\documentclass[11pt]{article}

\usepackage[final]{acl}
\usepackage{threeparttable}
\usepackage{booktabs}
\usepackage{multirow}
\usepackage{arydshln}

\usepackage{times}
\usepackage{latexsym}
\usepackage{listings}
\usepackage{float}
\newfloat{listing}{htbp}{lol}
\floatname{listing}{Listing}
\usepackage{enumitem}

\usepackage[T1]{fontenc}
\usepackage{pifont}

\usepackage[utf8]{inputenc}
\usepackage{amsmath}
\usepackage{microtype}

\usepackage{inconsolata}

\usepackage{graphicx}
\usepackage{placeins}
\usepackage{tikz}
\usetikzlibrary{shapes.geometric, arrows.meta, positioning, fit, backgrounds, calc}
\usepackage{tcolorbox}
\tcbuselibrary{breakable}
\usepackage{fontawesome5}
\usepackage{array}

\usepackage{titletoc}
\usepackage[dvipsnames,table]{xcolor}

\title{MedGame: Storytelling Gamification Empowered by \\ Large Language Models for Medical Education}

\author{
 \textbf{Qian Wu\textsuperscript{1}},
 \textbf{Xinrong Zhou\textsuperscript{1}},
 \textbf{Zizhan Ma\textsuperscript{1}},
 \textbf{Kai Chen\textsuperscript{1}},
 \textbf{Zheyao Gao\textsuperscript{1}}, 
   \textbf{Xun Lin\textsuperscript{1}}, \\
 \textbf{Hongqiu Wu\textsuperscript{4}},
 \textbf{Longfei Gou\textsuperscript{2}},
 \textbf{Yixiao Liu \textsuperscript{3}},
 \textbf{Ann Sin Nga Lau\textsuperscript{1}}, 
 \textbf{Qi Dou\textsuperscript{1}},
\\
 \textsuperscript{1}CUHK
 \textsuperscript{2}Southern Medical University 
 \textsuperscript{3}Peking University
 \textsuperscript{4}SJTU
\\
 \small{
   \textbf{(Work in progress)} \href{https://github.com/med-air/MedGame}{https://github.com/med-air/MedGame}
 }
}

\begin{document}
\maketitle
\begin{abstract}
Large Language Models (LLMs) show promise for medical education, but most existing systems focus on localized interactions such as question answering or single-turn feedback, rather than organizing an entire clinical case into a decision-centered learning trajectory. We introduce \textit{MedGame}, a framework that transforms static clinical cases into structured, executable storytelling games. MedGame uses a dual-engine design: a Medical Narrative Designer synthesizes case-grounded clinical storylines with states and decision nodes, while a Story Director converts them into dependency-aware multimodal orchestration plans rendered by our released interactive platform. We construct MedGame Bench, a 5,000-case benchmark and evaluation protocol for Medical Narrative Generation and Story Direction. Experiments show that task-specific fine-tuning substantially improves open-source LLMs on MedGame Bench and narrows the gap with commercial models. A pilot student study further shows that learners perceive MedGame as more engaging and useful than text-only alternatives. 
\end{abstract}

\section{Introduction}

The integration of Large Language Models (LLMs) into medical education has shown great promise~\cite{survey-mededu2,survey-mededu1}, supporting tasks such as clinical dialog tutoring~\cite{DDxTutor-ACL25} and communication practice~\cite{huang-etal-2024-benchmarking}. However, most existing systems operate at the level of localized interactions---individual questions, isolated dialog turns, or pointwise feedback---rather than organizing learning around a complete, decision-centered trajectory across an entire clinical encounter. This limitation is particularly salient for case-based learning, where the pedagogical value lies precisely in how a case progresses from presentation to resolution through a sequence of clinical decisions.

Clinical case summaries are a natural carrier of such trajectories: they encode rich diagnostic and therapeutic knowledge within a single coherent episode. Yet they are typically consumed as static text, presenting the reasoning process only in retrospect. Active clinical reasoning, in contrast, unfolds sequentially---learners must interpret partial information, decide what to ask or test next, and update their understanding as new evidence emerges. This mismatch between static delivery and dynamic reasoning motivates a technical question for LLM-based medical education: \textit{Can LLMs transform static clinical cases into structured, executable, and interactive learning trajectories?}

\begin{figure}[t]
\includegraphics[width=\columnwidth]{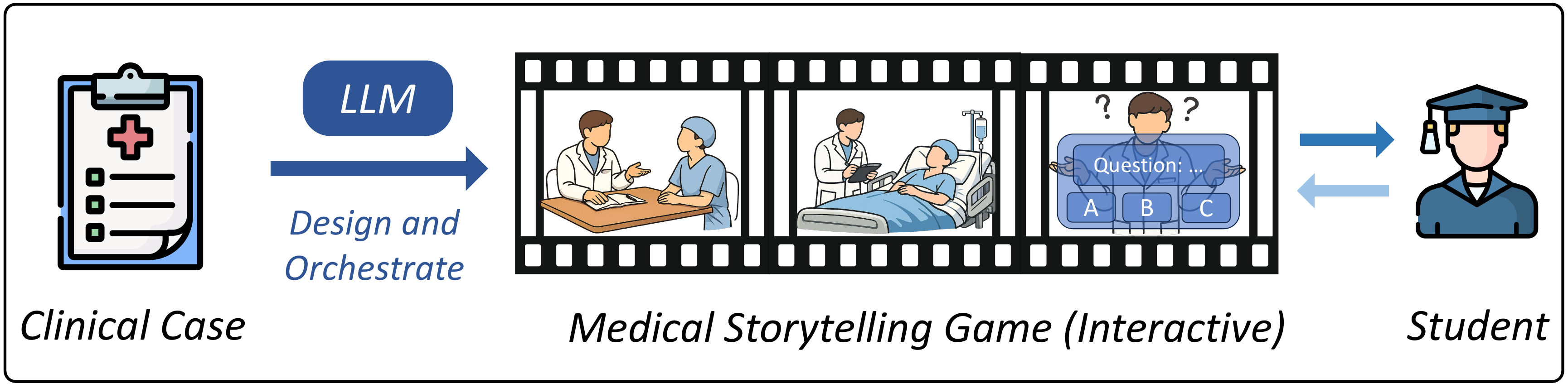}
\caption{MedGame at a glance. Given a static clinical case, MedGame uses LLMs to design and orchestrate an interactive medical storytelling game, where learners make decisions and receive feedback as the case unfolds.}
\label{fig:Intro}
\end{figure}

Transforming a case summary into such a trajectory requires more than fluent educational text: the output must be faithful to the source case, expose meaningful decision checkpoints, remain medically plausible, and stay machine-executable, making this a constrained structured generation problem.

Among possible representations, storytelling games naturally accommodate these requirements (Figure~\ref{fig:Intro}): scenes encode evolving clinical contexts, characters deliver role-specific information, decision nodes elicit learner reasoning, and feedback links choices to medical consequences. In this framing, gamification is not merely an entertainment layer; it serves as a structural interface for organizing case-based clinical reasoning into an unfolding sequence of states, choices, and outcomes~\cite{Gamification_and_Education,turning_stories_into_learning_journeys}.

In this paper, we propose \textit{MedGame}, a generative framework for transforming static clinical cases into structured storytelling games. Inspired by the professional film industry, where a continuity script bridges narrative intent and on-stage execution~\cite{Cinema_1,gomery2011movie}, MedGame adopts a dual-engine architecture that decouples clinical narrative ideation from technical orchestration. The \textit{Medical Narrative Designer} acts as the ``screenwriter'' by synthesizing raw clinical records into a hierarchical clinical storyline with explicit states and decision nodes. This structured storyline serves as the interface between the two engines: the \textit{Story Director} consumes it and maps it into a directed acyclic graph of multimodal generation primitives. By managing task dependencies through this graph, the Director supports identity preservation and causal coherence across downstream rendering and interaction.

Our \underline{contributions} are summarized as follows:

\textbf{(1)} We introduce \textit{MedGame}, a dual-engine framework for transforming static clinical cases into structured, executable storytelling games.
MedGame factorizes this transformation into Medical Narrative Generation, which produces a case-grounded clinical storyline with explicit states and decision nodes, and Story Direction, which converts the storyline into dependency-aware multimodal execution plans.

\textbf{(2)} We construct \textit{MedGame Bench}, a case-grounded benchmark and evaluation protocol for Medical Narrative Generation and Story Direction.
Built from 5,000 real patient cases, it evaluates clinically meaningful storyline generation and executable multimodal orchestration using metrics for structural validity, story adaptation, medical accuracy, educational quality, and task reasonability.

\textbf{(3)} We conduct systematic experiments showing that task-specific fine-tuning substantially improves open-source LLMs on MedGame Bench and narrows the gap with commercial models.
Expert and student studies further show that human revision strengthens generated educational content, while learners perceive the rendered MedGame experience as more engaging and useful than text-only interaction.

\textbf{(\textsc{Bonus})} To bridge concept and practice, we build and release an executable platform tailored to our MedGame framework to convert generated clinical storylines and orchestration plans into directly playable interactive games.
It allows generated cases to be inspected and experienced in an interactive environment, providing a practical interface for future research and educational prototyping.

\section{Related Works}

\subsection{LLMs for Medical Education}

\paragraph{Copilot-style Assistants}
LLM copilots act as knowledgeable mentors, providing real-time guidance and structured feedback. While early works focused on standardized exam performance \cite{kung2023performance, survey-mededu1}, recent research emphasizes pedagogical scaffolding. \citet{DDxTutor-ACL25} proposed to guide clinical reasoning via structured differential diagnosis steps, while \citet{jang-etal-2025-medtutor} introduced Retrieval-Augmented Generation (RAG)~\cite{RAG-nips20} based medical evidence generation. Beyond diagnostics, \citet{huang-etal-2024-benchmarking} developed \textit{ChatCoach} to provide feedback on communication skills, and \citet{yao-etal-2024-readme} utilized data-centric methods to simplify medical jargon for better patient understanding. Along this patient-education line, \citet{cai-etal-2023-paniniqa,jang-etal-2025-chatbot} further develops interactive question answering to help patients master discharge instructions with answer verification feedback.

\paragraph{Patient Simulations} While Standardized Patients (SPs) are the gold standard for clinical practice~\cite{anderson1994growing-standardized-patients,2025-JMIR-Standardized-Patient}, their reliance on human actors incurs high costs and limits scalability. LLM-driven Virtual Patients (VP) address these issues by enabling accessible, high-fidelity simulations. Recent comparative studies further examine how closely LLM-based VSPs can approximate human SPs in medical education~\cite{zhang-etal-2026-human}. In terms of domain-specific applications, \citet{yao-etal-2025-dischargesim} benchmarked VPs for discharge education, and \citet{yao2026rethinking} further frames patient education as multi-turn multimodal interaction, while \citet{EMNLP-patient-psi} simulated diverse cognitive models for mental health training. Beyond specific medical scenarios, recent studies have also focused on technical mechanisms to improve simulation realism. \citet{li2024agent} and \citet{ACL25-du2025evopatient} explored autonomous agent evolution and multi-agent co-evolution, respectively. Meanwhile, focusing on interaction dynamics, \citet{ACL2025Find-adaptiveVP} introduced a real-time adaptation mechanism to adjust virtual patients' hostility and cooperation levels based on trainees' communication strategies.

\paragraph{Generative Educational Content} LLMs facilitate the efficient production of diverse learning materials, significantly reducing the content creation burden on medical educators. For case-based learning, \citet{Medical-Teacher-Clinical-vignettes} demonstrated that LLM-generated clinical vignettes are comparable in quality to human-authored content. Similarly, in the domain of assessment, studies have validated the capability of LLMs to generate high-fidelity multiple-choice questions for specialized subjects such as anatomy~\cite{Medical-Teacher-Assessing-Generated-MCQs} and broader medical examinations~\cite{BMC-MedicalEducation-Review-GeneratingMedicalExams}. Extending generation capabilities to multimodal resources, \citet{HealthCards-EMNLP25} developed a framework to create visual flashcards for healthcare education via text-to-image pipelines.

\subsection{Game and Narrative Design with LLMs}
Gamification is a widely adopted strategy for enhancing engagement in general and medical education~\cite{Gamification_and_Education,gamification2,medgamification1,medgamification2}. Recent general-domain advances in LLM-based games provide essential technical foundations for this field.
On the development side, \citet{hongqiu-etal-2025-game} facilitated the democratization of content creation with \textit{ChatGE}, enabling game script generation via natural language. To improve interaction quality, \citet{wu-etal-2024-role, wu-etal-2025-towards-enhanced} proposed \textit{Drama-Interaction}, incorporating playwriting theories to elevate simple role-play into immersive narratives.
Crucially for narrative coherence, hierarchical planning frameworks such as \textit{StoryVerse}~\cite{wang2024storyverse} and \textit{CoDi}~\cite{wang2025codi} have addressed the tension between narrative freedom and coherence using Director-Actor architectures.
However, these systems are primarily designed for general-domain storytelling and do not account for the domain-specific constraints of medical education, such as clinical correctness, pedagogical sequencing, and multimodal enactment of professional roles.

\section{Problem Formulation}
\label{section: problem formulation and setup}

Drawing from narrative game design principles~\cite{AAAI-1999-interactive-drama,2005storyandNarrative,2019PlayerInteraction}, we formulate case-to-storytelling-game generation as a linear clinical storyline gated by decision nodes. The linear structure can be expressed as a sequence of clinical states gated by interactive sessions:
\begin{equation}
\label{eq:linear-story}
(\mathcal{C}_1 \xrightarrow{\mathcal{Q}_1} \mathcal{C}_2 \xrightarrow{\mathcal{Q}_2} \mathcal{C}_3 \xrightarrow{\mathcal{Q}_3} \dots \xrightarrow{\mathcal{Q}_{n-1}} \mathcal{C}_n)
\end{equation}
Each $\mathcal{C}_i$ denotes the accumulated \emph{clinical world state} after a sequence of confirmed observations and decisions (e.g., ``54-year-old female presenting with fatigue, memory loss, and normal initial labs''), while $\mathcal{Q}_i$ represents a gated decision checkpoint that must be resolved before the clinical state can advance (e.g., ``Which laboratory tests should be ordered?''). Each arrow $\xrightarrow{\mathcal{Q}_i}$ thus denotes a state transition gated by the learner's resolution of $\mathcal{Q}_i$: the next state $\mathcal{C}_{i+1}$ is determined by $\mathcal{C}_i$ together with that resolved decision checkpoint. This equation provides a \emph{flattened, process-level view} of storyline progression.

This linear formulation is motivated by the source format of medical cases rather than by an assumption that clinical reasoning itself is linear. Medical case reports and diagnostic records usually describe a deterministic clinical trajectory that has already occurred, organized along a temporal sequence of symptoms, examinations, diagnoses, treatments, and outcomes. Adapting such records into a linear storyline therefore preserves fidelity to the source case. In clinical decision-making, however, different learner actions may lead to different consequences, giving rise to branching structures in interactive simulation. Such branching story graphs can be constructed by composing the same state-transition units defined above, where different decisions instantiate different successor states. We discuss this extension in Appendix~\ref{appendix section: branching generalization}.

\begin{figure*}[t]
    \centering
    \includegraphics[width=\linewidth]{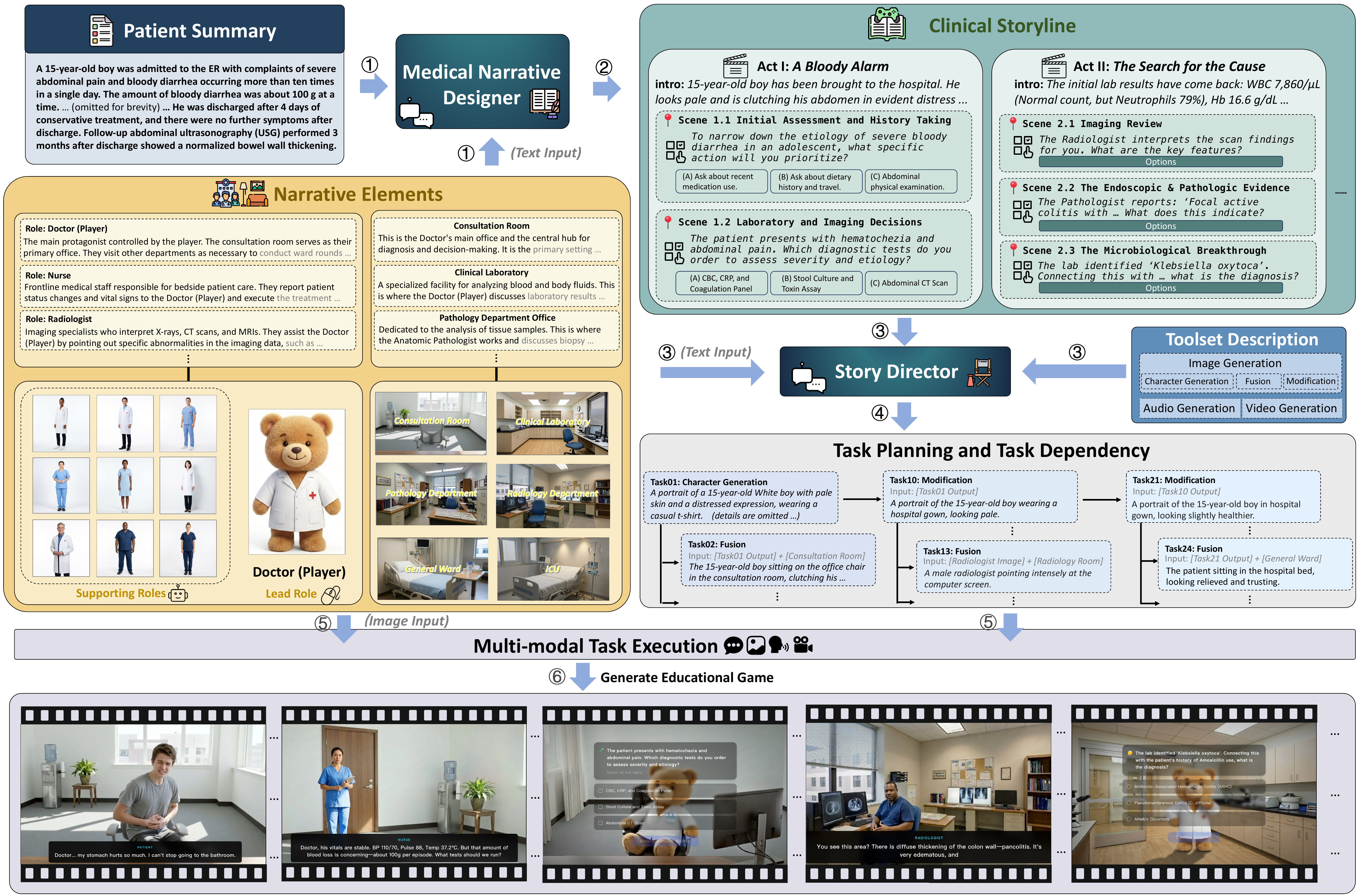}
\caption{\textbf{Overview of the MedGame framework architecture.}
MedGame follows a two-phase generative workflow.
In the script synthesis phase (Steps~\ding{172}--\ding{173}), the \textit{Medical Narrative Designer} transforms raw patient summaries and persistent narrative elements into a structured Clinical Storyline.
In the orchestration and execution phase (Steps~\ding{174}--\ding{177}), the \textit{Story Director} decomposes the storyline into multimodal generation primitives, organizes task planning and dependencies using a directed acyclic graph, and coordinates multimodal task execution to render the interactive educational experience.}

\label{fig:Main}
\end{figure*}

\subsection{Acts, Scenes, and Decision Nodes}
\label{subsection: Hierarchy of Acts, Scenes, and Nodes}
To operationalize the abstract progression in Eq.~\eqref{eq:linear-story} within a coherent clinical narrative, we organize each storyline into a hierarchical structure composed of \underline{Acts}, \underline{Scenes}, and \underline{Decision Nodes}, following established scriptwriting and narrative design practices. Table~\ref{tab:clinical_story_hierarchy} summarizes the role of each level in the generated Clinical Storyline.

\begin{table}[t]
\centering
\scriptsize
\setlength{\tabcolsep}{3pt}
\caption{Clinical Storyline hierarchy.}
\label{tab:clinical_story_hierarchy}
\begin{tabular}{@{}m{0.17\linewidth}m{0.39\linewidth}m{0.34\linewidth}@{}}
\toprule
\textbf{Level} & \textbf{Role in Storyline} & \textbf{Medical Example} \\
\midrule
Act ($A_i$) & Organizes a thematic macro-stage of the clinical pathway. & Diagnostic workup for a patient with unexplained fatigue. \\
\cdashline{1-3}
Scene ($S_{i,j}$) & Localizes one clinical step and its state transition. & Ordering laboratory tests after history taking and examination. \\
\cdashline{1-3}
Decision Node ($\mathcal{Q}$) & Presents a learner-facing decision checkpoint. & Selecting which laboratory tests should be ordered next. \\
\bottomrule
\end{tabular}
\end{table}

An \textbf{Act} $A_i$ defines a thematic macro-stage of the clinical pathway (e.g., initial presentation or treatment planning). Each Act comprises an ordered sequence of Scenes, each corresponding to an intermediate step within the same phase:
\begin{equation}
A_i \;=\; \langle S_{i,1},\, S_{i,2},\, \dots,\, S_{i,m_i} \rangle .
\end{equation}

A \textbf{Scene} $S_{i,j}$ represents a finer-grained clinical step within an Act, such as history taking, focused physical examination, diagnostic testing, or treatment planning. Each Scene localizes one transition in the clinical trajectory by presenting the context before an interaction, eliciting a learner decision, and revealing the updated clinical state after the decision is resolved.

A \textbf{Decision Node} is the learner-facing checkpoint inside a Scene. It asks the learner to make a clinical choice, such as selecting relevant tests, choosing a diagnosis, or deciding on a treatment action, and provides option-level feedback or explanatory consequences. In this way, Decision Nodes convert passive case information into active clinical reasoning steps.

Together, these abstractions define the \textit{Clinical Storyline} $\mathcal{T} = \langle A_1, A_2, \dots, A_N \rangle$, which serves as the \underline{narrative skeleton} of our gamified clinical cases and as the interface between the two engines of MedGame introduced in the next section. The complete and verifiable \textit{Pydantic}~\cite{colvin2023pydantic} schema is provided in Appendix~\ref{list: pydantic_models of clinical story}.

\section{The MedGame Framework}
\label{section: the MedGame Framework}

The overview of our framework is shown in Figure~\ref{fig:Main}. MedGame separates case adaptation from multimodal realization through two LLM-driven processes. The \textit{Medical Narrative Designer} first converts a raw patient summary into a structured, text-based interactive clinical storyline. The \textit{Story Director} then translates this storyline into a dependency-aware multimodal generation plan for rendering the interactive experience.

\subsection{Medical Narrative Designer}

Rather than directly asking an LLM to ``write a game'', the Medical Narrative Designer frames case adaptation as a source-grounded restructuring problem. Given a patient summary, the Designer must preserve the factual clinical trajectory of the case while turning passive case information into learner-facing interactions. This design separates three requirements that are otherwise entangled in free-form generation: fidelity to the source case, pedagogical decision design, and structural validity for downstream execution.

Formally, the Medical Narrative Designer, denoted $f_{\text{des}}$, transforms an unstructured patient summary $S_{\text{pat}}$ into a hierarchical \textit{Clinical Storyline} $\mathcal{T}$:
\begin{equation}
\mathcal{T} = f_{\text{des}}(S_{\text{pat}} \mid \Phi_{\text{des}}).
\end{equation}
The Designer performs this transformation through a structured adaptation process. It first anchors the generated content to clinically relevant facts in $S_{\text{pat}}$, then organizes the temporal progression of the case into Acts and Scenes, and finally introduces Decision Nodes at clinically meaningful moments. Each Decision Node asks the learner to choose an action, gather information, or reason over evidence, and is paired with feedback that connects the learner's choice to the corresponding clinical consequence or explanation.

The resulting storyline can already support a text-based interactive medical game: it contains a source-grounded clinical progression, learner choices, and feedback before any multimodal rendering is applied. The instruction $\Phi_{\text{des}}$ guides this transformation with four types of constraints. Clinical exemplars $\mathcal{E}$ demonstrate how source cases are adapted into interactive storylines; the structural Pydantic schema (Listing~\ref{list: pydantic_models of clinical story}) enforces a machine-readable hierarchy; medical rubrics $\mathcal{R}$ encourage clinically meaningful decisions and explanations; and the narrative elements $\{\mathcal{P}, \mathcal{L}\}$ define the available medical staff personas and clinical locations from which the Designer can select. These curated elements constrain character and scene choices, helping the generated storyline remain grounded and visually consistent. Together, these constraints are intended to keep the output faithful to the source case while making it usable by the Story Director. Details of the narrative elements are provided in Appendix~\ref{appendix section: Details of Narrative Elements}, and the full instruction is provided in Appendix~\ref{appendix section: prompt for narrative design}.

\subsection{Story Director}
\label{sec:director}

While the Clinical Storyline $\mathcal{T}$ can support a text-based interaction, MedGame further aims to render it as a multimodal simulation. The Story Director operates over a multimodal toolset $\mathcal{M}=\{f_{\text{vis}}, f_{\text{aud}}, f_{\text{vid}}\}$, whose interfaces expose visual, audio, and video generation capabilities. The Director, denoted $f_{\text{dir}}$, maps $\mathcal{T}$ into an executable generation plan represented as a directed acyclic graph (DAG) $\mathcal{G}=(V,E)$:
\begin{equation}
\mathcal{G} = f_{\text{dir}}(\mathcal{T} \mid \Phi_{\text{dir}}).
\end{equation}
Each node $v \in V$ is a self-contained generation primitive instantiated from the multimodal toolset, and each edge $(u \rightarrow v) \in E$ records an asset-level or state-level dependency between tasks. The instruction $\Phi_{\text{dir}}$ incorporates narrative elements, tool API specifications, and dependency rules to guide task decomposition. Details of the tool interfaces are provided in Appendix~\ref{appendix section: Details of Multimodal Toolset}, and the full instruction is provided in Appendix~\ref{appendix section: prompt for story director}.

This graph representation lets the Director plan beyond text-based interaction without assuming that all assets can be generated independently. For example, a later scene may need to reuse a character reference, inherit a scene layout, or modify a previously generated visual state. We therefore use \textbf{initial frame synthesis} (a static keyframe generated before downstream video synthesis) as a visual anchor for each generation task, and propagate upstream assets along validated dependencies to help maintain identity and narrative continuity. Detailed dependency specifications are provided in Appendix~\ref{appendix section: task dependency}.

Finally, the system materializes interactive assets by executing the generation primitives following $\mathcal{G}$. For each node $v_i$, the system selects a tool $f_{v_i} \in \mathcal{M}$ according to the node's modality and task type, and synthesizes the output asset $\mathbf{a}_{v_i}$ from the task specification and available upstream assets:
\[
\mathbf{a}_{v_i} = f_{v_i}\bigl(v_i \mid \{\mathbf{a}_u\}_{(u,v_i) \in E}\bigr),
\]
where $\{\mathbf{a}_u\}$ denotes the set of inherited assets from predecessor tasks. This dependency-aware execution provides a practical mechanism for coordinating multimodal generation, while keeping asset synthesis modular with respect to the underlying generation tools.

\section{MedGame Bench}
\label{section: MedGame Bench}

\begin{figure}[t]
    \centering
    \includegraphics[width=\linewidth]{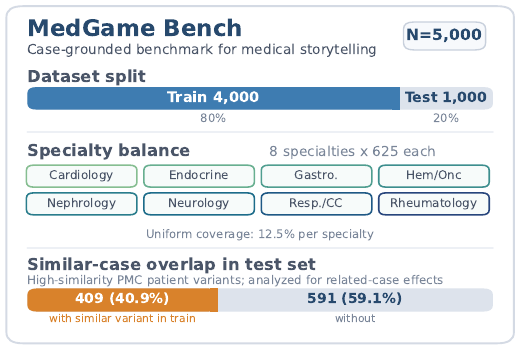}
    \caption{Overview of MedGame Bench. The benchmark contains 5,000 PMC-Patients cases split into 4,000 training and 1,000 test cases, balanced across eight medical specialties with 625 cases each. Among the test cases, 409 have at least one high-similarity patient variant in the training set; here, high-similarity variants refer to PMC-Patients relations with score 2, corresponding to patient summaries from the same source article. We analyze this related-case overlap in Appendix~\ref{appendix subsection: similar patient analysis}.}
    \label{fig:medgame_bench_overview}
\end{figure}

We construct \textit{MedGame Bench}, a case-grounded benchmark for evaluating whether LLMs can transform static medical cases into structured interactive storylines and executable orchestration plans for downstream rendering. The benchmark is built from public patient summaries in the PMC-Patients Dataset~\cite{PMC_Patient}, which consists of case-report-derived summaries from \textit{PubMed Central}\footnote{https://pmc.ncbi.nlm.nih.gov/}. To support broad clinical coverage, we sample 5,000 cases balanced across eight medical specialties---cardiology, endocrinology, gastroenterology, hematology/oncology, nephrology, neurology, respiratory and critical care, and rheumatology---and partition them into a fixed 4,000/1,000 train--test split used throughout Section~\ref{section: experiments} (Figure~\ref{fig:medgame_bench_overview}). We additionally track high-similarity patient variants across the split to assess whether related-case overlap affects evaluation scores. Details about generated-output statistics are provided in Appendix~\ref{appendix section: more details about dataset}.

\noindent\textbf{Benchmark Tasks.}
MedGame Bench contains two evaluation tracks aligned with the two-engine design of MedGame. \textbf{Medical Narrative Generation} evaluates whether a model can convert a patient summary into a structured Clinical Storyline with valid Acts, Scenes, Decision Nodes, learner choices, and feedback. \textbf{Story Direction} evaluates whether a model can translate a Clinical Storyline into a valid multimodal orchestration plan for downstream rendering, with appropriate resource assignment, tool selection, parameters, and dependency modeling.

\noindent\textbf{Evaluation Protocol.}
For Medical Narrative Generation, we evaluate generated storylines across four dimensions. \textbf{(1)} \underline{\textit{Structure Validation}} verifies JSON parsing (JSON), Pydantic schema conformance (Schema), and strict business logic constraints (Strict). \textbf{(2)} \underline{\textit{Story Adaptation}} assesses Clinical Content Integration (CCI), Character \& Scene Usage (CSU), and Narrative Quality (NQ). \textbf{(3)} \underline{\textit{Medical Accuracy}} evaluates Clinical Decision Accuracy (CDA), Option Design Accuracy (ODA), and Medical Explanation Accuracy (MEA). \textbf{(4)} \underline{\textit{Educational Quality}} measures Question Design Quality (QDQ) and Feedback Quality (FQ).

For Story Direction, we evaluate generated task parameters across two dimensions. \textbf{(1)} \underline{\textit{Structure Validation}} verifies whether outputs conform to the required format, measured at both sample level (Sample-wise) and individual task level (Task-wise). A sample passes validation only when all constituent tasks satisfy the schema and business logic constraints. \textbf{(2)} \underline{\textit{Task Reasonability}} assesses Resource Assignment (RA), API Type Selection (ATS), and Parameter Content (PC), measuring whether generated task plans select appropriate resources, follow tool API usage rules, and provide parameters that match the narrative context.

Structure Validation is performed with automated rule-based checking, while content-oriented indicators are scored by LLM-as-a-Judge~\cite{llm-as-a-judge} using GPT-5.2\footnote{https://openai.com/index/introducing-gpt-5-2/}, with each indicator rated on a 1--10 scale. Complete rubrics, validation rules, and evaluation prompts are provided in Appendix~\ref{appendix section: Evaluation Details}.

\begin{table*}[t]
    \centering
\caption{Quantitative results on Medical Narrative Generation. \ding{108} denotes open-source models and \ding{117} denotes commercial models. \textbf{Bold} indicates the best performance, \underline{underlined} the second-best, $^*$ denotes fine-tuned models.}
\resizebox{0.86\linewidth}{!}
{

\begin{tabular}{l|ccc|ccc|ccc|cc}
\toprule
\multirow{2}{*}{Model} & \multicolumn{3}{c|}{Structure Validation} & \multicolumn{3}{c|}{Story Adaptation} & \multicolumn{3}{c|}{Medical Accuracy} & \multicolumn{2}{c}{Educational} \\
\cmidrule(lr){2-4} \cmidrule(lr){5-7} \cmidrule(lr){8-10} \cmidrule(lr){11-12}
 & JSON$\uparrow$ & Schema$\uparrow$ & Strict$\uparrow$ & CCI$\uparrow$ & CSU$\uparrow$ & NQ$\uparrow$ & CDA$\uparrow$ & ODA$\uparrow$ & MEA$\uparrow$ & QDQ$\uparrow$ & FQ$\uparrow$ \\
\midrule
\ding{108} Qwen3-32B & 87.4 & 84.5 & 79.4 & 7.04 & 5.41 & 6.92 & 5.78 & 4.95 & 5.04 & 6.21 & 5.94 \\
\ding{108} Gemma-3-27B & 95.6 & 69.2 & 58.7 & 7.67 & 7.11 & 7.21 & 6.08 & 5.82 & 5.43 & 6.12 & 6.02 \\
\ding{108} MedGemma-27B & 95.3 & 76.1 & 59.3 & 7.25 & 7.02 & 7.53 & 6.43 & 6.22 & 5.78 & 6.53 & 6.61 \\
\ding{108} Qwen3.5-27B & 92.1 & 91.7 & 90.8 & 8.43 & 8.51 & 7.89 & 6.82 & 6.63 & 6.32 & 6.89 & 6.92 \\
\ding{117} Claude-Sonnet-4.5 & 99.7 & 99.5 & \underline{99.5} & \underline{8.74} & 6.99 & 7.99 & 7.01 & 6.77 & \underline{6.73} & \textbf{7.33} & \textbf{7.52} \\
\ding{117} Gemini-3-Pro & \textbf{100.0} & \textbf{100.0} & \textbf{100.0} & \textbf{8.78} & \textbf{8.65} & \textbf{8.02} & \textbf{7.26} & \textbf{7.20} & \textbf{6.92} & \underline{7.10} & \underline{7.36} \\
\cdashline{1-12}
\ding{108} Qwen3.5-9B$^*$ & 93.8 & 91.7 & 90.9 & 7.95 & 8.30 & 7.67 & 6.55 & 6.13 & 5.71 & 6.72 & 6.54 \\
\ding{108} Qwen3-32B$^*$ & 99.6 & 99.3 & 99.1 & 8.23 & 8.50 & 7.95 & 6.60 & 6.31 & 5.87 & 6.82 & 6.76 \\
\ding{108} Gemma-3-27B$^*$ & 99.7 & 99.4 & 98.9 & 7.77 & 8.41 & 7.86 & 6.39 & 6.01 & 5.56 & 6.79 & 6.57 \\
\ding{108} MedGemma-27B$^*$ & 99.2 & 99.0 & 98.5 & 8.18 & 8.42 & 7.92 & 6.64 & 6.32 & 5.90 & 6.88 & 6.83 \\
\ding{108} Qwen3.5-27B$^*$ & \underline{99.9} & \underline{99.8} & \underline{99.5} & 8.62 & \underline{8.62} & \underline{8.00} & \underline{7.05} & \underline{6.85} & 6.58 & 6.99 & 7.12 \\
\bottomrule
\end{tabular}

}
\label{tab:storygen_main}
\end{table*}

\section{Experiments}

\label{section: experiments}
We organize our experiments around two perspectives. \textbf{RQ1} and \textbf{RQ2} evaluate LLM capabilities on the two core MedGame generation tasks, including case-grounded medical narrative generation and multimodal story direction, before and after task-specific fine-tuning. \textbf{RQ3} and \textbf{RQ4} examine MedGame in collaborative and learner-facing use.

\subsection{Implementation Details}
We evaluate both MedGame generation tasks with a mixture of open-source general-purpose LLMs, medical LLMs, and commercial frontier models, including Qwen~\cite{yang2025qwen3}, Gemma~\cite{team2024gemma}, MedGemma~\cite{sellergren2025medgemma}, Claude, and Gemini~\cite{team2023gemini}. For Medical Narrative Generation, prompted baselines use a 2-shot setting; for Story Direction, prompted baselines use a zero-shot setting. We further fine-tune open-source models on the corresponding training split of MedGame Bench, with the complete model list shown in Tables~\ref{tab:storygen_main} and \ref{tab:filmgen_main}.

Open-source models in the fine-tuned setting are adapted with LoRA~\cite{hu2022lora} in Rank 32 and Alpha 64.
Local inference is performed with the vLLM framework~\cite{vllm}.
Open-source model training and inference are performed on a server with 2 H100 GPUs, while commercial models are accessed through official APIs.

\subsection{RQ1: How Do Existing LLMs Perform on MedGame Bench?}

\begin{table}[H]
    \centering
\caption{Quantitative results on Story Direction.}
\scriptsize
\resizebox{0.99\linewidth}{!}{
\begin{tabular}{l|cc|ccc}
\toprule
\multirow{2}{*}{Model} & \multicolumn{2}{c|}{Structure Validation} & \multicolumn{3}{c}{Task Reasonability} \\
\cmidrule(lr){2-3} \cmidrule(lr){4-6}
 & Sample-wise$\uparrow$ & Task-wise$\uparrow$ & RA$\uparrow$ & ATS$\uparrow$ & PC$\uparrow$ \\
\midrule
\ding{108} Gemma-3-27B & 56.50 & 93.90 & 5.12 & 4.66 & 7.08 \\
\ding{108} MedGemma-27B & 59.40 & 89.30 & 5.39 & 5.17 & 7.01 \\
\ding{108} Qwen3-32B & 77.20 & 94.51 & 6.16 & 6.98 & 6.95 \\
\ding{108} Qwen3.5-27B & 80.30 & 95.26 & 6.34 & 7.01 & 6.97 \\
\ding{117} Claude-Sonnet-4.5 & 99.40 & 99.87 & 7.89 & \underline{8.97} & 7.72 \\
\ding{117} Gemini-3-Pro & 99.20 & 99.93 & \textbf{7.93} & 8.95 & \underline{7.85} \\
\cdashline{1-6}
\ding{108} Qwen3.5-9B$^*$ & 95.00 & 97.90 & 7.67 & 8.85 & 7.81 \\
\ding{108} Qwen3-32B$^*$ & 99.10 & 99.80 & \underline{7.92} & 8.86 & 7.84 \\
\ding{108} Gemma-3-27B$^*$ & 99.50 & 99.74 & 7.69 & 8.77 & \textbf{7.86} \\
\ding{108} MedGemma-27B$^*$ & \underline{99.60} & \underline{99.97} & 7.71 & 8.84 & \underline{7.85} \\
\ding{108} Qwen3.5-27B$^*$ & \textbf{99.80} & \textbf{99.98} & 7.88 & \textbf{9.02} & \textbf{7.86} \\
\bottomrule
\end{tabular}
}
\label{tab:filmgen_main}
\end{table}

For RQ1, we evaluate existing LLMs without task-specific fine-tuning, corresponding to the results above the dashed line in Tables~\ref{tab:storygen_main} and \ref{tab:filmgen_main}. For Medical Narrative Generation, commercial frontier models achieve near-perfect structural validity, with Claude-Sonnet-4.5 and Gemini-3-Pro reaching 99.5\% and 100.0\% Strict validation. Open-source models lag across both structure and content metrics: Qwen3-32B reaches only 79.4\% Strict validation, Gemma-3-27B and MedGemma-27B remain below 60\%, and their medical accuracy scores are consistently lower than commercial frontier models. Even for Gemini-3-Pro, CDA, ODA, and MEA remain around 7, suggesting that generated medical content still needs careful review.

For Story Direction, the gap becomes more operational. Open-source models reach only 56.50--80.30\% Sample-wise validation and also trail commercial models on task reasonability, especially API Type Selection. By contrast, Claude-Sonnet-4.5 and Gemini-3-Pro exceed 99\% Sample-wise validation, showing that orchestration is feasible for stronger models but unreliable for open-source models without adaptation.

\textbf{Overall,} commercial frontier models show that MedGame is feasible for strong LLMs, while open-source models lag across content quality, task reasonability, and structural validity. The structural gap is especially consequential: without valid storylines and task plans, generated outputs cannot be reliably parsed, executed, or rendered, motivating RQ2's study of task-specific fine-tuning.

\subsection{RQ2: Can Open-Source Models Be Adapted through Fine-Tuning?}

Building on RQ1, we examine whether open-source models can acquire MedGame-specific generation conventions through task-specific fine-tuning on Gemini-3-Pro reference trajectories. The gains are consistent across both tracks (Tables~\ref{tab:storygen_main} and \ref{tab:filmgen_main}): Qwen3.5-27B$^*$ reaches 99.5\% Strict validation on Medical Narrative Generation, while Qwen3-32B$^*$ improves CSU from 5.41 to 8.50, suggesting better use of curated personas and clinical locations. More importantly, Qwen3.5-27B$^*$ nearly matches Gemini-3-Pro on narrative structure and adaptation, and slightly surpasses Claude-Sonnet-4.5 when summing the content-oriented Medical Narrative Generation indicators (59.83 vs.\ 59.08). For Story Direction, Qwen3.5-27B$^*$ achieves the best Sample-wise and Task-wise validation scores and the highest API Type Selection score, while remaining competitive on Resource Assignment and Parameter Content.

We also test whether these gains come from related-case overlap rather than task learning. Test cases with and without similar patient variants in the training split show no significant performance difference ($p>0.05$; Appendix~\ref{appendix subsection: similar patient analysis}), providing \textbf{no evidence of score inflation from related cases}. Appendix Table~\ref{tab:medical_llm_medical_accuracy} shows matched medical LLMs outperform general-purpose counterparts after the same Medical Narrative Generation fine-tuning (+0.43, +0.14, and +0.31 across 8B, 32B, and 27B pairs). Thus, task-specific fine-tuning helps open-source models learn MedGame's structure and interaction requirements, while \textbf{medical-domain specialization in the base model provides complementary gains in clinical accuracy}.

\subsection{RQ3: LLM-as-a-Judge Reliability and Expert-in-the-Loop Revision}

We first validate whether our LLM-as-a-Judge protocol provides a reliable signal for expert-facing analysis. Three medical Ph.D.\ holders evaluated 50 Medical Narrative Generation cases, and two experienced game developers evaluated 50 Story Direction cases, using the same rubrics as GPT-5.2 while blinded to model identities. As shown in Appendix Figure~\ref{fig:human_eval}, LLM scores correlate with human judgments, with strong agreement for Medical Narrative Generation (avg.\ $r=0.81$) and moderate agreement for Story Direction (avg.\ $r=0.61$).

Given the importance of clinical accuracy in medical education, we further study expert-in-the-loop refinement on 10 low-scoring generated story drafts per model from Gemini-3-Pro and Qwen3.5-27B$^*$. Experts marked targeted revision regions (e.g., inaccurate decisions, misleading options, incomplete explanations, and weak feedback), averaging 8.3 and 8.9 regions per sample, respectively. As shown in Table~\ref{tab:progressive_revision}, full expert revision substantially improves both medical accuracy and educational quality: for Gemini, the average Medical Accuracy / Educational scores increase from 6.40 / 6.45 to 8.07 / 7.95; for Qwen3.5-27B$^*$, they increase from 6.20 / 6.35 to 7.87 / 7.75. These results highlight expert-in-the-loop authoring as valuable for high-stakes medical education applications.

\subsection{RQ4: Students' Perception of MedGame}

\begin{figure}[t]
    \centering
    \includegraphics[width=\linewidth]{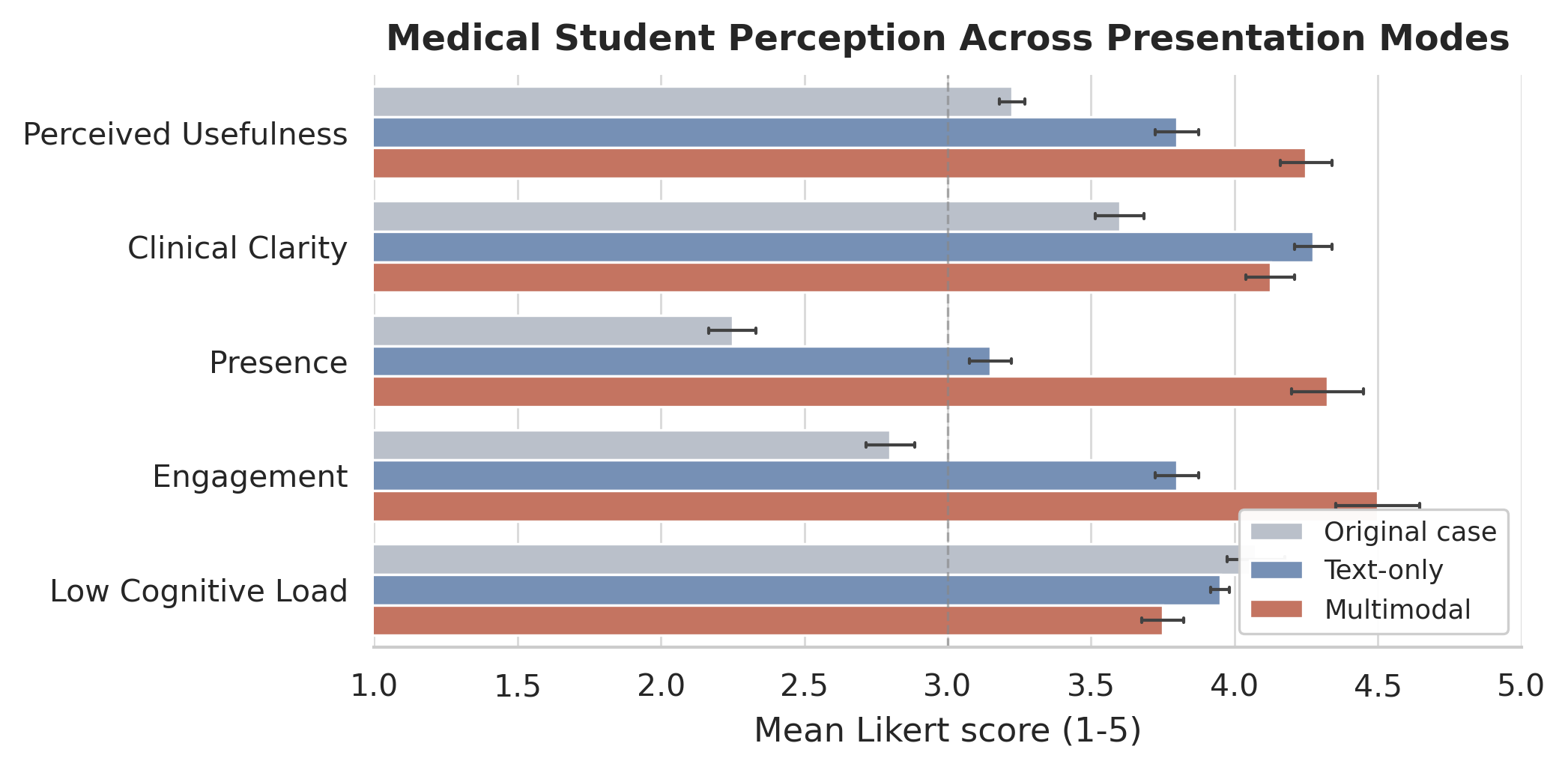}
    \caption{Medical student perception across three presentation modes. Eight senior medical students evaluated the same five clinical cases under original case, text-only MedGame, and multimodal MedGame conditions. Higher values indicate better perceived quality.}
    \label{fig:student_perception}
\end{figure}

To examine whether multimodal rendering provides value beyond static case material and text-only storylines, we conduct a paired pilot perception study with eight senior medical students. Each student evaluates the same five clinical cases under original case, text-only MedGame, and multimodal MedGame conditions, rating engagement, presence, clinical clarity, usefulness, and cognitive load (higher = lower) on a 1--5 Likert scale; the questionnaire items are provided in Appendix~\ref{appendix subsection: student perception questionnaire}. We average scores over cases for each student before statistical testing to avoid treating case-level ratings as independent samples.

As shown in Figure~\ref{fig:student_perception}, multimodal MedGame achieves the highest overall perception score (4.19), followed by text-only MedGame (3.79) and original case material (3.19); the multimodal advantage over text-only is significant under a one-sided paired Wilcoxon signed-rank test over student-level means ($p=0.0039$). \textbf{Multimodal rendering mainly improves engagement, presence, and perceived usefulness, while maintaining comparable clinical clarity but introducing a mild cognitive-load trade-off.} These results do not establish downstream learning gains, but they provide initial learner-facing evidence that the Story Director's multimodal rendering adds perceived value beyond text-only interaction.

\section{Conclusion}
We presented MedGame, a dual-engine framework that turns static cases into executable storytelling games. It separates Medical Narrative Generation from Story Direction, producing case-grounded storylines and dependency-aware multimodal orchestration plans. Task-specific fine-tuning improves open-source LLMs on structural validity, narrative adaptation, and orchestration reliability, while medical accuracy remains a bottleneck addressed by domain pretraining and expert revision. A learner-side pilot suggests multimodal rendering improves engagement and presence over text-only and static formats. Together, these results support MedGame as a practical step toward promising and scalable interactive medical educational content generation.
\section*{Limitations}

(1) This work focuses on the technical formulation and system-level realization of narrative-driven medical gamification, rather than on measuring long-term pedagogical outcomes. Given the time horizon and evaluation scope supported by the present study, we are unable to conduct longitudinal educational tracking experiments. Accordingly, our evaluation emphasizes generation-centric criteria and perceived learner experience, including structural validity, narrative coherence, medical accuracy, engagement, and usefulness. This scope is consistent with prior technical work on educational systems~\cite{LMScienceTutor,EMNLP2025-MathTutorRL}, where system design and content-generation quality are often evaluated before large-scale longitudinal learning studies. Assessing outcomes such as knowledge retention, clinical reasoning improvement, or skill transfer would require controlled studies with rigorous institutional oversight, and remains an important direction for future work.

(2) This work treats multimodal generation modules, including image, audio, and video generation, as external tools for rendering the generated storylines and orchestration plans. We do not systematically study how the performance of these tools affects the overall MedGame framework. Instead, our focus is on the NLP-centered capabilities of LLMs as medical narrative designers and story directors: converting clinical cases into structured interactive storylines and executable multimodal orchestration plans.

\section*{Ethical Considerations}

MedGame is intended as an educational authoring and simulation tool, not as an autonomous clinical decision-making system. Generated cases, explanations, and feedback should be reviewed by instructors or qualified medical experts before use with learners, consistent with oversight frameworks such as the EU Artificial Intelligence Act~\cite{smuha2025regulation}.

In our experiments, we study whether open-source models can learn MedGame-specific generation conventions from Gemini-3-Pro outputs. This use is intended solely for non-commercial research and experimental exploration. We use Gemini-3-Pro outputs in accordance with the applicable Gemini Terms of Service, and any data or model artifacts derived from them will be released only under a non-commercial license.

\bibliography{reference}

\begin{thebibliography}{55}
\providecommand{\natexlab}[1]{#1}

\bibitem[{Abd-Alrazaq et~al.(2023)Abd-Alrazaq, AlSaad, Alhuwail, Ahmed, Healy, Latifi, Aziz, Damseh, Alrazak, and Sheikh}]{survey-mededu2}
Alaa Abd-Alrazaq, Rawan AlSaad, Dari Alhuwail, Arfan Ahmed, Padraig~Mark Healy, Syed Latifi, Sarah Aziz, Rafat Damseh, Sadam~Alabed Alrazak, and Javaid Sheikh. 2023.
\newblock Large language models in medical education: opportunities, challenges, and future directions.
\newblock \emph{JMIR medical education}, 9(1):e48291.

\bibitem[{Anderson et~al.(1994)Anderson, Stillman, and Wang}]{anderson1994growing-standardized-patients}
M~Brownell Anderson, Paula~L Stillman, and Youde Wang. 1994.
\newblock Growing use of standardized patients in teaching and evaluation in medical education.
\newblock \emph{Teaching and Learning in Medicine: An International Journal}, 6(1):15--22.

\bibitem[{Artsi et~al.(2024)Artsi, Sorin, Konen, Glicksberg, Nadkarni, and Klang}]{BMC-MedicalEducation-Review-GeneratingMedicalExams}
Yaara Artsi, Vera Sorin, Eli Konen, Benjamin~S Glicksberg, Girish Nadkarni, and Eyal Klang. 2024.
\newblock Large language models for generating medical examinations: systematic review.
\newblock \emph{BMC medical education}, 24(1):354.

\bibitem[{Brunetti et~al.(2024)Brunetti, Ferrante, Avella, Indraccolo, and Del~Gatto}]{turning_stories_into_learning_journeys}
Riccardo Brunetti, Silvia Ferrante, Anna~Maria Avella, Allegra Indraccolo, and Claudia Del~Gatto. 2024.
\newblock Turning stories into learning journeys: the principles and methods of immersive education.
\newblock \emph{Frontiers in Psychology}, 15:1471459.

\bibitem[{Cai et~al.(2023)Cai, Yao, Liu, Wang, Reilly, Zhou, Li, Cao, Kapoor, Bajracharya, Berlowitz, and Yu}]{cai-etal-2023-paniniqa}
Pengshan Cai, Zonghai Yao, Fei Liu, Dakuo Wang, Meghan Reilly, Huixue Zhou, Lingxi Li, Yi~Cao, Alok Kapoor, Adarsha Bajracharya, Dan Berlowitz, and Hong Yu. 2023.
\newblock \href {https://doi.org/10.1162/tacl_a_00616} {{P}anini{QA}: Enhancing patient education through interactive question answering}.
\newblock \emph{Transactions of the Association for Computational Linguistics}, 11:1518--1536.

\bibitem[{Caponetto et~al.(2014)Caponetto, Earp, Ott et~al.}]{Gamification_and_Education}
Ilaria Caponetto, Jeffrey Earp, Michela Ott, and 1 others. 2014.
\newblock Gamification and education: A literature review.
\newblock In \emph{European conference on games based learning}, volume~1, pages 50--57.

\bibitem[{Carstensdottir et~al.(2019)Carstensdottir, Kleinman, and El-Nasr}]{2019PlayerInteraction}
Elin Carstensdottir, Erica Kleinman, and Magy~Seif El-Nasr. 2019.
\newblock Player interaction in narrative games: structure and narrative progression mechanics.
\newblock In \emph{Proceedings of the 14th international conference on the foundations of digital games}, pages 1--9.

\bibitem[{Chevalier et~al.(2024)Chevalier, Geng, Wettig, Chen, Mizera, Annala, Aragon, Fanlo, Frieder, Machado, Prabhakar, Thieu, Wang, Wang, Wu, Xia, Xia, Yu, Zhu, Ren, Arora, and Chen}]{LMScienceTutor}
Alexis Chevalier, Jiayi Geng, Alexander Wettig, Howard Chen, Sebastian Mizera, Toni Annala, Max Aragon, Arturo~Rodriguez Fanlo, Simon Frieder, Simon Machado, Akshara Prabhakar, Ellie Thieu, Jiachen~T. Wang, Zirui Wang, Xindi Wu, Mengzhou Xia, Wenhan Xia, Jiatong Yu, Junjie Zhu, and 3 others. 2024.
\newblock \href {https://openreview.net/forum?id=WFyolnFZOR} {Language models as science tutors}.
\newblock In \emph{Forty-first International Conference on Machine Learning}.

\bibitem[{Colvin et~al.(2023)Colvin, Jolibois, Ramezani, Garcia~Badaracco, Dorsey, Montague, Matveenko, Trylesinski, Runkle, Hewitt et~al.}]{colvin2023pydantic}
Samuel Colvin, Eric Jolibois, Hasan Ramezani, Adrian Garcia~Badaracco, Terrence Dorsey, David Montague, Serge Matveenko, Marcelo Trylesinski, Sydney Runkle, David Hewitt, and 1 others. 2023.
\newblock Pydantic.
\newblock \emph{Zenodo}.

\bibitem[{Co{\c{s}}kun et~al.(2025)Co{\c{s}}kun, K{\i}yak, and Budako{\u{g}}lu}]{Medical-Teacher-Clinical-vignettes}
{\"O}zlem Co{\c{s}}kun, Yavuz~Selim K{\i}yak, and I{\c{s}}{\i}l~{\.I}rem Budako{\u{g}}lu. 2025.
\newblock Chatgpt to generate clinical vignettes for teaching and multiple-choice questions for assessment: A randomized controlled experiment.
\newblock \emph{Medical teacher}, 47(2):268--274.

\bibitem[{Dinucu-Jianu et~al.(2025)Dinucu-Jianu, Macina, Daheim, Hakimi, Gurevych, and Sachan}]{EMNLP2025-MathTutorRL}
David Dinucu-Jianu, Jakub Macina, Nico Daheim, Ido Hakimi, Iryna Gurevych, and Mrinmaya Sachan. 2025.
\newblock From problem-solving to teaching problem-solving: Aligning llms with pedagogy using reinforcement learning.
\newblock In \emph{Proceedings of the 2025 Conference on Empirical Methods in Natural Language Processing}, pages 272--292.

\bibitem[{Du et~al.(2025)Du, LujieZheng, Hu, Xu, Li, Sun, Chen, Wu, Cai, and Ying}]{ACL25-du2025evopatient}
Zhuoyun Du, LujieZheng LujieZheng, Renjun Hu, Yuyang Xu, Xiawei Li, Ying Sun, Wei Chen, Jian Wu, Haolei Cai, and Haochao Ying. 2025.
\newblock Llms can simulate standardized patients via agent coevolution.
\newblock In \emph{Proceedings of the 63rd Annual Meeting of the Association for Computational Linguistics (Volume 1: Long Papers)}, pages 17278--17306.

\bibitem[{Elzayyat et~al.(2025)Elzayyat, Mohammad, and Zaqout}]{Medical-Teacher-Assessing-Generated-MCQs}
Maram Elzayyat, Janatul~Naeim Mohammad, and Sami Zaqout. 2025.
\newblock Assessing llm-generated vs. expert-created clinical anatomy mcqs: a student perception-based comparative study in medical education.
\newblock \emph{Medical Education Online}, 30(1):2554678.

\bibitem[{Gomery and Pafort-Overduin(2011)}]{gomery2011movie}
Douglas Gomery and Clara Pafort-Overduin. 2011.
\newblock \emph{Movie history: A survey}.
\newblock Routledge.

\bibitem[{Hong et~al.(2025)Hong, Wu, and Zhao}]{hongqiu-etal-2025-game}
Jiale Hong, Hongqiu Wu, and Hai Zhao. 2025.
\newblock \href {https://doi.org/10.18653/v1/2025.acl-long.218} {Game development as human-{LLM} interaction}.
\newblock In \emph{Proceedings of the 63rd Annual Meeting of the Association for Computational Linguistics (Volume 1: Long Papers)}, pages 4333--4354, Vienna, Austria. Association for Computational Linguistics.

\bibitem[{Hu et~al.(2022)Hu, Shen, Wallis, Allen-Zhu, Li, Wang, Wang, Chen et~al.}]{hu2022lora}
Edward~J Hu, Yelong Shen, Phillip Wallis, Zeyuan Allen-Zhu, Yuanzhi Li, Shean Wang, Lu~Wang, Weizhu Chen, and 1 others. 2022.
\newblock Lora: Low-rank adaptation of large language models.
\newblock \emph{ICLR}, 1(2):3.

\bibitem[{Huang et~al.(2024)Huang, Wang, Liu, Wang, and Wang}]{huang-etal-2024-benchmarking}
Hengguan Huang, Songtao Wang, Hongfu Liu, Hao Wang, and Ye~Wang. 2024.
\newblock \href {https://doi.org/10.18653/v1/2024.findings-acl.94} {Benchmarking large language models on communicative medical coaching: A dataset and a novel system}.
\newblock In \emph{Findings of the Association for Computational Linguistics: ACL 2024}, pages 1624--1637, Bangkok, Thailand. Association for Computational Linguistics.

\bibitem[{{Intelligent Internet}(2025{\natexlab{a}})}]{2025II-Medical-32B-Preview}
{Intelligent Internet}. 2025{\natexlab{a}}.
\newblock \href {https://huggingface.co/Intelligent-Internet/II-Medical-32B-Preview} {Ii-medical-32b-preview: Medical reasoning model}.

\bibitem[{{Intelligent Internet}(2025{\natexlab{b}})}]{2025II-Medical-8B}
{Intelligent Internet}. 2025{\natexlab{b}}.
\newblock \href {https://huggingface.co/Intelligent-Internet/II-Medical-8B} {Ii-medical-8b: Medical reasoning model}.

\bibitem[{Jang et~al.(2025{\natexlab{a}})Jang, Shangguan, Tegtmeyer, Gupta, Czerminski, Chheang, and Cohan}]{jang-etal-2025-medtutor}
Dongsuk Jang, Ziyao Shangguan, Kyle Tegtmeyer, Anurag Gupta, Jan~T Czerminski, Sophie Chheang, and Arman Cohan. 2025{\natexlab{a}}.
\newblock \href {https://doi.org/10.18653/v1/2025.emnlp-demos.24} {{M}ed{T}utor: A retrieval-augmented {LLM} system for case-based medical education}.
\newblock In \emph{Proceedings of the 2025 Conference on Empirical Methods in Natural Language Processing: System Demonstrations}, pages 319--353, Suzhou, China. Association for Computational Linguistics.

\bibitem[{Jang et~al.(2025{\natexlab{b}})Jang, Tran, Mistry, Gandluri, Zhang, Sultana, Kwon, Zhang, Yao, and Yu}]{jang-etal-2025-chatbot}
Won~Seok Jang, Hieu Tran, Manav~Shaileshkumar Mistry, Sai~Kiran Gandluri, Yifan Zhang, Sharmin Sultana, Sunjae Kwon, Yuan Zhang, Zonghai Yao, and Hong Yu. 2025{\natexlab{b}}.
\newblock \href {https://doi.org/10.18653/v1/2025.findings-emnlp.351} {Chatbot to help patients understand their health}.
\newblock In \emph{Findings of the Association for Computational Linguistics: EMNLP 2025}, pages 6598--6627, Suzhou, China. Association for Computational Linguistics.

\bibitem[{Kung et~al.(2023)Kung, Cheatham, Medenilla, Sillos, De~Leon, Elepa{\~n}o, Madriaga, Aggabao, Diaz-Candido, Maningo et~al.}]{kung2023performance}
Tiffany~H Kung, Morgan Cheatham, Arielle Medenilla, Czarina Sillos, Lorie De~Leon, Camille Elepa{\~n}o, Maria Madriaga, Rimel Aggabao, Giezel Diaz-Candido, James Maningo, and 1 others. 2023.
\newblock Performance of chatgpt on usmle: potential for ai-assisted medical education using large language models.
\newblock \emph{PLoS digital health}, 2(2):e0000198.

\bibitem[{Kwon et~al.(2023)Kwon, Li, Zhuang, Sheng, Zheng, Yu, Gonzalez, Zhang, and Stoica}]{vllm}
Woosuk Kwon, Zhuohan Li, Siyuan Zhuang, Ying Sheng, Lianmin Zheng, Cody~Hao Yu, Joseph~E. Gonzalez, Hao Zhang, and Ion Stoica. 2023.
\newblock Efficient memory management for large language model serving with pagedattention.
\newblock In \emph{Proceedings of the ACM SIGOPS 29th Symposium on Operating Systems Principles}.

\bibitem[{Lee et~al.(2025{\natexlab{a}})Lee, Lee, Lai, Chen, Chen, and Yau}]{medgamification2}
Ching-Yi Lee, Ching-Hsin Lee, Hung-Yi Lai, Po-Jui Chen, Mi-Mi Chen, and Sze-Yuen Yau. 2025{\natexlab{a}}.
\newblock Emerging trends in gamification for clinical reasoning education: a scoping review.
\newblock \emph{BMC Medical Education}, 25(1):435.

\bibitem[{Lee et~al.(2025{\natexlab{b}})Lee, Lee, Kim, Ko, Eun, Kim, Cho, Zhu, Kraut, Suh, Kim, and Lim}]{ACL2025Find-adaptiveVP}
Keyeun Lee, Seolhee Lee, Esther~Hehsun Kim, Yena Ko, Jinsu Eun, Dahee Kim, Hyewon Cho, Haiyi Zhu, Robert~E. Kraut, Eunyoung~E. Suh, Eun-mee Kim, and Hajin Lim. 2025{\natexlab{b}}.
\newblock \href {https://doi.org/10.18653/v1/2025.findings-acl.118} {Adaptive-{VP}: A framework for {LLM}-based virtual patients that adapts to trainees' dialogue to facilitate nurse communication training}.
\newblock In \emph{Findings of the Association for Computational Linguistics: ACL 2025}, pages 2319--2352, Vienna, Austria. Association for Computational Linguistics.

\bibitem[{Lewis et~al.(2020)Lewis, Perez, Piktus, Petroni, Karpukhin, Goyal, K{\"u}ttler, Lewis, Yih, Rockt{\"a}schel et~al.}]{RAG-nips20}
Patrick Lewis, Ethan Perez, Aleksandra Piktus, Fabio Petroni, Vladimir Karpukhin, Naman Goyal, Heinrich K{\"u}ttler, Mike Lewis, Wen-tau Yih, Tim Rockt{\"a}schel, and 1 others. 2020.
\newblock Retrieval-augmented generation for knowledge-intensive nlp tasks.
\newblock \emph{Advances in neural information processing systems}, 33:9459--9474.

\bibitem[{Li et~al.(2024)Li, Lai, Li, Ren, Zhang, Kang, Wang, Li, Zhang, Ma et~al.}]{li2024agent}
Junkai Li, Yunghwei Lai, Weitao Li, Jingyi Ren, Meng Zhang, Xinhui Kang, Siyu Wang, Peng Li, Ya-Qin Zhang, Weizhi Ma, and 1 others. 2024.
\newblock Agent hospital: A simulacrum of hospital with evolvable medical agents.
\newblock \emph{arXiv preprint arXiv:2405.02957}.

\bibitem[{Lindley(2005)}]{2005storyandNarrative}
Craig~A Lindley. 2005.
\newblock Story and narrative structures in computer games.
\newblock \emph{Bushoff, Brunhild. ed}.

\bibitem[{Lucas et~al.(2024)Lucas, Upperman, and Robinson}]{survey-mededu1}
Harrison~C Lucas, Jeffrey~S Upperman, and Jamie~R Robinson. 2024.
\newblock A systematic review of large language models and their implications in medical education.
\newblock \emph{Medical education}, 58(11):1276--1285.

\bibitem[{Ma et~al.(2026)Ma, Lin, Xu, Xie, and Yu}]{DBLP:conf/aaai/MaLXXY26}
Yingjie Ma, Xun Lin, Yong Xu, Weicheng Xie, and Zitong Yu. 2026.
\newblock {PA-FAS:} towards interpretable and generalizable multimodal face anti-spoofing via path-augmented reinforcement learning.
\newblock In \emph{{AAAI} Conference on Artificial Intelligence}, pages 7856--7864.

\bibitem[{Sellergren et~al.(2025)Sellergren, Kazemzadeh, Jaroensri, Kiraly, Traverse, Kohlberger, Xu, Jamil, Hughes, Lau et~al.}]{sellergren2025medgemma}
Andrew Sellergren, Sahar Kazemzadeh, Tiam Jaroensri, Atilla Kiraly, Madeleine Traverse, Timo Kohlberger, Shawn Xu, Fayaz Jamil, C{\'\i}an Hughes, Charles Lau, and 1 others. 2025.
\newblock Medgemma technical report.
\newblock \emph{arXiv preprint arXiv:2507.05201}.

\bibitem[{Smuha(2025)}]{smuha2025regulation}
Nathalie~A Smuha. 2025.
\newblock Regulation 2024/1689 of the eur. parl. \& council of june 13, 2024 (eu artificial intelligence act).
\newblock \emph{International Legal Materials}, 64(5):1234--1381.

\bibitem[{Staiger(1979)}]{Cinema_1}
Janet Staiger. 1979.
\newblock Dividing labor for production control: Thomas ince and the rise of the studio system.
\newblock \emph{Cinema Journal}, 18(2):16--25.

\bibitem[{Szilas(1999)}]{AAAI-1999-interactive-drama}
Nicolas Szilas. 1999.
\newblock Interactive drama on computer: beyond linear narrative.
\newblock In \emph{AAAI Fall symposium on narrative intelligence}, volume 144, pages 150--156.

\bibitem[{Team et~al.(2023)Team, Anil, Borgeaud, Alayrac, Yu, Soricut, Schalkwyk, Dai, Hauth, Millican et~al.}]{team2023gemini}
Gemini Team, Rohan Anil, Sebastian Borgeaud, Jean-Baptiste Alayrac, Jiahui Yu, Radu Soricut, Johan Schalkwyk, Andrew~M Dai, Anja Hauth, Katie Millican, and 1 others. 2023.
\newblock Gemini: a family of highly capable multimodal models.
\newblock \emph{arXiv preprint arXiv:2312.11805}.

\bibitem[{Team et~al.(2024)Team, Mesnard, Hardin, Dadashi, Bhupatiraju, Pathak, Sifre, Rivi{\`e}re, Kale, Love et~al.}]{team2024gemma}
Gemma Team, Thomas Mesnard, Cassidy Hardin, Robert Dadashi, Surya Bhupatiraju, Shreya Pathak, Laurent Sifre, Morgane Rivi{\`e}re, Mihir~Sanjay Kale, Juliette Love, and 1 others. 2024.
\newblock Gemma: Open models based on gemini research and technology.
\newblock \emph{arXiv preprint arXiv:2403.08295}.

\bibitem[{Wang et~al.(2025{\natexlab{a}})Wang, Li, Lin, Zhang, Han, Wang, Liu, Tan, Pu, Li et~al.}]{2025-JMIR-Standardized-Patient}
Chenxu Wang, Shuhan Li, Nuoxi Lin, Xinyu Zhang, Ying Han, Xiandi Wang, Di~Liu, Xiaomei Tan, Dan Pu, Kang Li, and 1 others. 2025{\natexlab{a}}.
\newblock Application of large language models in medical training evaluation—using chatgpt as a standardized patient: Multimetric assessment.
\newblock \emph{Journal of medical Internet research}, 27:e59435.

\bibitem[{Wang et~al.(2026)Wang, Shi, Tao, Gao, Zhang, Lin, Feng, Yuan, Yu, and Cao}]{DBLP:conf/aaai/WangSTGZLFYYC26}
Hongyang Wang, Yichen Shi, Zhuofu Tao, Yuhao Gao, Liepiao Zhang, Xun Lin, Jun Feng, Xiaochen Yuan, Zitong Yu, and Xiaochun Cao. 2026.
\newblock {FaceShield:} explainable face anti-spoofing with multimodal large language models.
\newblock In \emph{{AAAI} Conference on Artificial Intelligence}, pages 9811--9819.

\bibitem[{Wang et~al.(2024{\natexlab{a}})Wang, Milani, Chiu, Zhi, Eack, Labrum, Murphy, Jones, Hardy, Shen, Fang, and Chen}]{EMNLP-patient-psi}
Ruiyi Wang, Stephanie Milani, Jamie~C. Chiu, Jiayin Zhi, Shaun~M. Eack, Travis Labrum, Samuel~M Murphy, Nev Jones, Kate~V Hardy, Hong Shen, Fei Fang, and Zhiyu Chen. 2024{\natexlab{a}}.
\newblock \href {https://doi.org/10.18653/v1/2024.emnlp-main.711} {{PATIENT}-$\psi$: Using large language models to simulate patients for training mental health professionals}.
\newblock In \emph{Proceedings of the 2024 Conference on Empirical Methods in Natural Language Processing}, pages 12772--12797, Miami, Florida, USA. Association for Computational Linguistics.

\bibitem[{Wang et~al.(2024{\natexlab{b}})Wang, Hsu, Fang, and Kuo}]{medgamification1}
Yung-Fu Wang, Ya-Fang Hsu, Kwo-Ting Fang, and Liang-Tseng Kuo. 2024{\natexlab{b}}.
\newblock Gamification in medical education: identifying and prioritizing key elements through delphi method.
\newblock \emph{Medical Education Online}, 29(1):2302231.

\bibitem[{Wang et~al.(2024{\natexlab{c}})Wang, Zhou, and Ledo}]{wang2024storyverse}
Z.~Wang, Y.~Zhou, and D.~Ledo. 2024{\natexlab{c}}.
\newblock \href {https://doi.org/10.1145/3613904.3642436} {{S}tory{V}erse: Towards co-authoring dynamic plot with {LLM}-based character simulation via narrative planning}.
\newblock In \emph{Proceedings of the CHI Conference on Human Factors in Computing Systems (CHI '24)}. ACM.

\bibitem[{Wang et~al.(2025{\natexlab{b}})}]{wang2025codi}
Z.~Wang and 1 others. 2025{\natexlab{b}}.
\newblock {C}o{D}i: A director-actor framework for goal-driven interactive story generation with {LLM}s.
\newblock In \emph{Proceedings of the AAAI Conference on Artificial Intelligence and Interactive Digital Entertainment (AIIDE)}.

\bibitem[{Wu et~al.(2025{\natexlab{a}})Wu, Wu, Xu, Zhang, and Zhao}]{wu-etal-2025-towards-enhanced}
Hongqiu Wu, Weiqi Wu, Tianyang Xu, Jiameng Zhang, and Hai Zhao. 2025{\natexlab{a}}.
\newblock \href {https://doi.org/10.18653/v1/2025.acl-long.546} {Towards enhanced immersion and agency for {LLM}-based interactive drama}.
\newblock In \emph{Proceedings of the 63rd Annual Meeting of the Association for Computational Linguistics (Volume 1: Long Papers)}, pages 11166--11182, Vienna, Austria. Association for Computational Linguistics.

\bibitem[{Wu et~al.(2025{\natexlab{b}})Wu, Gao, Gou, and Dou}]{DDxTutor-ACL25}
Qian Wu, Zheyao Gao, Longfei Gou, and Qi~Dou. 2025{\natexlab{b}}.
\newblock {DD}x{T}utor: Clinical reasoning tutoring system with differential diagnosis-based structured reasoning.
\newblock In \emph{Proceedings of the 63rd Annual Meeting of the Association for Computational Linguistics (Volume 1: Long Papers)}, pages 30934--30957.

\bibitem[{Wu et~al.(2025{\natexlab{c}})Wu, Gao, Gou, Hou, Lau, and Dou}]{HealthCards-EMNLP25}
Qian Wu, Zheyao Gao, Longfei Gou, Yifan Hou, Ann Sin~Nga Lau, and Qi~Dou. 2025{\natexlab{c}}.
\newblock {H}ealth{C}ards: Exploring text-to-image generation as visual aids for healthcare knowledge democratizing and education.
\newblock In \emph{Proceedings of the 2025 Conference on Empirical Methods in Natural Language Processing}, pages 27524--27546.

\bibitem[{Wu et~al.(2024)Wu, Wu, Jiang, Liu, Zhao, and Zhang}]{wu-etal-2024-role}
Weiqi Wu, Hongqiu Wu, Lai Jiang, Xingyuan Liu, Hai Zhao, and Min Zhang. 2024.
\newblock \href {https://doi.org/10.18653/v1/2024.findings-acl.196} {From role-play to drama-interaction: An {LLM} solution}.
\newblock In \emph{Findings of the Association for Computational Linguistics: ACL 2024}, pages 3271--3290, Bangkok, Thailand. Association for Computational Linguistics.

\bibitem[{Yang et~al.(2025{\natexlab{a}})Yang, Li, Yang, Zhang, Hui, Zheng, Yu, Gao, Huang, Lv et~al.}]{yang2025qwen3}
An~Yang, Anfeng Li, Baosong Yang, Beichen Zhang, Binyuan Hui, Bo~Zheng, Bowen Yu, Chang Gao, Chengen Huang, Chenxu Lv, and 1 others. 2025{\natexlab{a}}.
\newblock Qwen3 technical report.
\newblock \emph{arXiv preprint arXiv:2505.09388}.

\bibitem[{Yang et~al.(2025{\natexlab{b}})Yang, Kong, Gao, Cheng, Liu, Zhang, Kang, Luo, Cai, He, and Wei}]{yang2025infinitetalk}
Shaoshu Yang, Zhe Kong, Feng Gao, Meng Cheng, Xiangyu Liu, Yong Zhang, Zhuoliang Kang, Wenhan Luo, Xunliang Cai, Ran He, and Xiaoming Wei. 2025{\natexlab{b}}.
\newblock Infinitetalk: Audio-driven video generation for sparse-frame video dubbing.
\newblock \emph{arXiv preprint arXiv:2508.14033}.

\bibitem[{Yao et~al.(2024)Yao, Kantu, Wei, Tran, Duan, Kwon, Yang, and Yu}]{yao-etal-2024-readme}
Zonghai Yao, Nandyala~Siddharth Kantu, Guanghao Wei, Hieu Tran, Zhangqi Duan, Sunjae Kwon, Zhichao Yang, and Hong Yu. 2024.
\newblock \href {https://doi.org/10.18653/v1/2024.findings-emnlp.737} {{README}: Bridging medical jargon and lay understanding for patient education through data-centric {NLP}}.
\newblock In \emph{Findings of the Association for Computational Linguistics: EMNLP 2024}, pages 12609--12629, Miami, Florida, USA. Association for Computational Linguistics.

\bibitem[{Yao et~al.(2025)Yao, Sun, Jang, Kwon, Kwon, and Yu}]{yao-etal-2025-dischargesim}
Zonghai Yao, Michael Sun, Won~Seok Jang, Sunjae Kwon, Soie Kwon, and Hong Yu. 2025.
\newblock \href {https://doi.org/10.18653/v1/2025.emnlp-main.546} {{D}ischarge{S}im: A simulation benchmark for educational doctor{--}patient communication at discharge}.
\newblock In \emph{Proceedings of the 2025 Conference on Empirical Methods in Natural Language Processing}, pages 10783--10809, Suzhou, China. Association for Computational Linguistics.

\bibitem[{Yao et~al.(2026)Yao, Tang, Lin, Luo, Wang, Huang, Ong, and Yu}]{yao2026rethinking}
Zonghai Yao, Zhipeng Tang, Chengtao Lin, Xiong Luo, Benlu Wang, Juncheng Huang, Chin~Siang Ong, and Hong Yu. 2026.
\newblock Rethinking patient education as multi-turn multi-modal interaction.
\newblock \emph{arXiv preprint arXiv:2604.14656}.

\bibitem[{Zeybek and Sayg{\i}(2024)}]{gamification2}
Nil{\"u}fer Zeybek and Elif Sayg{\i}. 2024.
\newblock Gamification in education: Why, where, when, and how?—a systematic review.
\newblock \emph{Games and Culture}, 19(2):237--264.

\bibitem[{Zhang et~al.(2026)Zhang, Liu, Wang, Lei, Xie, and Wang}]{zhang-etal-2026-human}
Bingquan Zhang, Xiaoxiao Liu, Yuchi Wang, Zhou Lei, Qianqian Xie, and Benyou Wang. 2026.
\newblock \href {https://doi.org/10.18653/v1/2026.acl-long.1243} {Human or {LLM} as standardized patients? a comparative study in medical education}.
\newblock In \emph{Proceedings of the 64th Annual Meeting of the {A}ssociation for {C}omputational {L}inguistics (Volume 1: Long Papers)}, pages 26988--27012, San Diego, California, United States. Association for Computational Linguistics.

\bibitem[{Zhao et~al.(2023)Zhao, Jin, Chen, Peng, and Yu}]{PMC_Patient}
Zhengyun Zhao, Qiao Jin, Fangyuan Chen, Tuorui Peng, and Sheng Yu. 2023.
\newblock \href {https://api.semanticscholar.org/CorpusID:266360591} {A large-scale dataset of patient summaries for retrieval-based clinical decision support systems.}
\newblock \emph{Scientific data}, 10 1:909.

\bibitem[{Zheng et~al.(2023)Zheng, Chiang, Sheng, Zhuang, Wu, Zhuang, Lin, Li, Li, Xing et~al.}]{llm-as-a-judge}
Lianmin Zheng, Wei-Lin Chiang, Ying Sheng, Siyuan Zhuang, Zhanghao Wu, Yonghao Zhuang, Zi~Lin, Zhuohan Li, Dacheng Li, Eric Xing, and 1 others. 2023.
\newblock Judging llm-as-a-judge with mt-bench and chatbot arena.
\newblock \emph{Advances in neural information processing systems}, 36:46595--46623.

\end{thebibliography}
\appendix
\startcontents[appendices]

\titlecontents{section}[0pt]{\bfseries}{\thecontentslabel\quad}{}{\titlerule*[1pc]{.}\contentspage}

\titlecontents{subsection}[2em]{}{\thecontentslabel\quad}{}{\titlerule*[1pc]{.}\contentspage}

\newpage
\section*{\centering Appendix Overview}

\printcontents[appendices]{}{1}{\setcounter{tocdepth}{2}}

\section{Game Design Logic}
\label{sec:appendix-game}

\subsection{Clinical Story Structure}
\label{appendix section: clinical story structure}

\begin{figure}[htbp]
    \centering
    \begin{tcolorbox}[
        colback=white,
        colframe=gray!50,
        arc=4mm,
        boxrule=0.5pt,
        left=2mm,
        right=2mm,
        top=2mm,
        bottom=2mm
    ]
    \includegraphics[width=\linewidth]{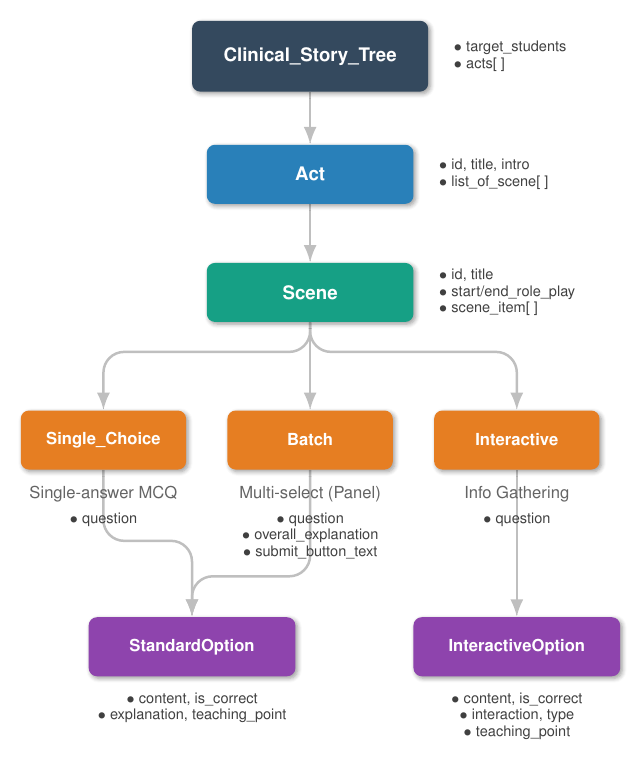}
    \end{tcolorbox}
    \caption{Hierarchical structure of the Clinical Story. The tree consists of five levels: the root \texttt{Clinical\_Story\_Tree} contains multiple \texttt{Act}s (chapters), each \texttt{Act} contains multiple \texttt{Scene}s (stages), each \texttt{Scene} contains \texttt{scene\_item}s (interaction elements), and each item contains \texttt{Option}s for user interaction.}
    \label{fig:clinical_story_hierarchy}
\end{figure}

The Clinical Story serves as the foundational data structure for our interactive medical education game. It adopts a hierarchical ``chapter-level'' design that organizes clinical narratives into a structured tree, enabling systematic progression through medical scenarios. Figure~\ref{fig:clinical_story_hierarchy} illustrates the overall hierarchy of this structure.

\paragraph{Level 0: Clinical\_Story\_Tree.} The root node defines the overall narrative and specifies the \texttt{target\_students} field (e.g., medical students, residents, or physicians) to calibrate the difficulty and depth of clinical content.

\paragraph{Level 1: Act.} An \texttt{Act} represents a major chapter or phase of the clinical encounter (e.g., ``Initial Assessment'', ``Diagnosis'', ``Treatment Planning''). Each act includes an \texttt{intro} field that provides contextual background and sets the scene before the player enters.

\paragraph{Level 2: Scene.} A \texttt{Scene} is a specific stage within an act, representing a discrete clinical decision point or interaction moment. Each scene contains: (1) \texttt{start\_role\_play}---narrative text that sets up the context before the interaction begins; (2) \texttt{scene\_item}---a list of interaction elements the player must complete; and (3) \texttt{end\_role\_play}---transitional narrative that concludes the scene and bridges to the next. The role-play fields support three presentation types for downstream multimodal rendering: \textbf{text} (pure text display), \textbf{film} (cinematic video generation), and \textbf{interleaved} (mixed text and dialogue sequences).

\paragraph{Level 3: Scene Items.} Each \texttt{scene\_item} corresponds to the Decision Node $\mathcal{Q}$ defined in Section~\ref{subsection: Hierarchy of Acts, Scenes, and Nodes}, representing an interaction mode. We define three types: (1) \textbf{single\_choice}---a single-answer MCQ for discrete clinical decision points; (2) \textbf{batch}---a multi-select question where the player chooses multiple correct items simultaneously (e.g., ordering a panel of diagnostic tests), with an \texttt{overall\_explanation} field for holistic feedback; and (3) \textbf{interactive}---an information-gathering interaction where each option reveals different information (patient responses, examination findings), simulating clinical inquiry.

\paragraph{Level 4: Options.} Options are the atomic interaction units. We define two types: (1) \textbf{StandardOption} (for \texttt{single\_choice} and \texttt{batch}) contains \texttt{content}, \texttt{is\_correct}, \texttt{explanation} (tactical feedback), and \texttt{teaching\_point} (knowledge summary); (2) \textbf{InteractiveOption} (for \texttt{interactive}) contains \texttt{content}, \texttt{is\_correct}, \texttt{interaction} (the returned information), \texttt{interaction\_type} (\texttt{dialogue} or \texttt{text}), and \texttt{teaching\_point}.

The complete Pydantic schema for this structure is provided in Listing~\ref{list: pydantic_models of clinical story}.

\begin{listing}[htbp]
    \centering
    \begin{tcolorbox}[
        colback=gray!5!white,
        colframe=gray!50,
        arc=4mm,
        boxrule=0.5pt,
        left=0mm,
        right=0mm,
        top=0mm,
        bottom=0mm
    ]
\lstset{
    language=Python,
    basicstyle=\ttfamily\scriptsize,
    numbers=left,
    numbersep=5pt,
    numberstyle=\tiny\color{gray},
    frame=none,
    framesep=2mm,
    breaklines=true,
    breakatwhitespace=true,
    tabsize=2,
    showstringspaces=false,
    columns=fullflexible,
    keepspaces=true
}
\begin{lstlisting}
from typing import List, Literal, Optional, Union
from pydantic import BaseModel, Field

# ========================================================
# 1. Define two different types of "Option" models
# ========================================================

class StandardOption(BaseModel):
    """
    Option used for standard decision points.
    """
    content: str = Field(description="The text content")
    is_correct: bool = Field(description="Is correct answer")
    explanation: str = Field(description="Tactical feedback")
    teaching_point: str = Field(description="Teaching point")

class InteractiveOption(BaseModel):
    """
    Option used for interactive information gathering.
    """
    content: str = Field(description="Text content")
    is_correct: bool = Field(description="Usually True")
    interaction: str = Field(description="Result text")
    interaction_type: Literal["dialogue", "text"]
    teaching_point: str = Field(description="Teaching point")

# ========================================================
# 2. Define "Scene Item" models
# ========================================================

class MultiChoice_SingleAnswer(BaseModel):
    interaction_type: Literal["single_choice"]
    question: str = Field(description="Question text")
    options: List[StandardOption]

class BatchQuestion(BaseModel):
    interaction_type: Literal["batch"]
    question: str = Field(description="Question text")
    options: List[StandardOption]
    submit_button_text: Optional[str] = Field(None)
    overall_explanation: str = Field(description="Holistic logic")

class InteractiveQuestion(BaseModel):
    interaction_type: Literal["interactive"]
    question: str = Field(description="Prompt text")
    options: List[InteractiveOption]

# ========================================================
# 3. Combine into Scene, Act, and Story Tree
# ========================================================

class Scene(BaseModel):
    id: int
    title: str
    start_role_play: str
    scene_item: List[Union[MultiChoice_SingleAnswer,
                           BatchQuestion,
                           InteractiveQuestion]]
    end_role_play: str

class Act(BaseModel):
    id: int
    title: str
    intro: str
    list_of_scene: List[Scene]

class Clinical_Story_Tree(BaseModel):
    target_students: str
    acts: List[Act]
\end{lstlisting}
    \end{tcolorbox}
    \caption{Pydantic data models for the Clinical Storyline structure.}
    \label{list: pydantic_models of clinical story}
\end{listing}

\subsection{Details of Narrative Elements}
\label{appendix section: Details of Narrative Elements}

As introduced in Section~\ref{section: the MedGame Framework}, Narrative Elements $\{\mathcal{P}, \mathcal{L}\}$ consist of \textbf{Medical Staff Personas} ($\mathcal{P}$) and \textbf{Clinical Locations} ($\mathcal{L}$). These pre-defined assets provide a consistent clinical setting across all generated stories, ensuring visual coherence and reducing the complexity of downstream multimodal rendering. We detail each element below.

\subsubsection{Clinical Locations ($\mathcal{L}$)}

We define six clinical locations that cover the essential environments encountered in hospital-based medical practice. Each location is rendered as a high-quality background image with designated character placement positions.

\begin{figure*}[htbp]
\centering
\begin{tcolorbox}[
    colback=white,
    colframe=gray!50,
    arc=4mm,
    boxrule=0.5pt,
    left=3mm,
    right=3mm,
    top=3mm,
    bottom=3mm
]

\noindent
\begin{minipage}[c]{0.22\textwidth}
    \includegraphics[width=\linewidth]{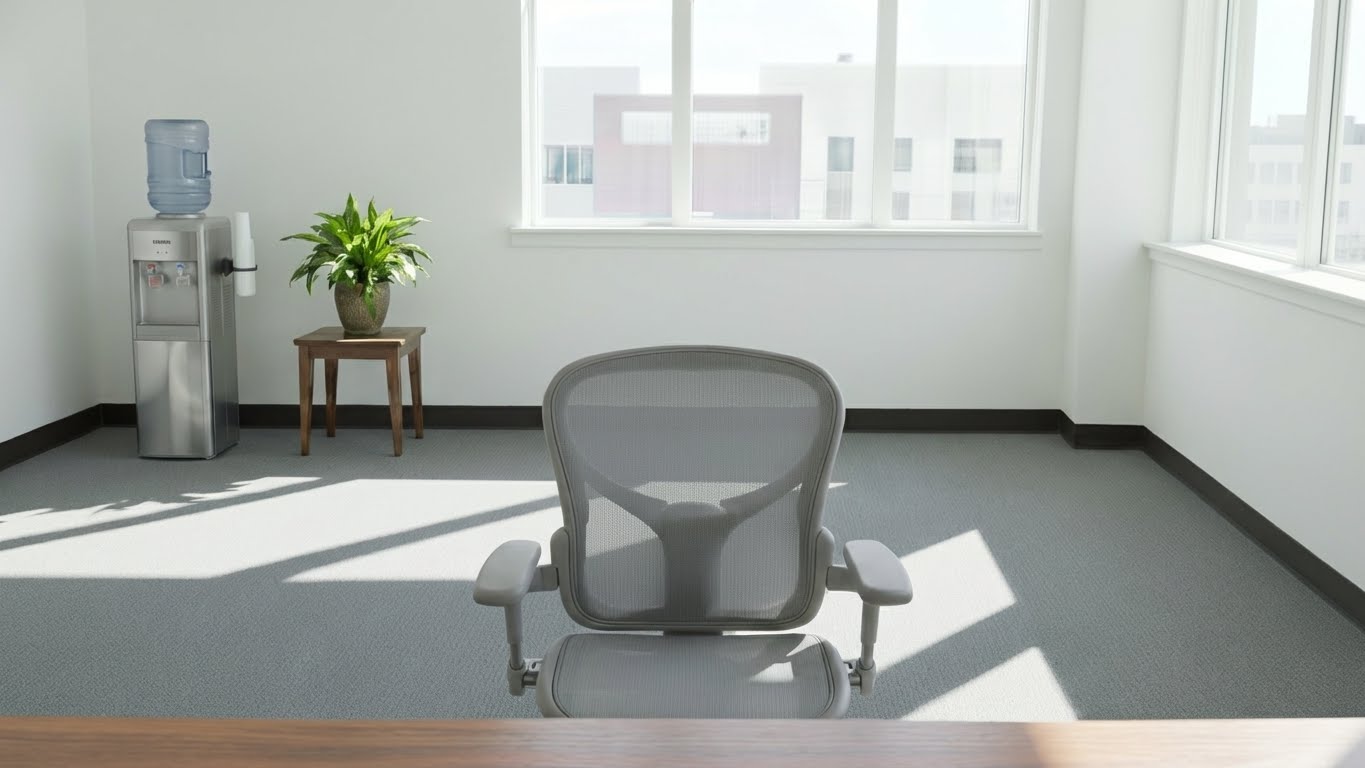}
\end{minipage}%
\hfill
\begin{minipage}[c]{0.22\textwidth}
    \includegraphics[width=\linewidth]{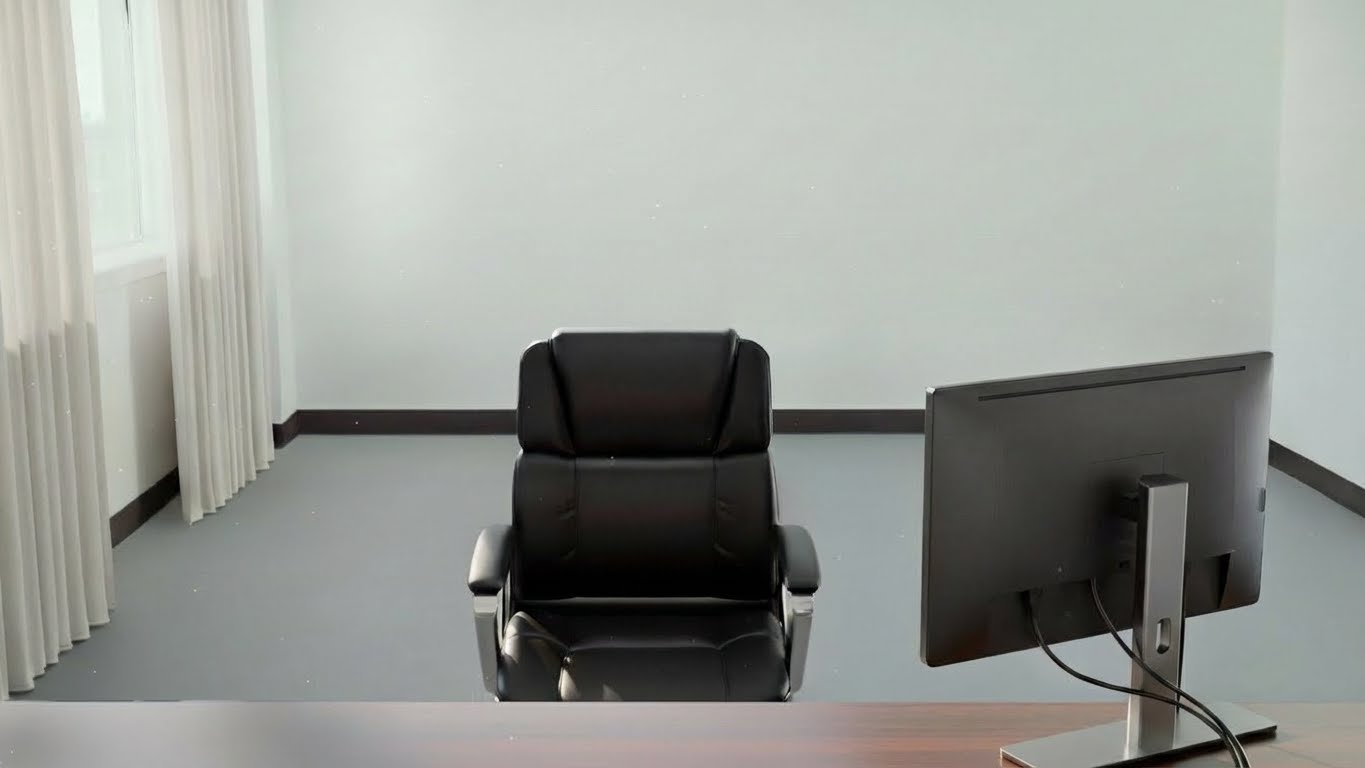}
\end{minipage}%
\hfill
\begin{minipage}[c]{0.52\textwidth}
    \small\textbf{(a) Consultation Room}
    
    {\scriptsize\textcolor{RoyalBlue}{\textit{Visual:}} A minimalist, bright, and professional office environment featuring a dark wooden desk, modern office chair, and guest seating with soft natural lighting.
    
    \textcolor{ForestGreen}{\textit{Narrative:}} The Doctor's (Player's) main office and central hub for diagnosis. Primary setting for medical history taking, physical examinations, explaining conditions, and reviewing reports.}
\end{minipage}

\smallskip
\noindent\makebox[\linewidth]{\textcolor{gray!40}{\rule{0.95\linewidth}{0.3pt}}}
\smallskip

\noindent
\begin{minipage}[c]{0.22\textwidth}
    \includegraphics[width=\linewidth]{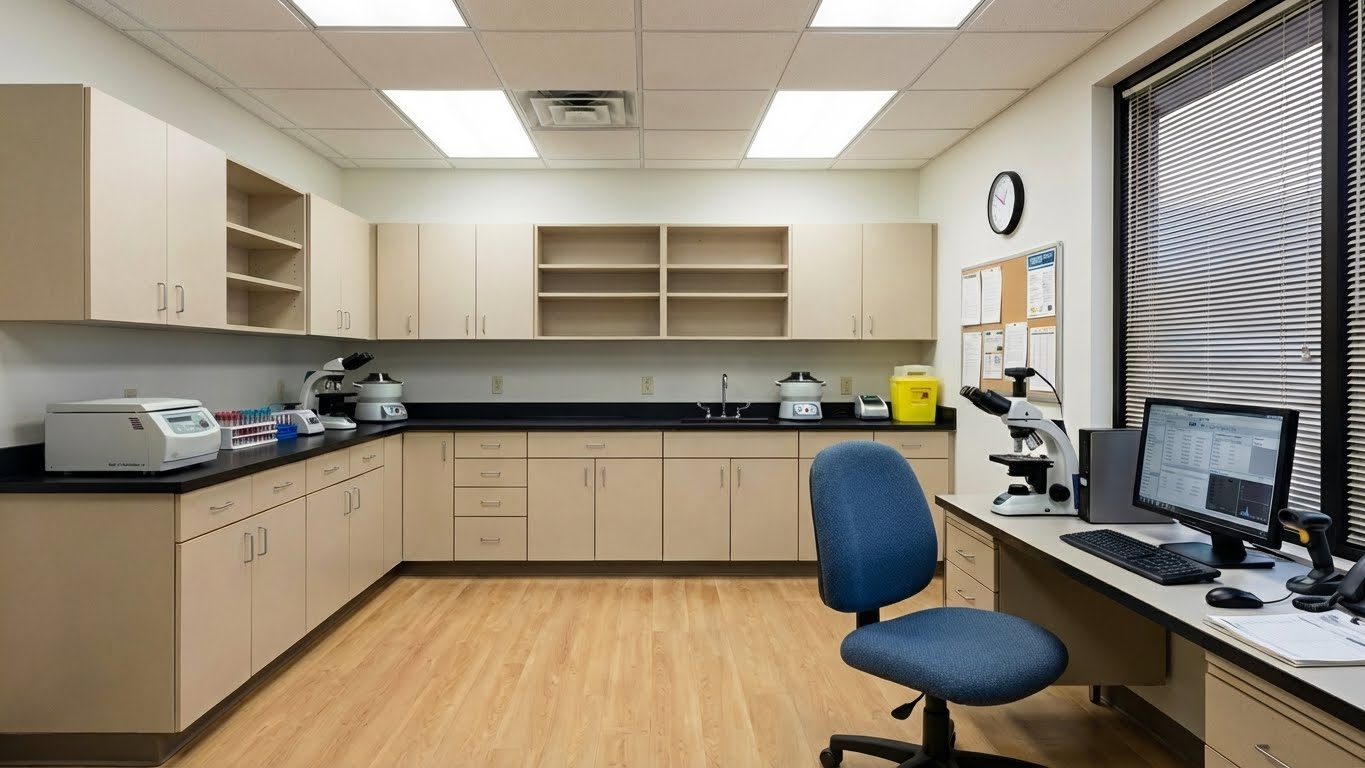}
\end{minipage}%
\hfill
\begin{minipage}[c]{0.22\textwidth}
    \includegraphics[width=\linewidth]{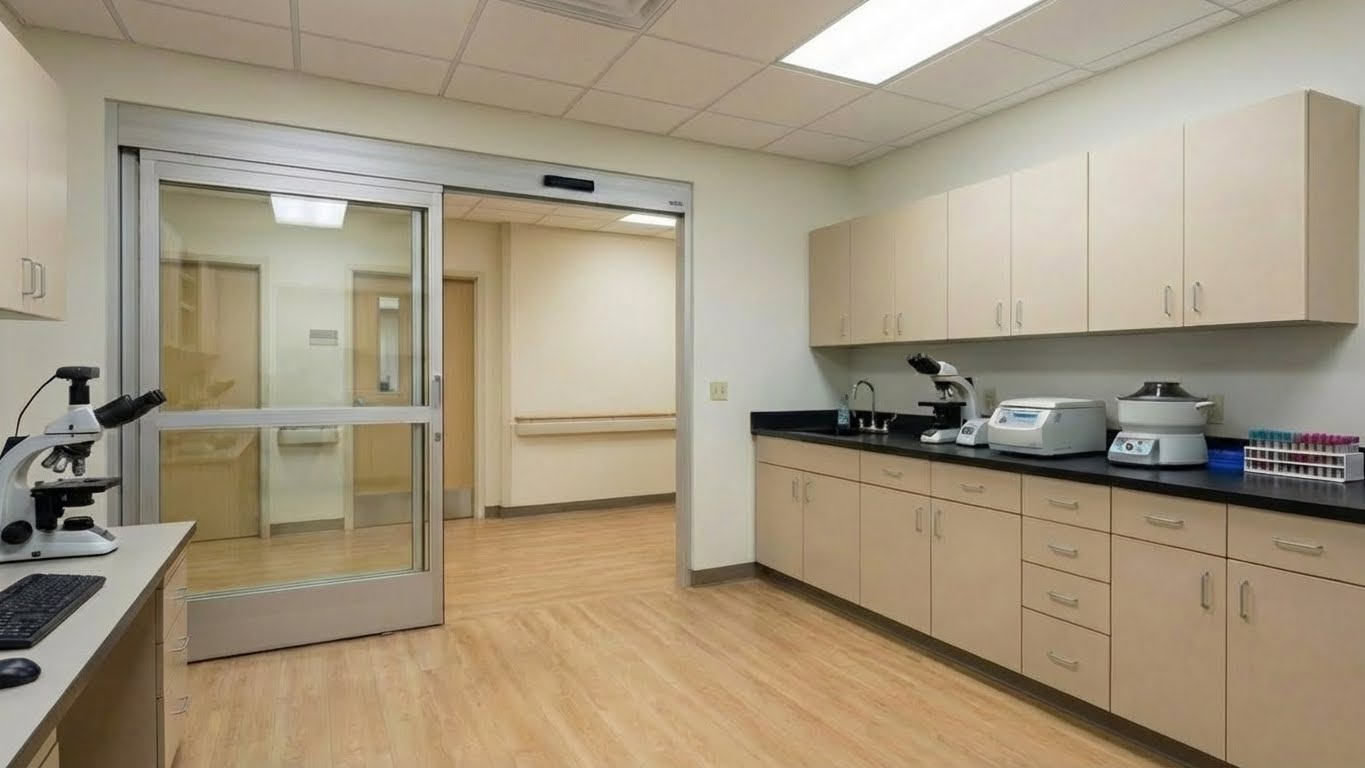}
\end{minipage}%
\hfill
\begin{minipage}[c]{0.52\textwidth}
    \small\textbf{(b) Clinical Laboratory}
    
    {\scriptsize\textcolor{RoyalBlue}{\textit{Visual:}} A spacious functional laboratory equipped with medical cabinets, microscopes, test tubes, and analysis instruments, with a workstation for the clinical pathologist.
    
    \textcolor{ForestGreen}{\textit{Narrative:}} Specialized facility for analyzing blood and body fluids. Where the Doctor (Player) discusses laboratory results (blood smears, biochemistry) with the Lab Physician.}
\end{minipage}

\smallskip
\noindent\makebox[\linewidth]{\textcolor{gray!40}{\rule{0.95\linewidth}{0.3pt}}}
\smallskip

\noindent
\begin{minipage}[c]{0.22\textwidth}
    \includegraphics[width=\linewidth]{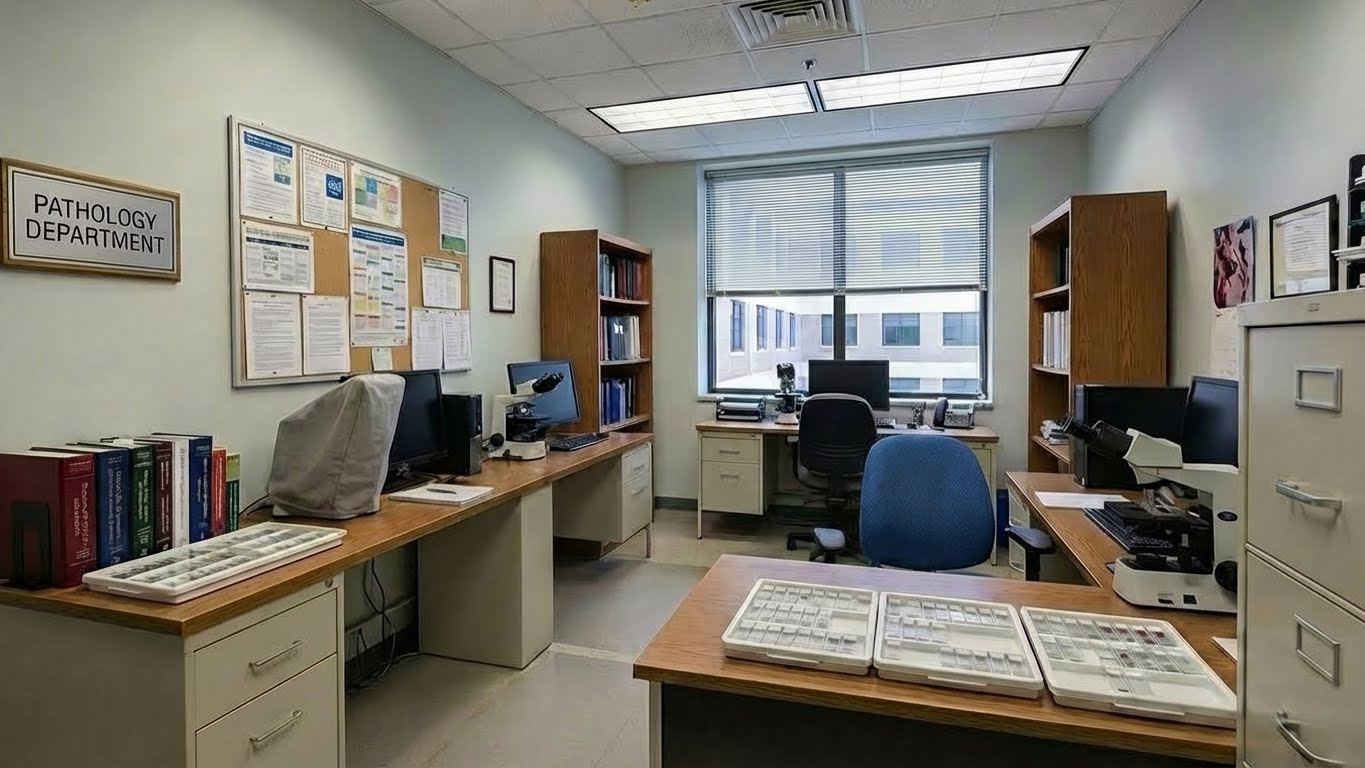}
\end{minipage}%
\hfill
\begin{minipage}[c]{0.22\textwidth}
    \includegraphics[width=\linewidth]{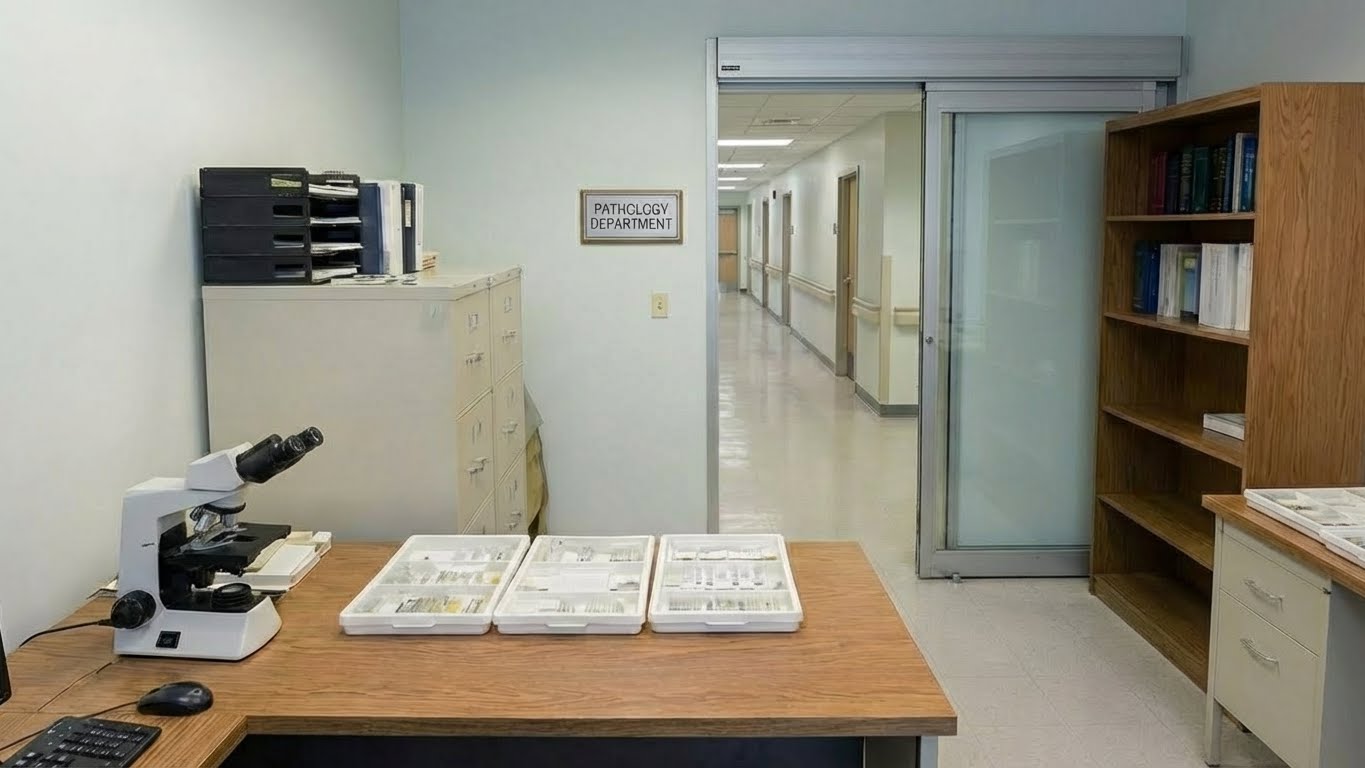}
\end{minipage}%
\hfill
\begin{minipage}[c]{0.52\textwidth}
    \small\textbf{(c) Pathology Department Office}
    
    {\scriptsize\textcolor{RoyalBlue}{\textit{Visual:}} A specialized office space focused on microscopic analysis, with a desk featuring a microscope, specimen slides, and connection to the hospital corridor via glass sliding door.
    
    \textcolor{ForestGreen}{\textit{Narrative:}} Dedicated to tissue sample analysis. Where the Anatomic Pathologist discusses biopsy results or specimen findings with the attending Doctor (Player).}
\end{minipage}

\smallskip
\noindent\makebox[\linewidth]{\textcolor{gray!40}{\rule{0.95\linewidth}{0.3pt}}}
\smallskip

\noindent
\begin{minipage}[c]{0.22\textwidth}
    \includegraphics[width=\linewidth]{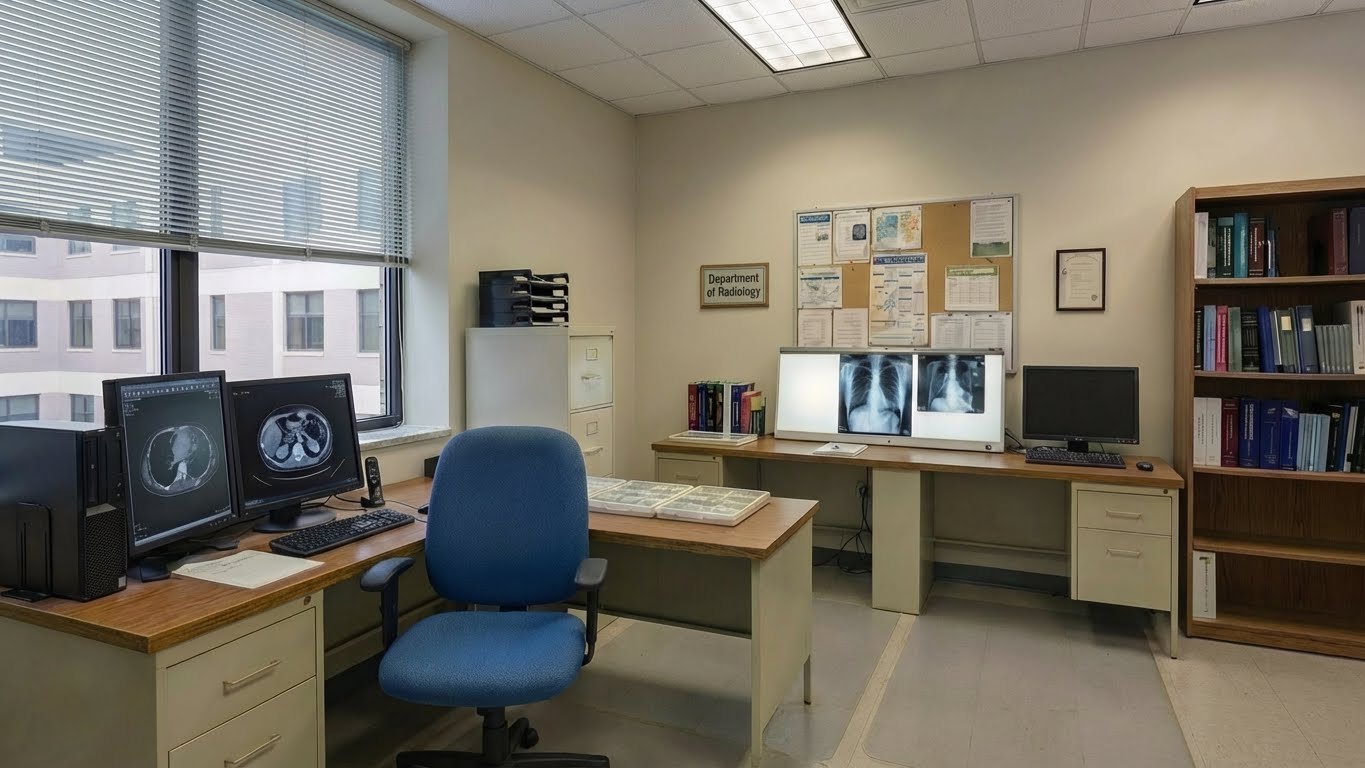}
\end{minipage}%
\hfill
\begin{minipage}[c]{0.22\textwidth}
    \includegraphics[width=\linewidth]{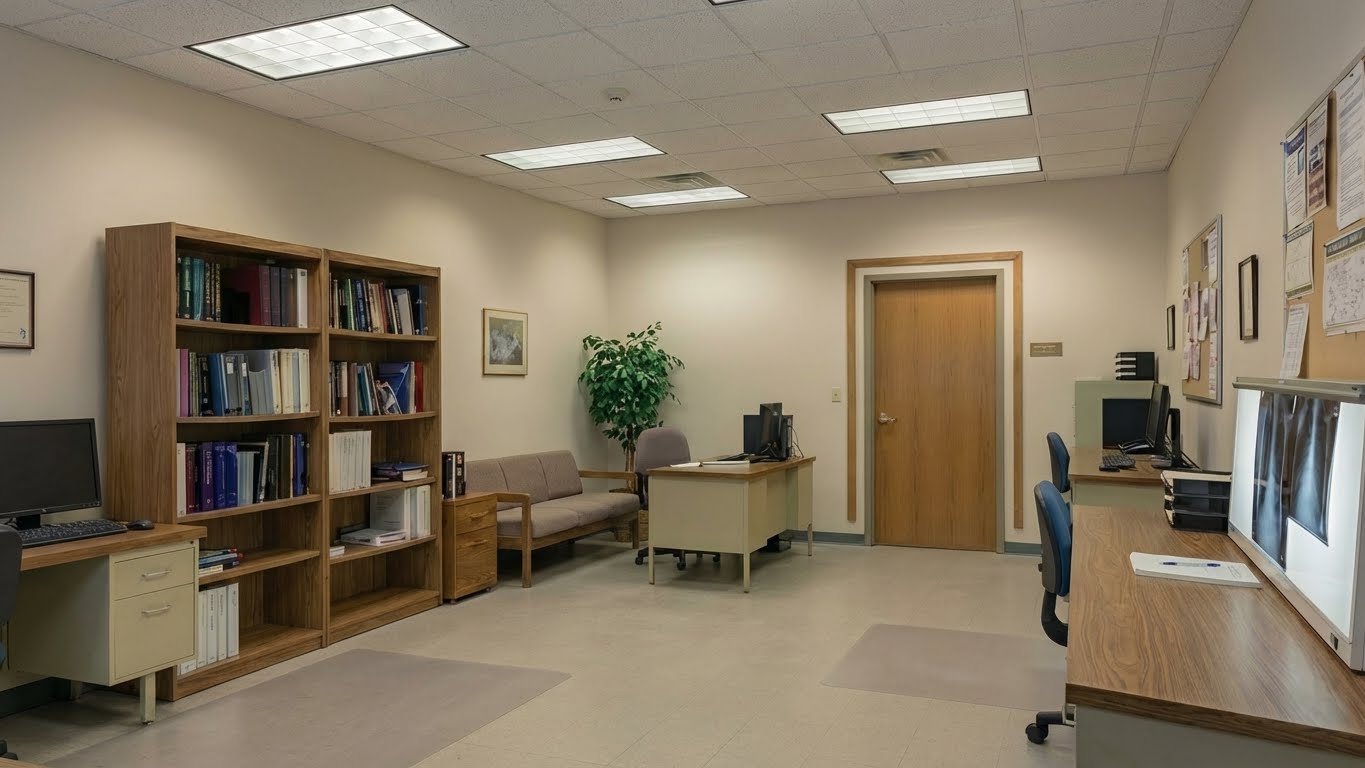}
\end{minipage}%
\hfill
\begin{minipage}[c]{0.52\textwidth}
    \small\textbf{(d) Radiology Department Office}
    
    {\scriptsize\textcolor{RoyalBlue}{\textit{Visual:}} A technical workspace for imaging analysis featuring light boxes for X-rays, dual-monitor workstations for CT scans, and bookshelves for reference materials.
    
    \textcolor{ForestGreen}{\textit{Narrative:}} Used for reviewing medical imaging data. The Radiologist analyzes X-rays and CT scans and explains imaging results to the treating Doctor (Player).}
\end{minipage}

\smallskip
\noindent\makebox[\linewidth]{\textcolor{gray!40}{\rule{0.95\linewidth}{0.3pt}}}
\smallskip

\noindent
\begin{minipage}[c]{0.22\textwidth}
    \includegraphics[width=\linewidth]{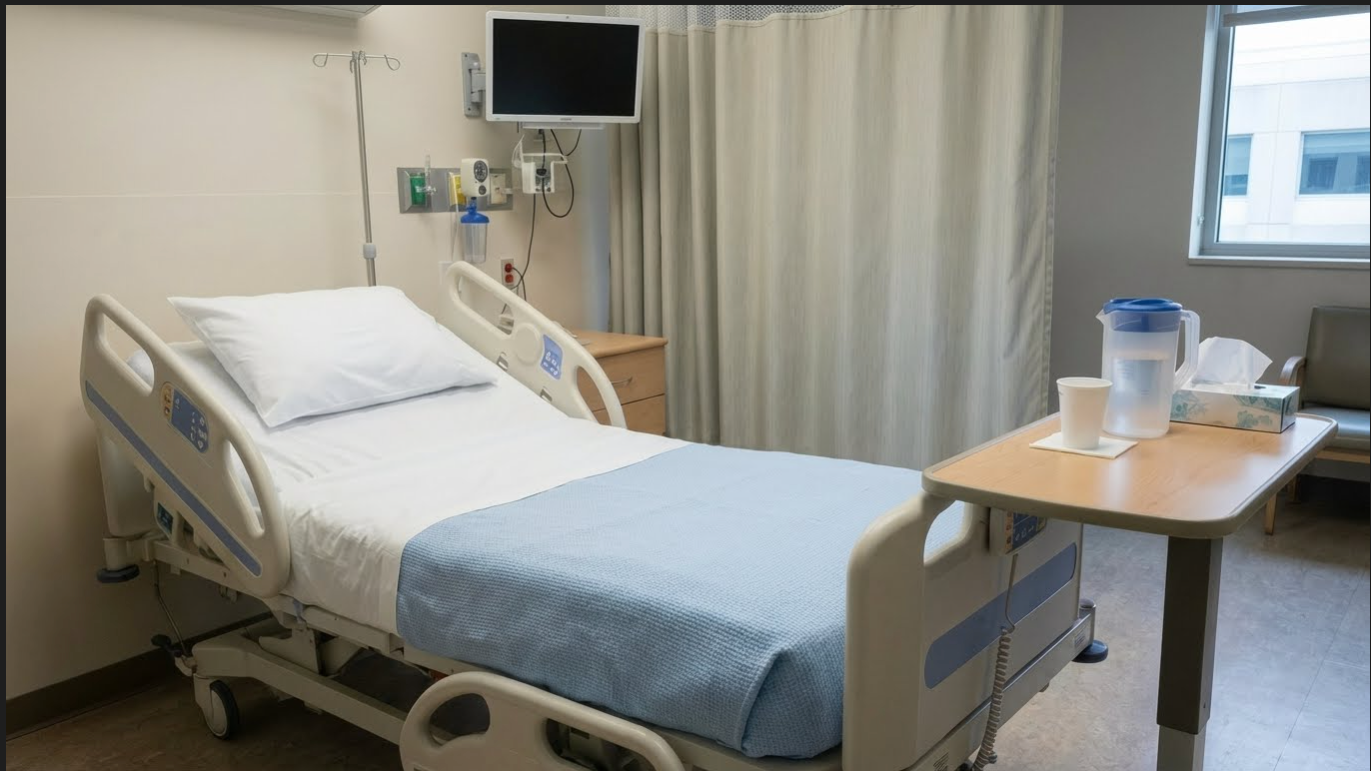}
\end{minipage}%
\hfill
\begin{minipage}[c]{0.22\textwidth}
    \includegraphics[width=\linewidth]{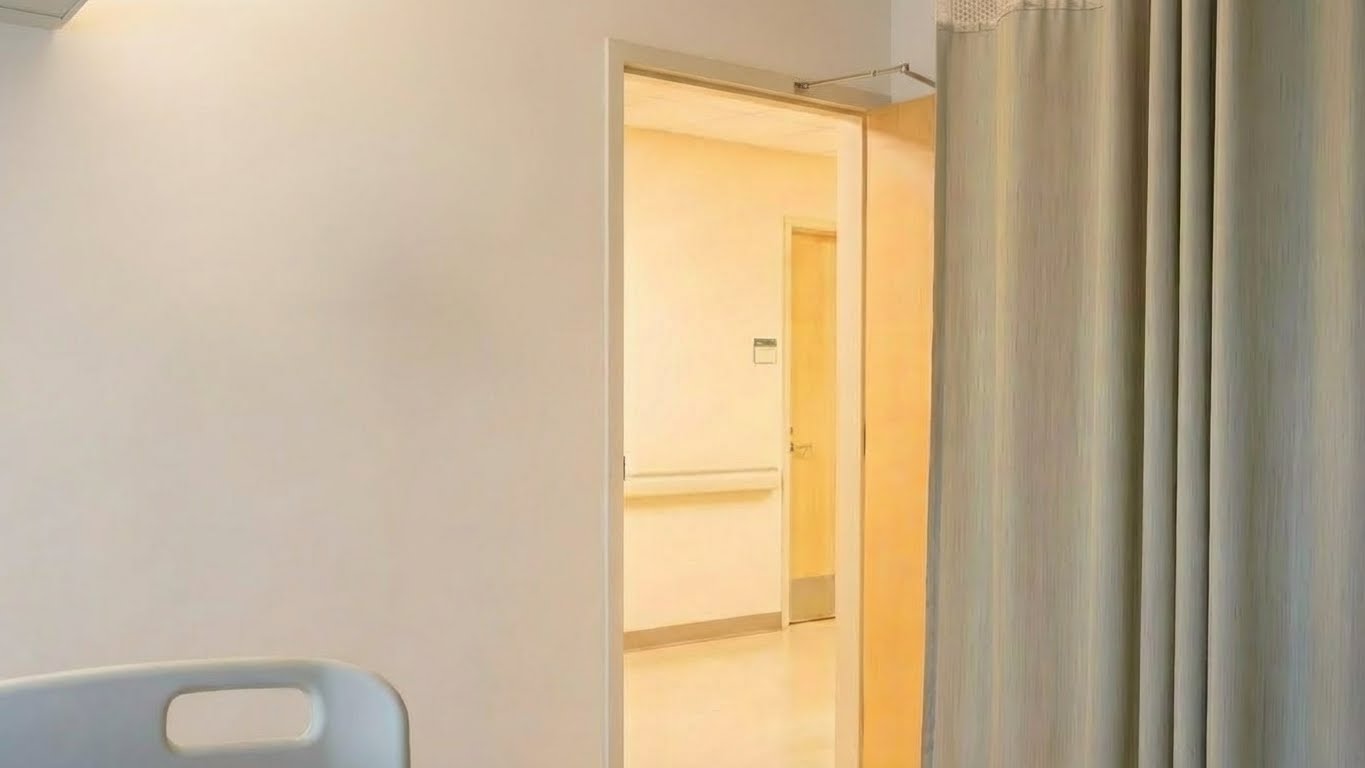}
\end{minipage}%
\hfill
\begin{minipage}[c]{0.52\textwidth}
    \small\textbf{(e) Hospital General Ward}
    
    {\scriptsize\textcolor{RoyalBlue}{\textit{Visual:}} A clean, well-lit standard inpatient room with a hospital bed, blue bedding, overbed table, privacy curtains, and guest chair. Large windows provide natural light.
    
    \textcolor{ForestGreen}{\textit{Narrative:}} Standard setting for patient recovery and monitoring. Used for bedside rounds, nurse status reports, and routine communication between Doctor (Player) and patient.}
\end{minipage}

\smallskip
\noindent\makebox[\linewidth]{\textcolor{gray!40}{\rule{0.95\linewidth}{0.3pt}}}
\smallskip

\noindent
\begin{minipage}[c]{0.22\textwidth}
    \includegraphics[width=\linewidth]{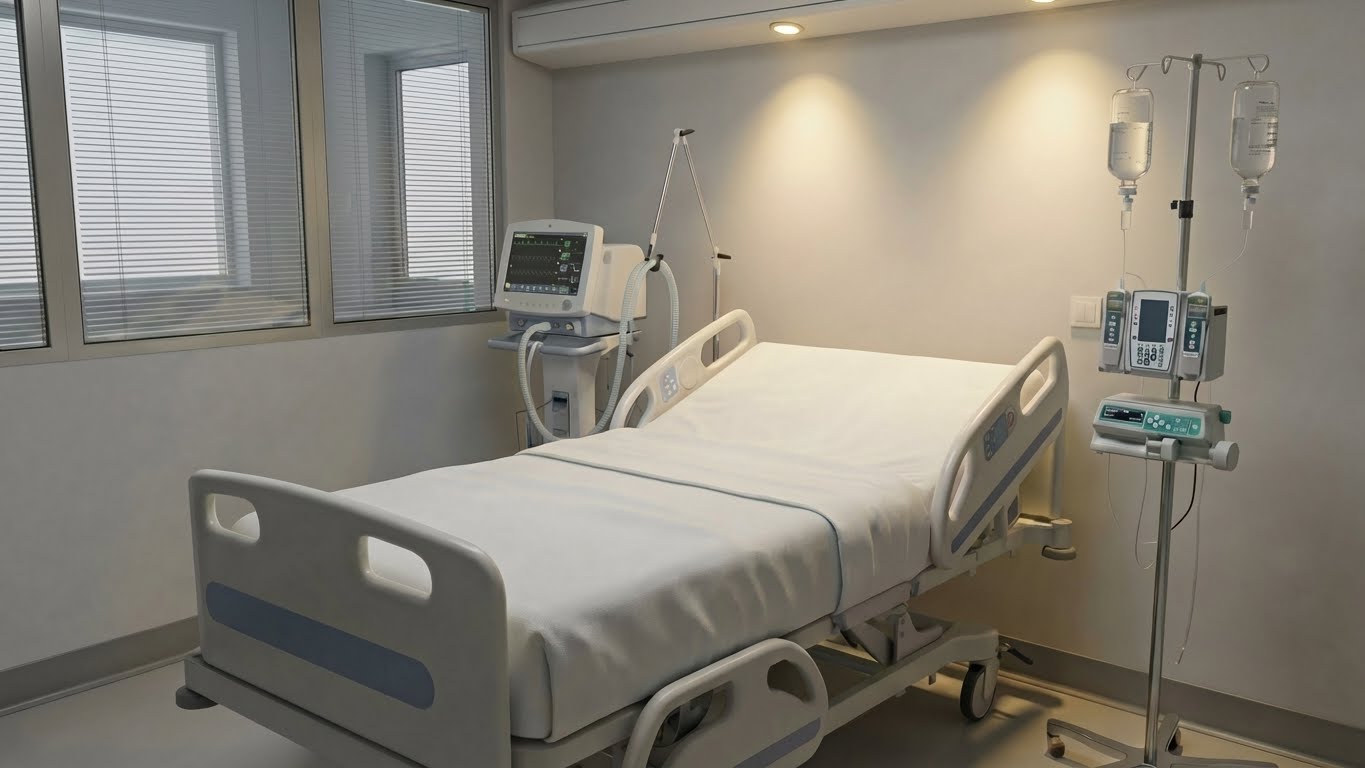}
\end{minipage}%
\hfill
\begin{minipage}[c]{0.22\textwidth}
    \includegraphics[width=\linewidth]{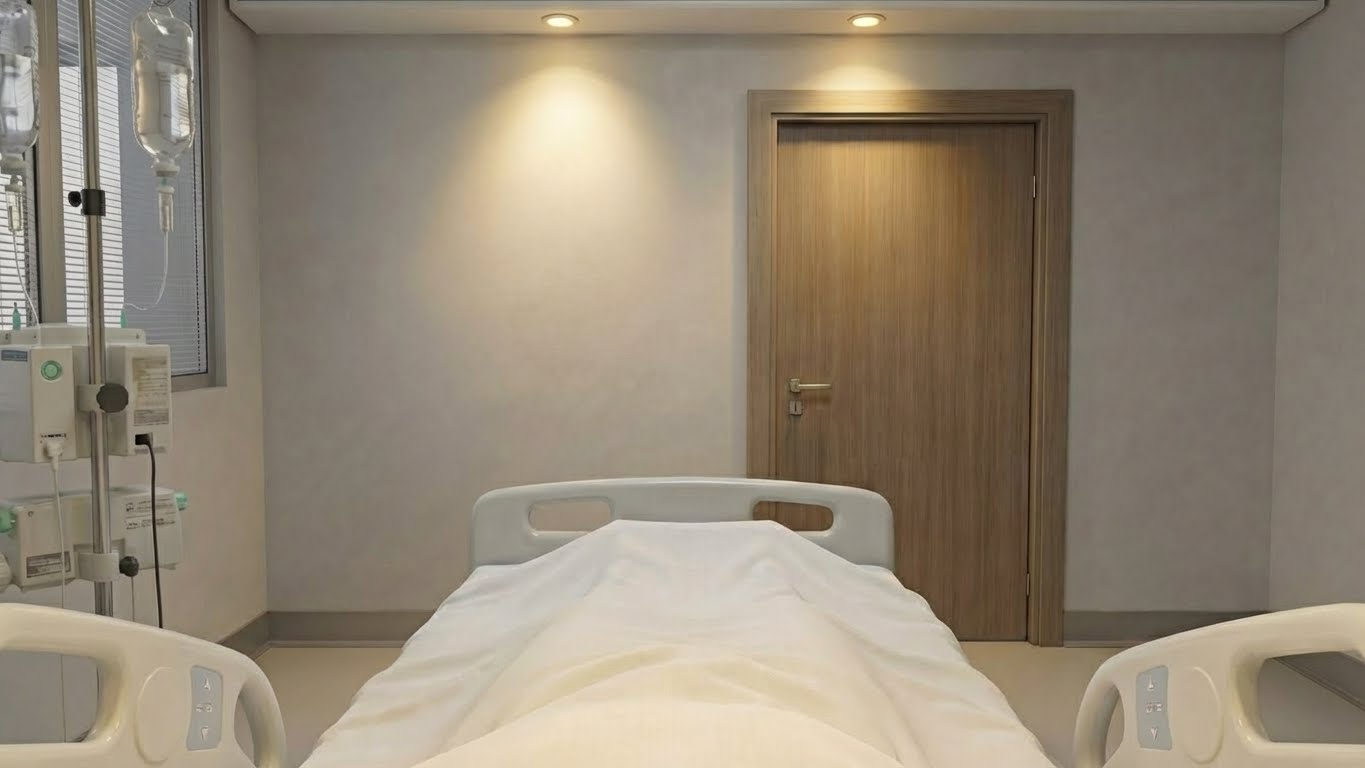}
\end{minipage}%
\hfill
\begin{minipage}[c]{0.52\textwidth}
    \small\textbf{(f) Intensive Care Unit (ICU)}
    
    {\scriptsize\textcolor{RoyalBlue}{\textit{Visual:}} A high-dependency environment with critical care equipment. Patient bed flanked by mechanical ventilators, infusion pump towers, and IV poles with focused, warm lighting.
    
    \textcolor{ForestGreen}{\textit{Narrative:}} Reserved for critical patient care with life-support systems. Used for vital sign monitoring and urgent communications between Doctor (Player) and patient.}
\end{minipage}

\end{tcolorbox}
\caption{Clinical locations ($\mathcal{L}$) used in MedGame. Each row shows two perspectives: \textit{facing-patient/staff} (left) and \textit{facing-doctor} (right). Descriptions include \textcolor{RoyalBlue}{\textit{Visual}} appearance and \textcolor{ForestGreen}{\textit{Narrative}} purpose.}
\label{fig:clinical_locations}
\end{figure*}

Each location is defined with visual appearance and narrative purpose descriptions that guide the LLM during story generation. Figure~\ref{fig:clinical_locations} illustrates all six clinical locations used in MedGame, showing both the \textit{facing-patient} perspective (first-person view from doctor) and the \textit{facing-doctor} perspective (view toward the doctor/player).

\subsubsection{Medical Staff Personas ($\mathcal{P}$)}

We define five medical staff roles that interact with the player throughout the clinical narrative. Each role has multiple visual variants with diverse gender, age, and ethnicity to promote inclusive representation.

\begin{figure*}[htbp]
\centering
\begin{tcolorbox}[
    colback=white,
    colframe=gray!50,
    arc=4mm,
    boxrule=0.5pt,
    left=3mm,
    right=3mm,
    top=3mm,
    bottom=3mm
]

\noindent
\begin{minipage}[c]{0.15\textwidth}
    \centering
    \includegraphics[width=\linewidth]{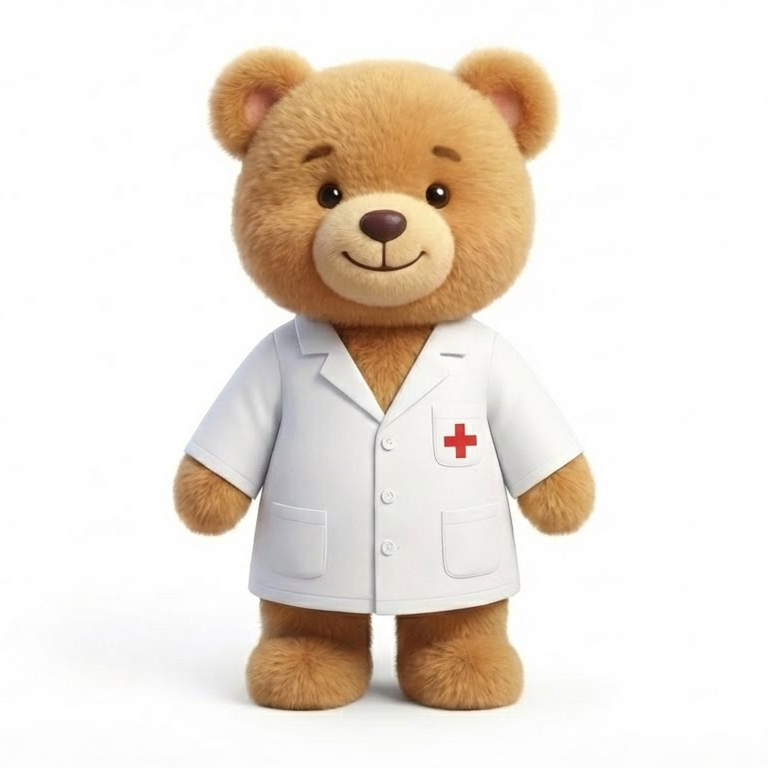}
\end{minipage}%
\hfill
\begin{minipage}[c]{0.80\textwidth}
    \small\textbf{(a) Doctor (Player)}
    
    {\scriptsize The protagonist controlled by the learner. Makes all clinical decisions including diagnosis, ordering tests, and treatment planning. Appears in all clinical environments. Represented as a friendly bear character to create an approachable learning atmosphere.}
\end{minipage}

\smallskip
\noindent\makebox[\linewidth]{\textcolor{gray!40}{\rule{0.95\linewidth}{0.3pt}}}
\smallskip

\noindent
\begin{minipage}[c]{0.15\textwidth}
    \centering
    \includegraphics[width=\linewidth]{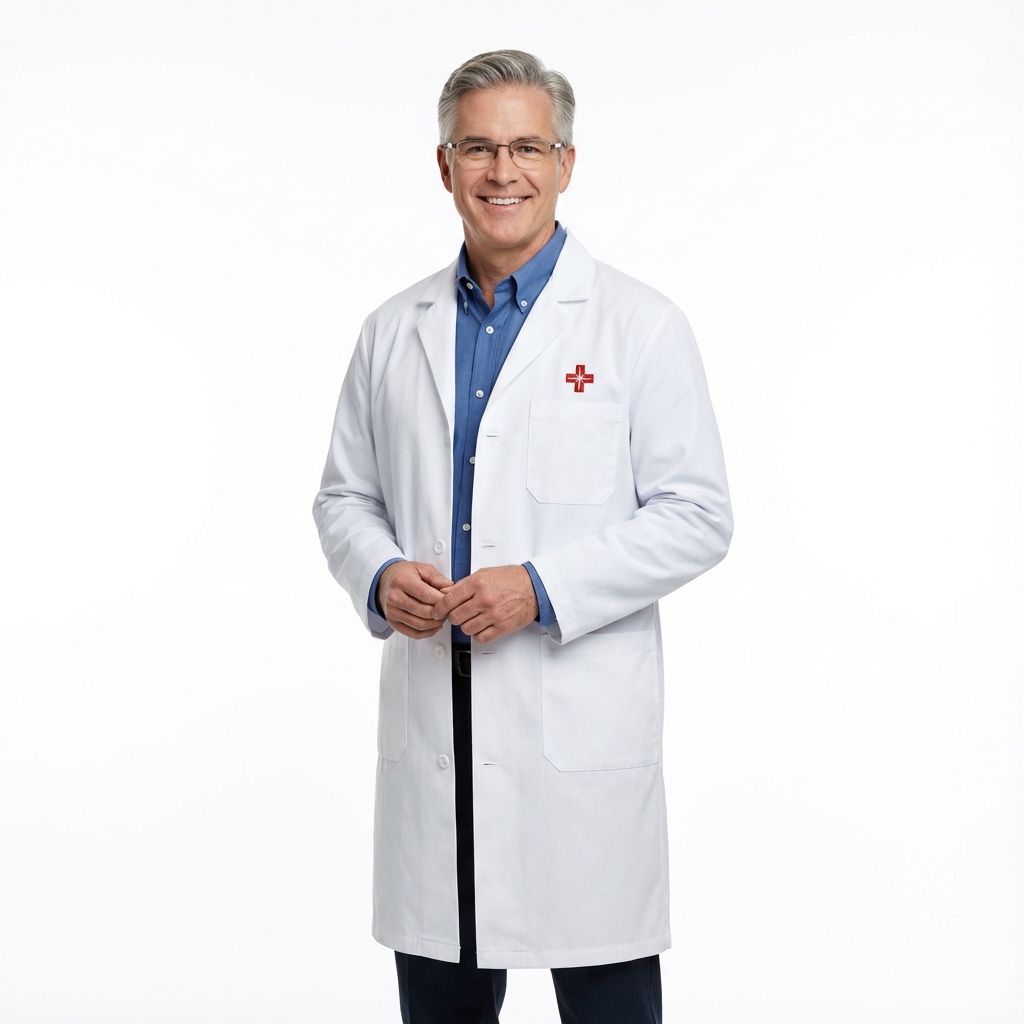}
\end{minipage}%
\hfill
\begin{minipage}[c]{0.15\textwidth}
    \centering
    \includegraphics[width=\linewidth]{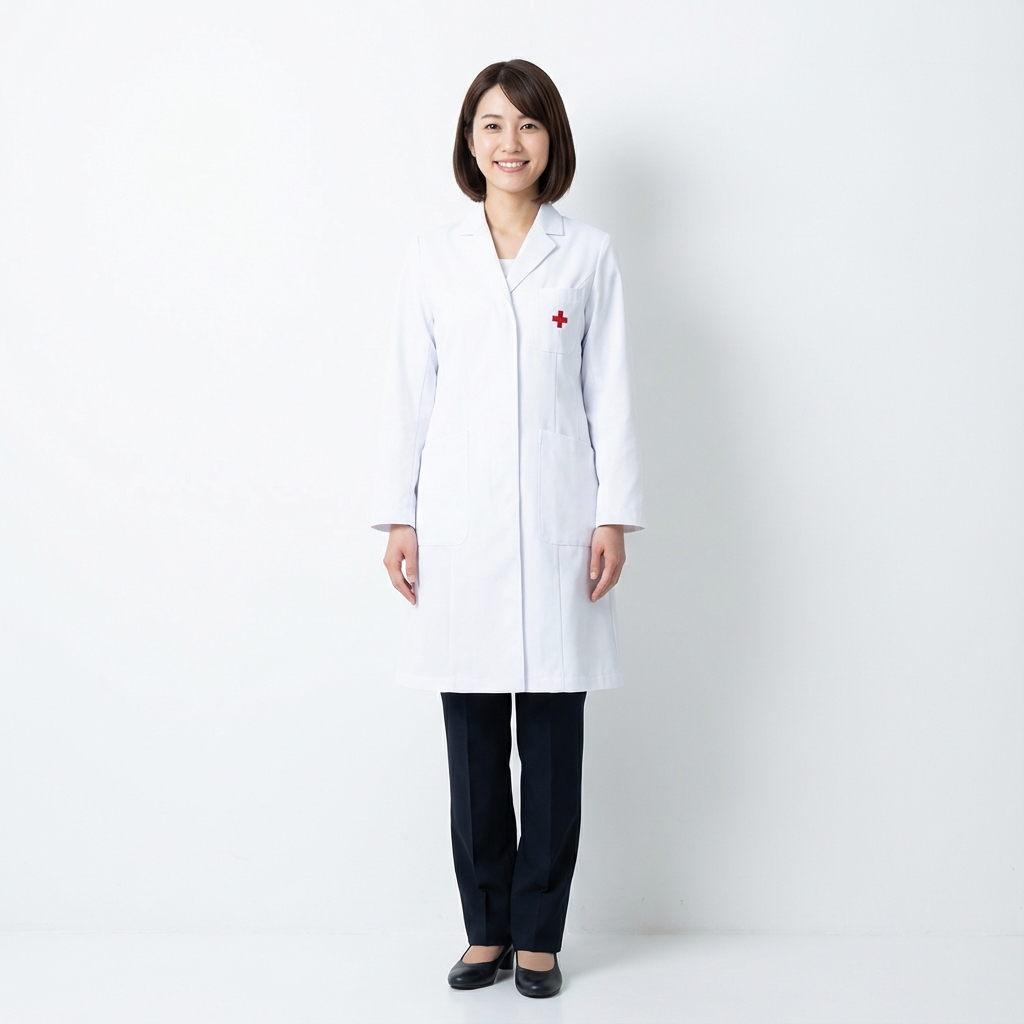}
\end{minipage}%
\hfill
\begin{minipage}[c]{0.65\textwidth}
    \small\textbf{(b) Pathologist}
    
    {\scriptsize Specialist in anatomical pathology who analyzes tissue specimens and biopsies. Provides definitive diagnoses for malignancies and tissue abnormalities. 2 variants (Male/Female). Locations: Pathology Department, Consultation Room.}
\end{minipage}

\smallskip
\noindent\makebox[\linewidth]{\textcolor{gray!40}{\rule{0.95\linewidth}{0.3pt}}}
\smallskip

\noindent
\begin{minipage}[c]{0.15\textwidth}
    \centering
    \includegraphics[width=\linewidth]{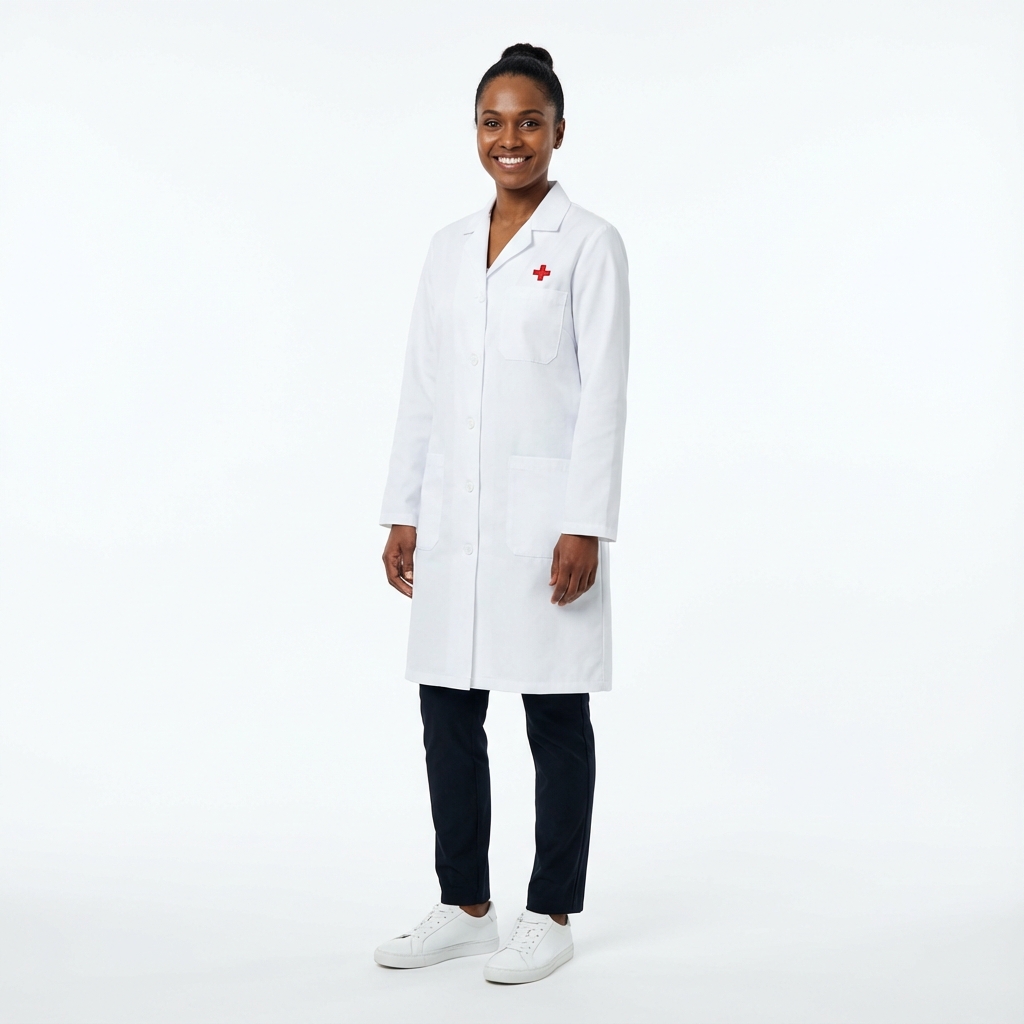}
\end{minipage}%
\hfill
\begin{minipage}[c]{0.15\textwidth}
    \centering
    \includegraphics[width=\linewidth]{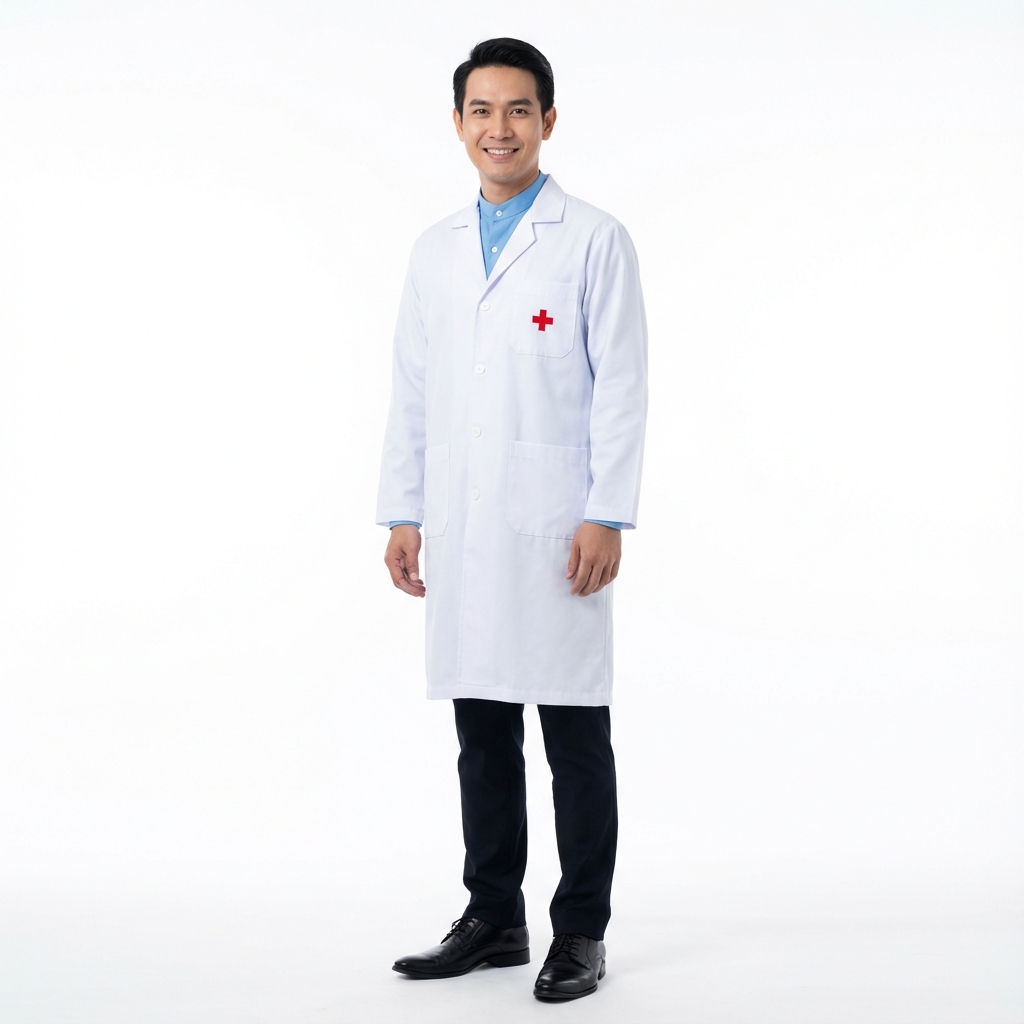}
\end{minipage}%
\hfill
\begin{minipage}[c]{0.65\textwidth}
    \small\textbf{(c) Lab Physician}
    
    {\scriptsize Clinical laboratory specialist who interprets blood tests, urinalysis, and other body fluid analyses. Explains abnormal markers and their clinical significance. 2 variants (Male/Female). Locations: Clinical Laboratory, Consultation Room.}
\end{minipage}

\smallskip
\noindent\makebox[\linewidth]{\textcolor{gray!40}{\rule{0.95\linewidth}{0.3pt}}}
\smallskip

\noindent
\begin{minipage}[c]{0.15\textwidth}
    \centering
    \includegraphics[width=\linewidth]{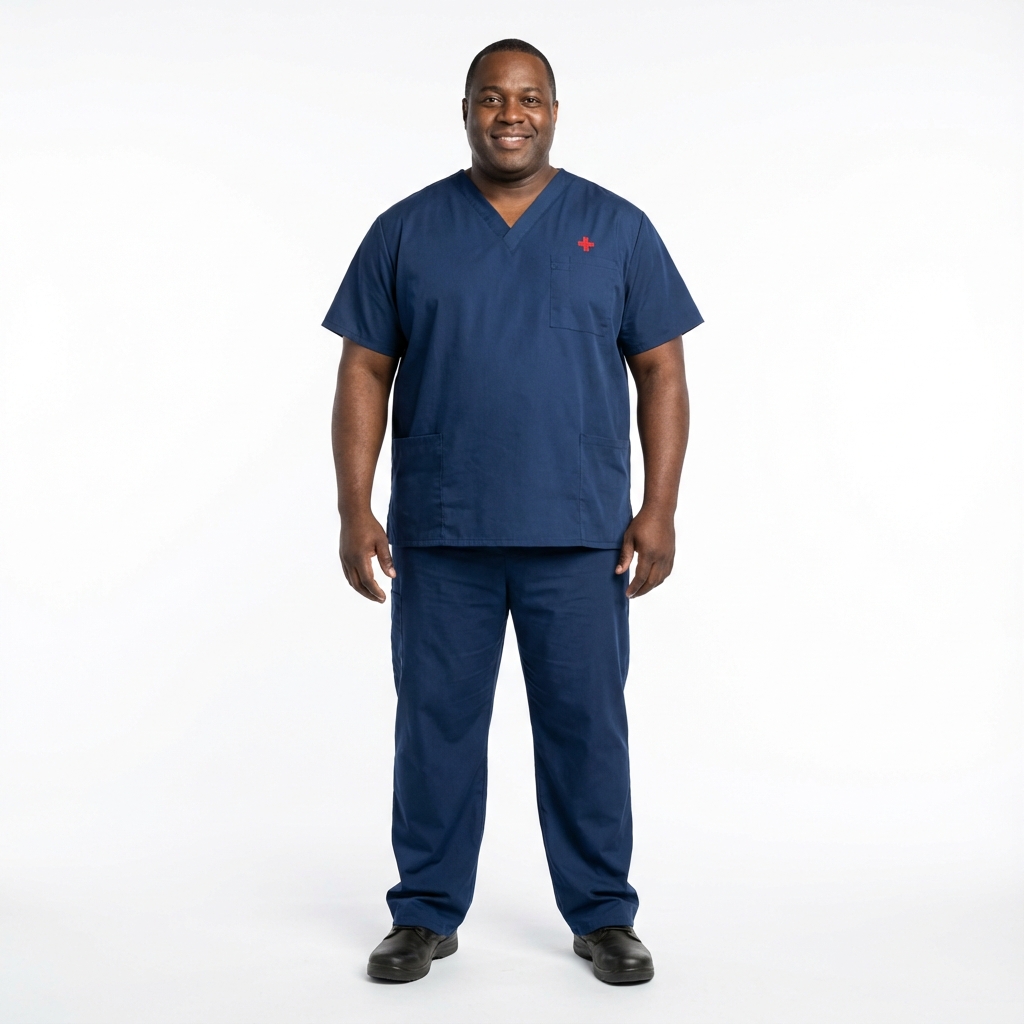}
\end{minipage}%
\hfill
\begin{minipage}[c]{0.15\textwidth}
    \centering
    \includegraphics[width=\linewidth]{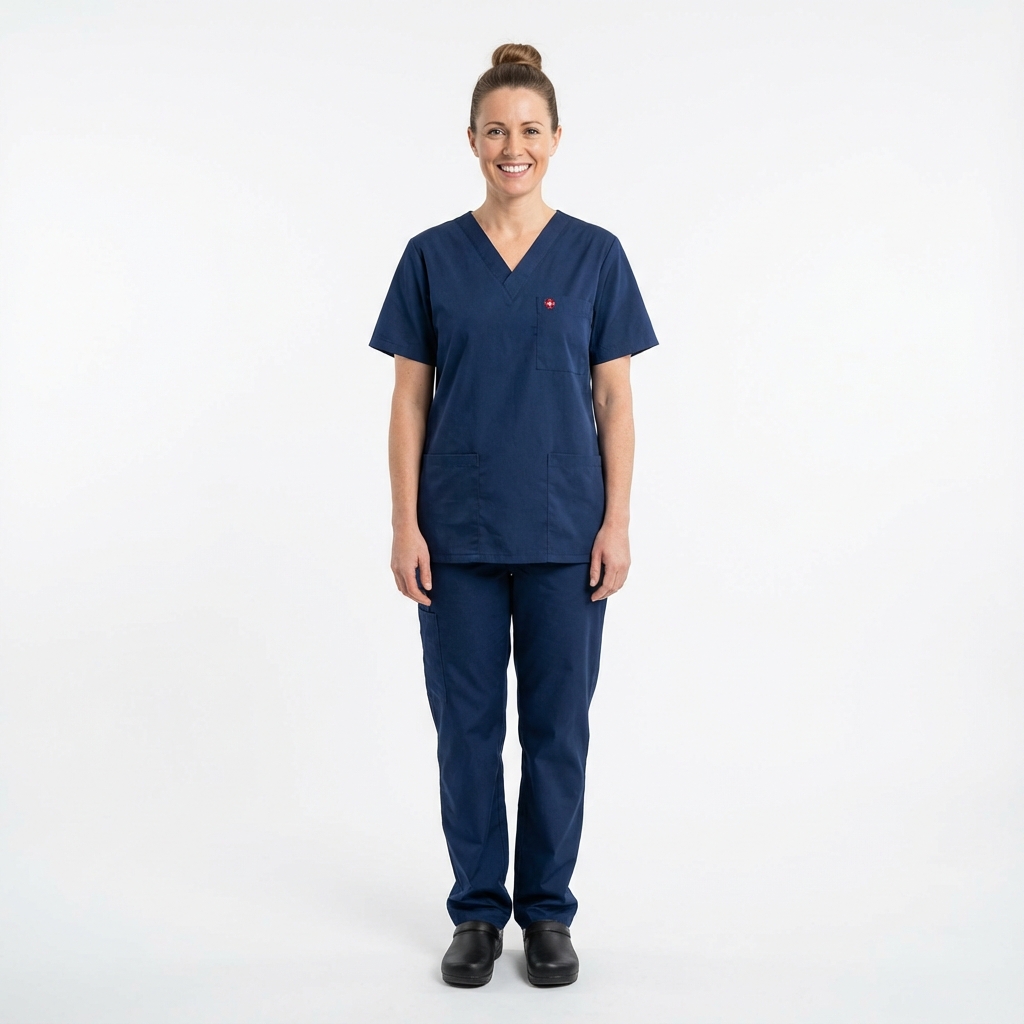}
\end{minipage}%
\hfill
\begin{minipage}[c]{0.65\textwidth}
    \small\textbf{(d) Radiologist}
    
    {\scriptsize Medical imaging specialist who interprets X-rays, CT scans, MRIs, and ultrasounds. Identifies anatomical abnormalities and provides imaging-based diagnoses. 2 variants (Male/Female). Locations: Radiology Department, Consultation Room.}
\end{minipage}

\smallskip
\noindent\makebox[\linewidth]{\textcolor{gray!40}{\rule{0.95\linewidth}{0.3pt}}}
\smallskip

\noindent
\begin{minipage}[c]{0.11\textwidth}
    \centering
    \includegraphics[width=\linewidth]{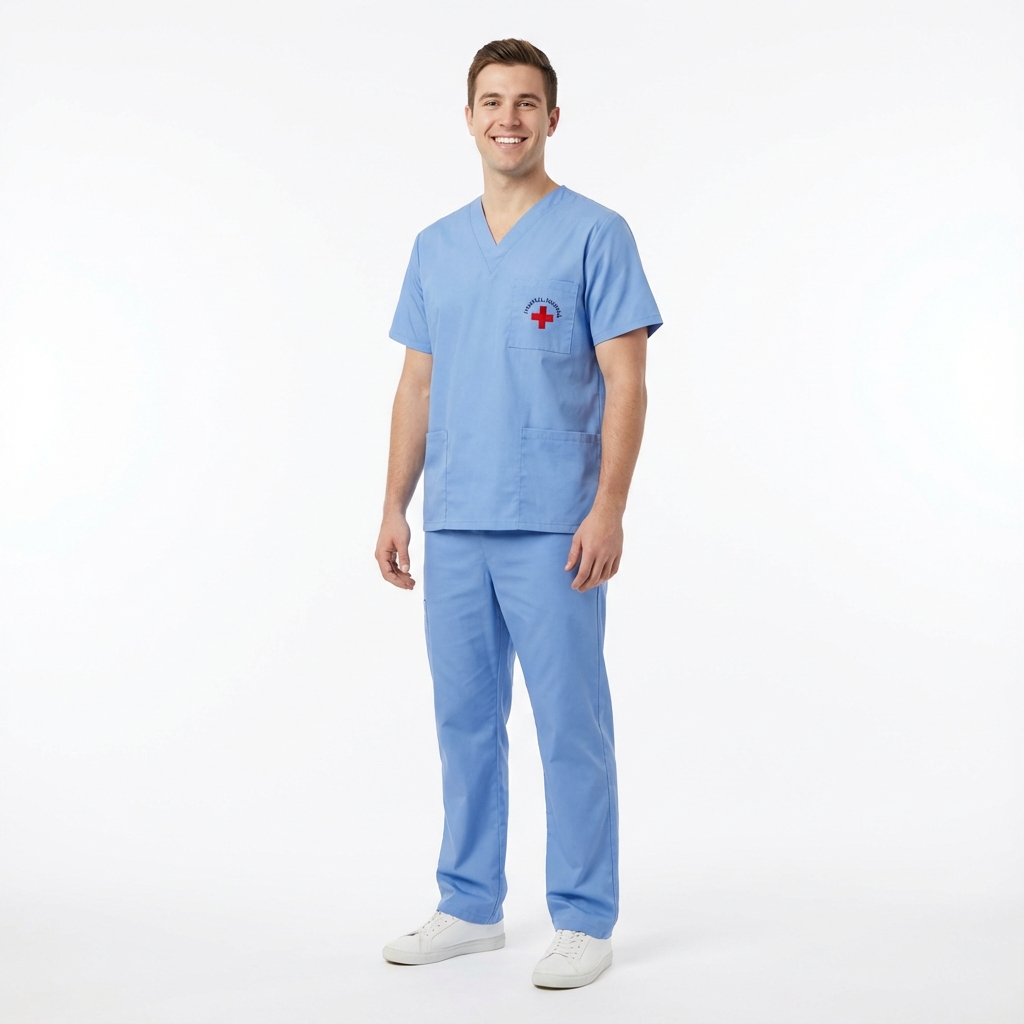}
\end{minipage}%
\hfill
\begin{minipage}[c]{0.11\textwidth}
    \centering
    \includegraphics[width=\linewidth]{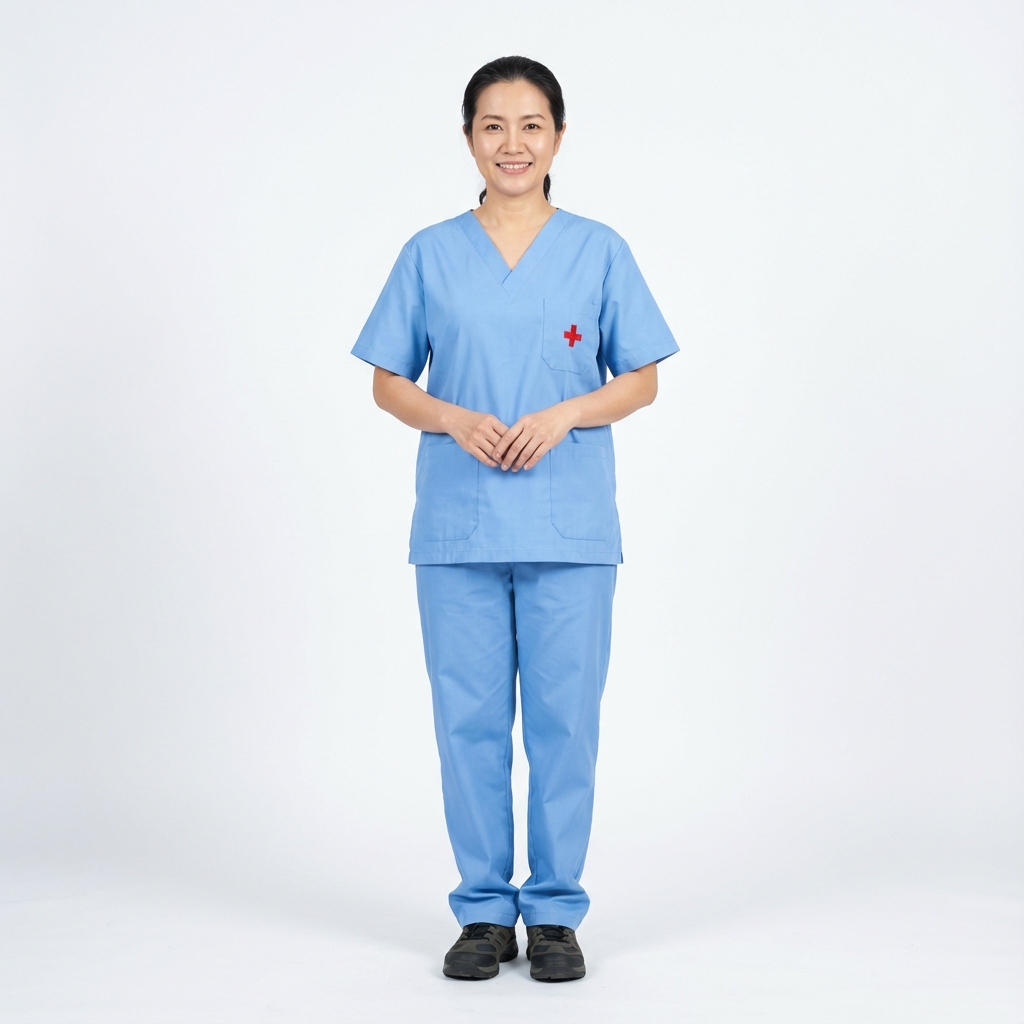}
\end{minipage}%
\hfill
\begin{minipage}[c]{0.11\textwidth}
    \centering
    \includegraphics[width=\linewidth]{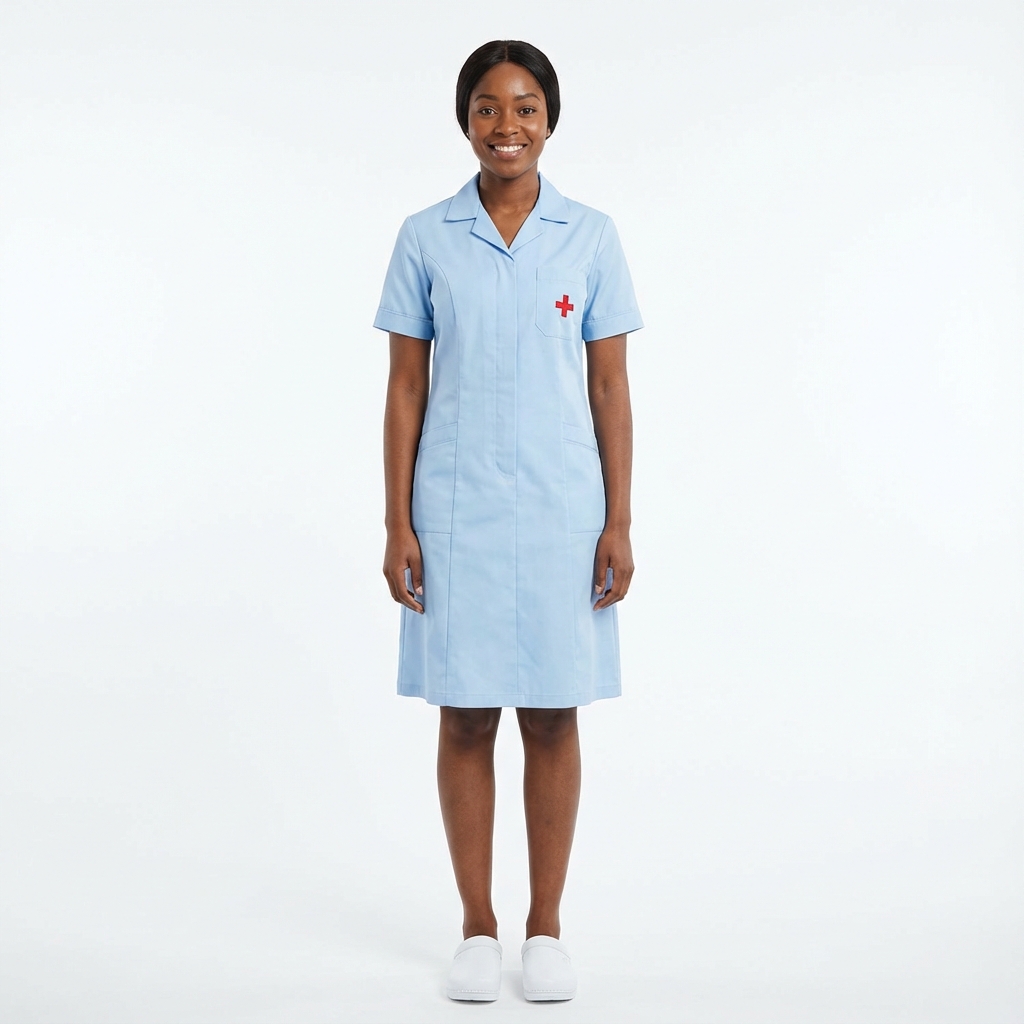}
\end{minipage}%
\hfill
\begin{minipage}[c]{0.62\textwidth}
    \small\textbf{(e) Nurse}
    
    {\scriptsize Provides bedside patient care, monitors vital signs, administers medications, and reports patient status changes to the doctor. 3 variants with diverse representation. Locations: General Ward, ICU, Consultation Room.}
\end{minipage}

\end{tcolorbox}
\caption{Medical staff personas ($\mathcal{P}$) used in MedGame. Each role has multiple visual variants to promote inclusive representation. The Doctor (Player) is represented as a bear character to create an approachable learning atmosphere.}
\label{fig:medical_staff_personas}
\end{figure*}

Figure~\ref{fig:medical_staff_personas} illustrates all medical staff personas used in MedGame, with multiple visual variants for diversity.

\noindent
\textbf{Design Rationale.} By constraining the narrative elements to a fixed set of locations and personas, we achieve three objectives: (1) \textit{Visual Consistency}---all generated stories share a unified aesthetic, preventing jarring style variations; (2) \textit{Asset Efficiency}---pre-rendered backgrounds and character sprites eliminate the need for per-story image generation; (3) \textit{Narrative Coherence}---the story generation model can reference concrete locations and roles, producing more grounded and realistic clinical scenarios.

\FloatBarrier

\subsection{Details of Multimodal Toolset}
\label{appendix section: Details of Multimodal Toolset}

As introduced in Section~\ref{section: the MedGame Framework}, the multimodal toolset $\mathcal{M} = \{f_{vis}, f_{aud}, f_{vid}\}$ serves as a rendering layer that materializes textual clinical logic into sensory representations. We detail each tool below, with emphasis on the visual generation tool $f_{vis}$ which serves as the anchor for downstream audio and video synthesis.

\subsubsection{Visual Generation Tool ($f_{vis}$)}

The visual generation tool $f_{vis}$ serves as the \textbf{anchor} for our multimodal rendering pipeline. As described in Section~\ref{sec:director}, each generation task begins with an \textit{initial frame synthesis} that fixes canonical visual attributes---such as character identity and scene layout---providing the structural foundation for subsequent audio and video generation.

We implement $f_{vis}$ using Gemini-3-Pro-Image\footnote{https://deepmind.google/models/gemini-image/pro/}, which supports three generation modes: \textbf{CharacterGen}, \textbf{Fusion}, and \textbf{Modification}. Table~\ref{tab:image-tool-modes} summarizes each mode's interface and usage scenarios.

\begin{table}[htbp]
\centering
\small
\caption{Three generation modes of the visual generation tool $f_{vis}$.}
\label{tab:image-tool-modes}
\resizebox{\linewidth}{!}{%
\begin{tabular}{@{}lp{5cm}p{5cm}@{}}
\toprule
\textbf{Mode} & \textbf{Input} & \textbf{Output} \\
\midrule
CharacterGen & Text description (age, gender, ethnicity, appearance, expression, clothing) & Character portrait with white background \\
\addlinespace
Fusion & Character image(s) + Scene image + Position/Posture description & Composed scene with character(s) naturally integrated \\
\addlinespace
Modification & Base image + Modification target + Modification details & Updated image with specified changes \\
\bottomrule
\end{tabular}%
}
\end{table}

The following API specification is provided to the Story Director LLM to guide the generation of image parameters:

\begin{tcolorbox}[title = {API Specification for Visual Generation Tool $f_{vis}$}, 
                  fonttitle=\normalsize, 
                  fontupper=\small,
                  colback=gray!5!white,
                  colframe=black!75!white,
                  breakable]

\textbf{\textcolor{RoyalBlue}{1. character\_gen}} --- Generate character portrait with pure white background.\\
\textit{Use Case:} Create patient images for the story (first appearance).\\
\textit{Parameters:} \texttt{age} (int), \texttt{ethnicity} (string), \texttt{gender} (string), \texttt{appearance} (string), \texttt{expression} (string), \texttt{clothing} (string), \texttt{shot\_type} (string, default: ``portrait'').

\noindent\makebox[\linewidth]{\textcolor{gray}{\leaders\hbox{\rule{1pt}{0pt}\rule[-0.2ex]{5pt}{0.5pt}}\hfill}}

\textbf{\textcolor{RoyalBlue}{2. fusion}} --- Fuse one or multiple person images into a scene at specified positions.\\
\textit{Use Case:} Place characters into clinical environments. Supports single-person and multi-person scenarios.\\
\textit{Parameters:} \texttt{person\_image\_path} (string or list), \texttt{scene\_image\_path} (string), \texttt{location\_description} (string or list), \texttt{posture\_expression} (string or list).\\
\textit{Note:} For multi-person fusion, all list parameters must have equal length.

\noindent\makebox[\linewidth]{\textcolor{gray}{\leaders\hbox{\rule{1pt}{0pt}\rule[-0.2ex]{5pt}{0.5pt}}\hfill}}

\textbf{\textcolor{RoyalBlue}{3. modification}} --- Modify the state of person(s) in an existing image.\\
\textit{Use Case:} Adjust expression, posture, or clothing while preserving identity. Essential for reflecting narrative progression.\\
\textit{Modifiable Attributes:} Facial expression, body posture, action state, clothing style, interaction with objects.\\
\textit{Parameters:} \texttt{input\_image\_path} (string, supports \texttt{[task\_XXX\_output]} placeholder), \texttt{modification\_target} (string or list), \texttt{modification\_details} (string or list).

\noindent\makebox[\linewidth]{\textcolor{gray}{\leaders\hbox{\rule{1pt}{0pt}\rule[-0.2ex]{5pt}{0.5pt}}\hfill}}

\textbf{Path Placeholders:}
\begin{itemize}[noitemsep, topsep=0pt, leftmargin=*]
    \item \texttt{[character\_source\_dir]} --- Pre-defined medical staff assets (e.g., \texttt{Radiologist\_1.png})
    \item \texttt{[scene\_source\_dir]} --- Pre-rendered clinical backgrounds
    \item \texttt{[task\_XXX\_output]} --- Dependency reference to previous task's output
\end{itemize}

\end{tcolorbox}

\begin{figure*}[htbp]
\centering
\begin{tcolorbox}[
    colback=white,
    colframe=gray!50,
    arc=4mm,
    boxrule=0.5pt,
    left=3mm,
    right=3mm,
    top=3mm,
    bottom=3mm
]

\noindent
\begin{minipage}[c]{0.45\textwidth}
    \centering
    \small\textbf{(a) CharacterGen: Text $\rightarrow$ Portrait}
    
    \smallskip
    {\scriptsize\textit{Input:} Text description specifying age, gender, ethnicity, appearance, expression, and clothing.}
    
    \smallskip
    \fbox{\parbox{0.9\linewidth}{\tiny\texttt{age: 15, gender: Male, ethnicity: White, appearance: ``pale skin, brown hair, looking sick'', expression: ``distressed, painful grimace'', clothing: ``casual grey t-shirt''}}}
\end{minipage}%
\hfill
\begin{minipage}[c]{0.08\textwidth}
    \centering
    {\Large $\Rightarrow$}
\end{minipage}%
\hfill
\begin{minipage}[c]{0.15\textwidth}
    \centering
    \includegraphics[width=\linewidth]{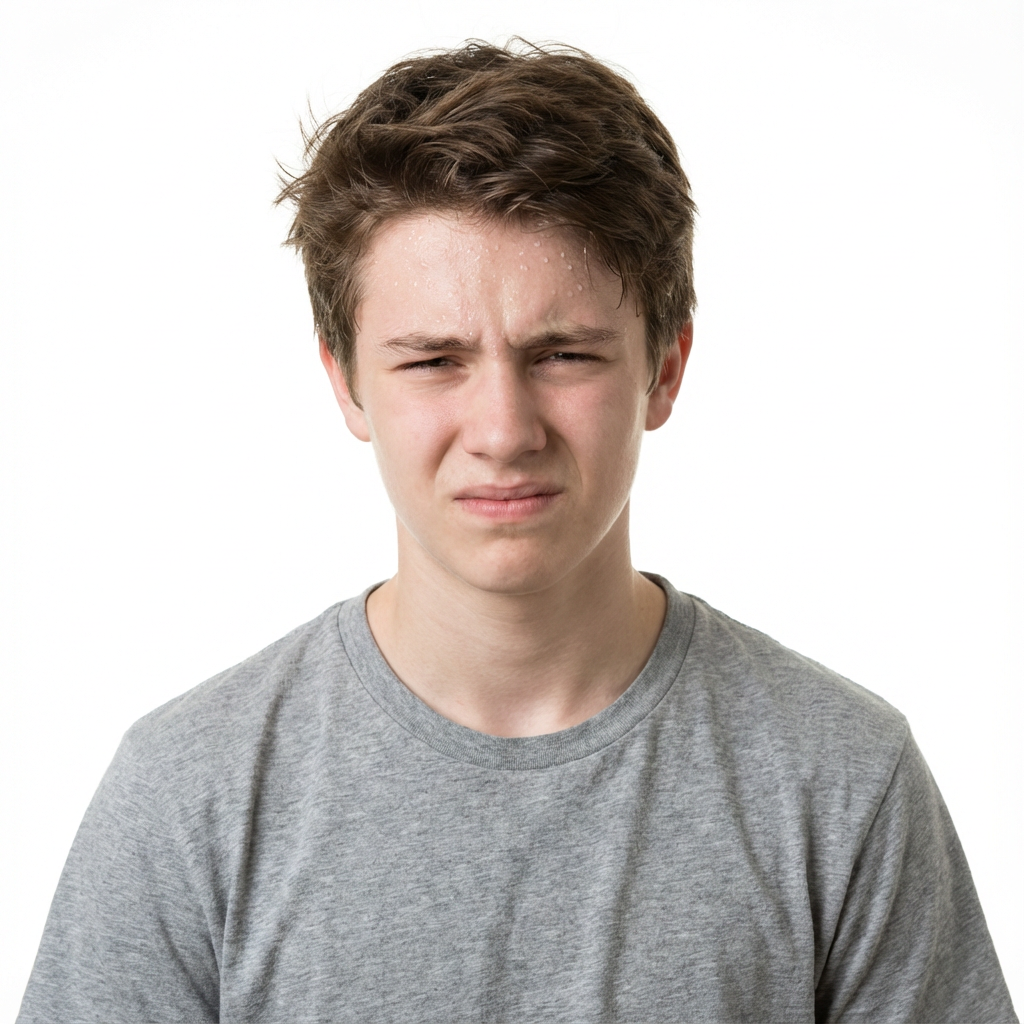}
    
    {\tiny Output: Patient portrait}
\end{minipage}

\medskip
\noindent\makebox[\linewidth]{\textcolor{gray!40}{\rule{0.95\linewidth}{0.3pt}}}
\medskip

\noindent
\begin{minipage}[c]{0.15\textwidth}
    \centering
    \includegraphics[width=\linewidth]{Appendix/Figures/ImageToolExamples/charactergen_output.png}
    
    {\tiny Character}
\end{minipage}%
\hfill
\begin{minipage}[c]{0.04\textwidth}
    \centering
    {\large $+$}
\end{minipage}%
\hfill
\begin{minipage}[c]{0.22\textwidth}
    \centering
    \includegraphics[width=\linewidth]{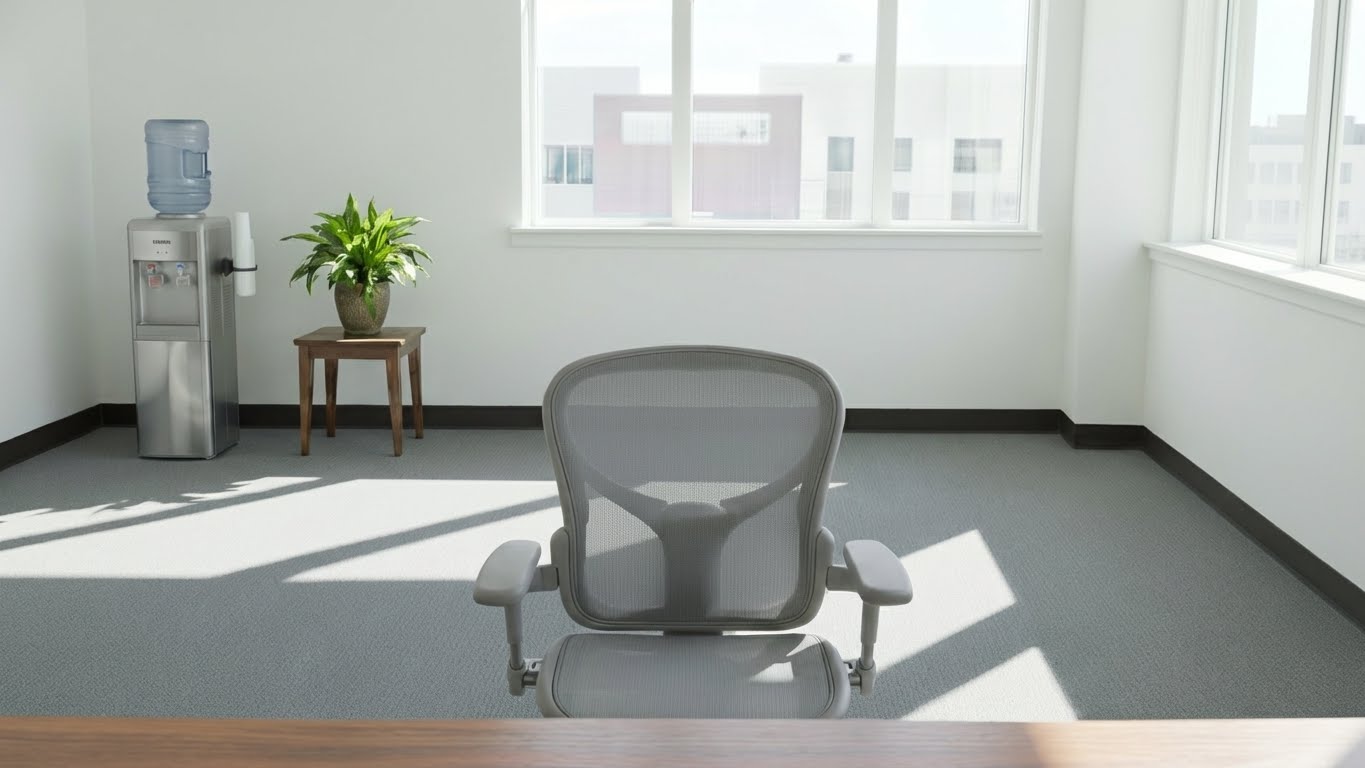}
    
    {\tiny Scene}
\end{minipage}%
\hfill
\begin{minipage}[c]{0.08\textwidth}
    \centering
    {\Large $\Rightarrow$}
\end{minipage}%
\hfill
\begin{minipage}[c]{0.22\textwidth}
    \centering
    \includegraphics[width=\linewidth]{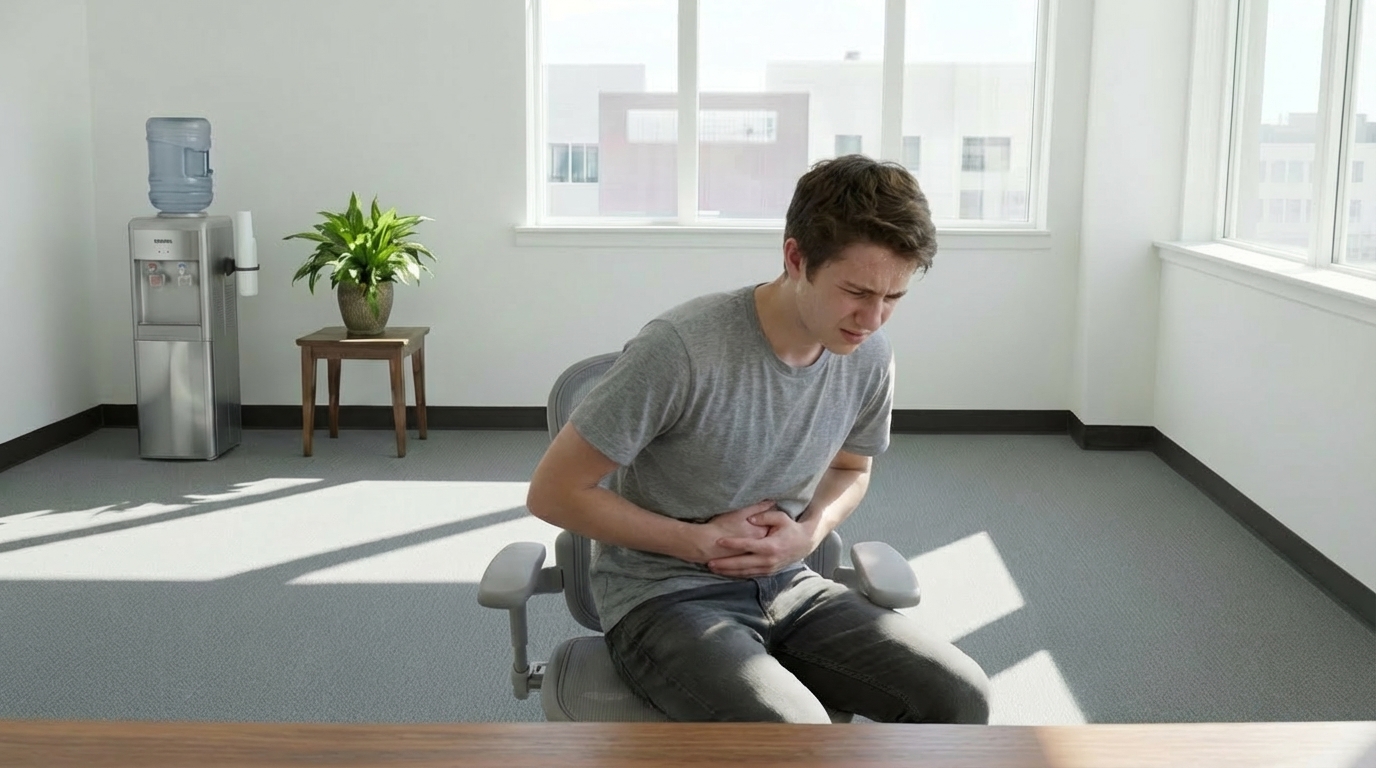}
    
    {\tiny Output}
\end{minipage}%
\hfill
\begin{minipage}[c]{0.22\textwidth}
    \centering
    \small\textbf{(b) Fusion: Patient}
    
    {\scriptsize Fuse generated patient portrait into consultation room scene.}
\end{minipage}

\medskip
\noindent\makebox[\linewidth]{\textcolor{gray!40}{\rule{0.95\linewidth}{0.3pt}}}
\medskip

\noindent
\begin{minipage}[c]{0.15\textwidth}
    \centering
    \includegraphics[width=\linewidth]{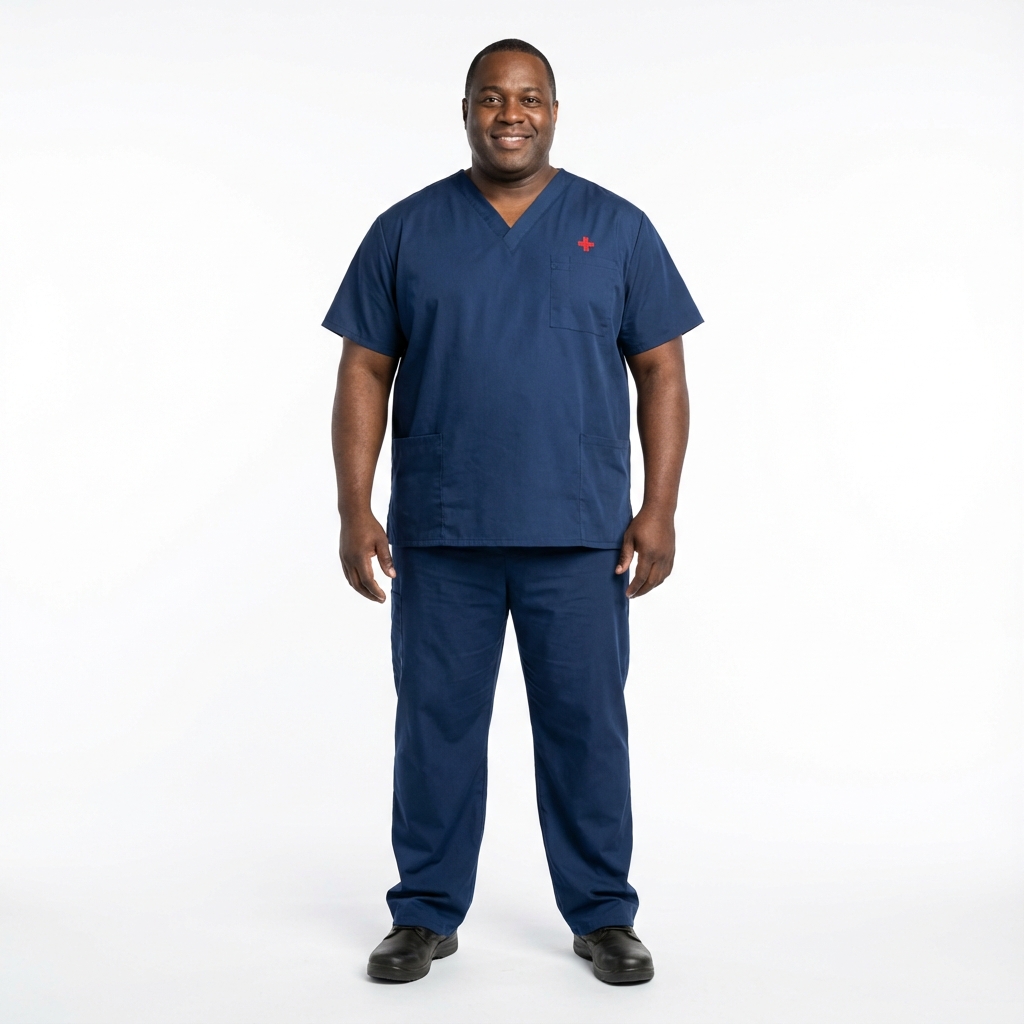}
    
    {\tiny Character}
\end{minipage}%
\hfill
\begin{minipage}[c]{0.04\textwidth}
    \centering
    {\large $+$}
\end{minipage}%
\hfill
\begin{minipage}[c]{0.22\textwidth}
    \centering
    \includegraphics[width=\linewidth]{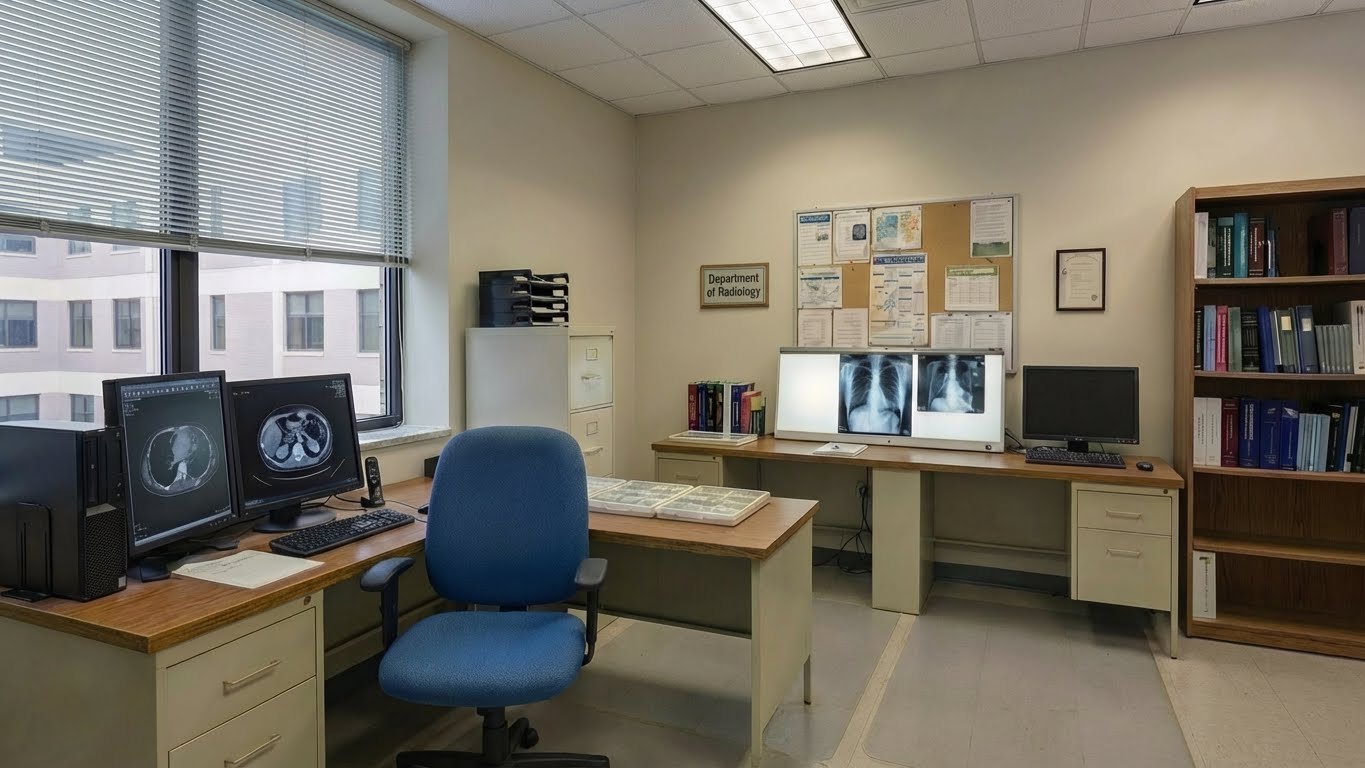}
    
    {\tiny Scene}
\end{minipage}%
\hfill
\begin{minipage}[c]{0.08\textwidth}
    \centering
    {\Large $\Rightarrow$}
\end{minipage}%
\hfill
\begin{minipage}[c]{0.22\textwidth}
    \centering
    \includegraphics[width=\linewidth]{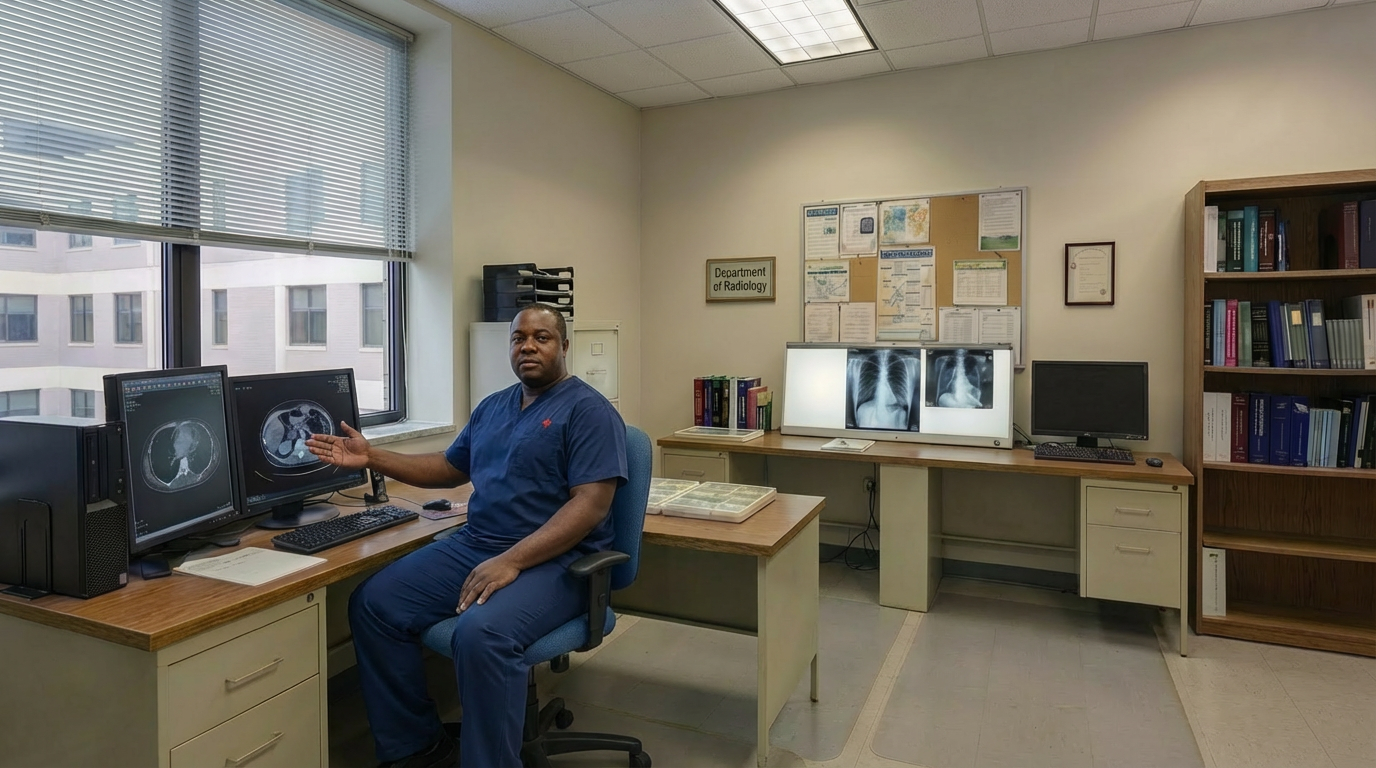}
    
    {\tiny Output}
\end{minipage}%
\hfill
\begin{minipage}[c]{0.22\textwidth}
    \centering
    \small\textbf{(c) Fusion: Staff}
    
    {\scriptsize Fuse pre-defined radiologist asset into radiology room.}
\end{minipage}

\medskip
\noindent\makebox[\linewidth]{\textcolor{gray!40}{\rule{0.95\linewidth}{0.3pt}}}
\medskip

\noindent
\begin{minipage}[c]{0.15\textwidth}
    \centering
    \includegraphics[width=\linewidth]{Appendix/Figures/ImageToolExamples/charactergen_output.png}
    
    {\tiny Input: Act I portrait}
\end{minipage}%
\hfill
\begin{minipage}[c]{0.08\textwidth}
    \centering
    {\Large $\Rightarrow$}
\end{minipage}%
\hfill
\begin{minipage}[c]{0.15\textwidth}
    \centering
    \includegraphics[width=\linewidth]{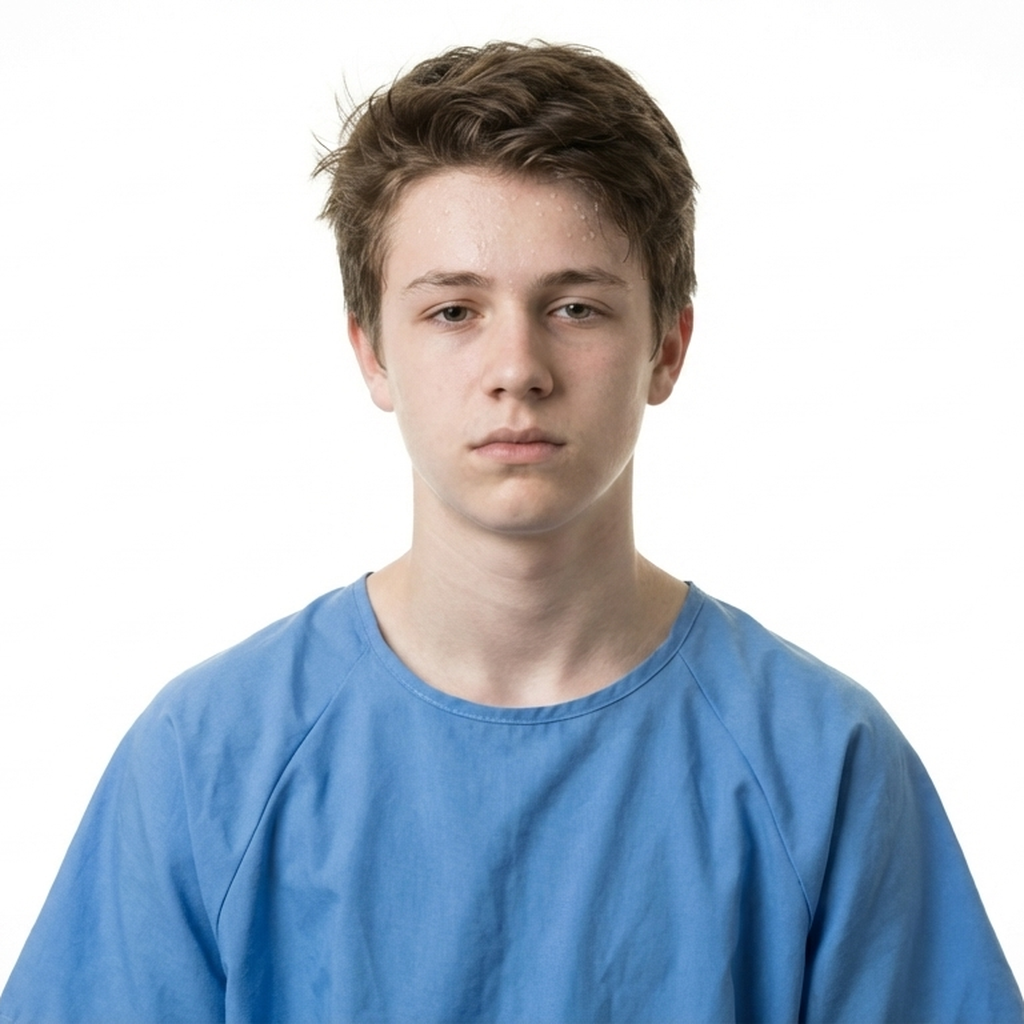}
    
    {\tiny Output: Act II portrait}
\end{minipage}%
\hfill
\begin{minipage}[c]{0.42\textwidth}
    \centering
    \small\textbf{(d) Modification: Clothing Change}
    
    {\scriptsize Modify patient's clothing from casual t-shirt to hospital gown while preserving identity. Reflects narrative progression from Act I to Act II (patient admission).}
\end{minipage}

\end{tcolorbox}
\caption{Visual generation tool $f_{vis}$ examples from Case 5915692-2. \textbf{(a)} CharacterGen creates patient portraits from text. \textbf{(b-c)} Fusion composes characters into scenes. \textbf{(d)} Modification updates attributes while preserving identity.}
\label{fig:image-tool-examples}
\end{figure*}

Figure~\ref{fig:image-tool-examples} illustrates the three generation modes with input-output examples from Case 5915692-2.

\subsubsection{Audio Generation Tool ($f_{aud}$)}

The audio generation tool $f_{aud}$ synthesizes character speech from dialogue text. The pipeline consists of two stages:

\paragraph{Stage 1: Emotion Tagging.}
An LLM analyzes each dialogue segment within its narrative context and inserts emotion tags to guide expressive speech synthesis. Available tags include general emotions (\texttt{[happy]}, \texttt{[worried]}, \texttt{[thoughtful]}), delivery styles (\texttt{[whisper]}, \texttt{[softly]}, \texttt{[firmly]}), and role-appropriate expressions. Medical staff are constrained to professional emotions (\texttt{[warmly]}, \texttt{[calmly]}, \texttt{[reassuringly]}), while patients may express stronger emotions (\texttt{[anxiously]}, \texttt{[painfully]}, \texttt{[hopefully]}).

\begin{tcolorbox}[fontupper=\small, colback=gray!5!white, colframe=black!50!white]
\textbf{Example --- Before:} ``I've been experiencing these symptoms for weeks now...''

\textbf{Example --- After:} ``\texttt{[anxiously]} I've been experiencing these symptoms for weeks now...''
\end{tcolorbox}

\paragraph{Stage 2: Voice Synthesis.}
We use ElevenLabs\footnote{\url{https://elevenlabs.io}} Text-to-Dialogue API for multi-speaker voice synthesis. Each character role is mapped to a distinct voice ID:
\begin{itemize}[noitemsep, topsep=2pt, leftmargin=*]
    \item \textbf{Patient}: Unique voice cloned per case to maintain identity consistency across acts.
    \item \textbf{Medical Staff}: Pre-assigned voices from the character reference (e.g., Pathologist\_1, Radiologist\_2).
\end{itemize}
The API returns synthesized audio with word-level timestamps, enabling precise lip-sync alignment for downstream video generation.

\subsubsection{Video Generation Tool ($f_{vid}$)}

The video generation tool $f_{vid}$ creates character animations with lip-sync and expressive motion. We employ InfiniteTalk~\cite{yang2025infinitetalk}, an audio-driven video generation model for sparse-frame dubbing that synthesizes videos with accurate lip synchronization while simultaneously aligning head movements, body posture, and facial expressions with the audio. The model takes three inputs:

\begin{enumerate}[noitemsep, topsep=2pt, leftmargin=*]
    \item \textbf{Audio}: The synthesized speech from $f_{aud}$, which drives lip movements.
    \item \textbf{First Frame}: A static character image (from $f_{vis}$), defining the visual appearance.
    \item \textbf{Text Description}: A natural language prompt describing the character's state, posture, and expression.
\end{enumerate}

\paragraph{Description Generation.}
An LLM generates video descriptions by analyzing:
\begin{itemize}[noitemsep, topsep=2pt, leftmargin=*]
    \item \textbf{Character States}: Whether each character in frame is \textit{talking} (audio contains their dialogue) or \textit{listening} (only appears visually).
    \item \textbf{Emotion Tags}: Extracted from the dialogue to inform facial expressions.
    \item \textbf{Posture Information}: From the image generation parameters (e.g., ``sitting hunched over, clutching abdomen'').
\end{itemize}

\begin{tcolorbox}[fontupper=\small, colback=gray!5!white, colframe=black!50!white]
\textbf{Single Character:} ``The male patient sits hunched, talking painfully while clutching his stomach.''

\textbf{Multiple Characters:} ``The doctor bear on the left listens, nodding thoughtfully; the female patient on the right talks anxiously.''

\textbf{Doctor Thinking:} ``The doctor bear sits at the desk, gazing at the screen in deep thought.''
\end{tcolorbox}

\paragraph{Generation Constraints.}
The video model is optimized for continuous emotional states and micro-expressions rather than complex physical actions:
\begin{itemize}[noitemsep, topsep=2pt, leftmargin=*]
    \item \textbf{Suitable}: Emotion descriptions, listening gestures (nodding, tilting head), static postures.
    \item \textbf{Unsuitable}: Complex sequential actions (medical examinations), instantaneous movements (standing up, turning to leave).
\end{itemize}
For scenes requiring complex actions (e.g., physical examinations), we fall back to text-based presentation as specified in the narrative design principles.

\subsection{The Instruction used for Medical Narrative Design: $\Phi_{\text{des}}$}
\label{appendix section: prompt for narrative design}

The instruction $\Phi_{\text{des}}$ guides the LLM to transform a raw patient summary into a structured Clinical Story Tree. The complete prompt template is shown in Figure~\ref{fig:prompt_phi_des}.

\begin{figure*}[p]
\begin{tcolorbox}[title = {Instruction Template: $\Phi_{\text{des}}$}, 
                  fonttitle=\small, 
                  fontupper=\scriptsize,
                  colback=gray!5!white,
                  colframe=black!75!white]

\textbf{\# Objective:}

We are designing story-based scenarios for medical education. Our task is to convert the given \textbf{\textcolor{RoyalBlue}{\{input\_type\}}} into a structured, gamified narrative. The goal is to allow medical students to acquire knowledge through an immersive experience.

\textbf{\# Examples}

\textbf{\textcolor{RoyalBlue}{\{examples\_section\}}}

\textbf{\# Pydantic Model Definition for Gamified Plot}

Below is the Pydantic definition of the JSON structure I require. \textbf{Constraint:} You must strictly follow this schema, ensuring valid JSON output. Pay special attention to the \texttt{interaction\_type} Literal fields (e.g., ``single\_choice'', ``batch'', ``interactive'') as they define the question structure.

\textbf{\textcolor{RoyalBlue}{\{pydantic\_model\_example\}}} \hfill $\rightarrow$ \textit{See Listing~\ref{list: pydantic_models of clinical story}}

\textbf{\# Requirements and Design Principles}

\textbf{\#\# 1. Fidelity to the Source Material}

The progression of the plot---such as the treatment plans adopted and the test results---must align with the description in the original text. (However, distractor options used for assessment can be simulated/fictional).

\textbf{\#\# 2. Guidelines for Assessment Distractors}
\begin{itemize}[noitemsep, topsep=2pt, leftmargin=*]
\item \textbf{Avoid leading language:} Do not use words like ``merely,'' ``only,'' or ``just'' in distractors, as they make it too easy for the user to identify the wrong answer.
\item \textbf{Clinical Relevance:} Distractors should represent meaningful clinical decision points or common misconceptions, rather than obviously impossible options.
\end{itemize}

\textbf{\#\# 3. Design Principles for Acts and Scenes}
\begin{itemize}[noitemsep, topsep=2pt, leftmargin=*]
\item An \textbf{Act} can contain multiple \textbf{Scenes}.
\item In our design logic, a \textbf{Scene} represents an independent step in the clinical workflow where the player constitutes a decision-making entity.
\item An \textbf{Act} typically represents a major phase shift. It usually implies the patient leaving the current space (e.g., your clinic) and reappearing later due to specific reasons (e.g., a follow-up visit or an emergency return).
\end{itemize}

\textbf{\#\# 4. Dialogue vs. Static Text}

This is a critical component.
\begin{itemize}[noitemsep, topsep=2pt, leftmargin=*]
\item The \textbf{Roleplay} and \textbf{Interaction} sections of the structured plot can involve dialogue. Our video generation model will simulate these conversations with the plot characters.
\item However, for complex technical operations---such as physical examinations or detailed lab reports---prefer using \textbf{pure text} presentation (or use a support character, like a nurse or lab technician, to briefly report the results via dialogue).
\item \textbf{Requirement:} Please aim for \textbf{>50\% of the plot to involve dialogue} to increase the immersive nature of the game.
\end{itemize}

\textbf{\#\# 5. Character Design Constraints}
\begin{itemize}[noitemsep, topsep=2pt, leftmargin=*]
\item To simplify the burden on our video generation module and system design, try to ensure the patient appears alone (minimize the presence of family members/escorts unless necessary).
\item In scenarios where the patient cannot appear or speak (e.g., coma, physical distress, or surgery), use third-party characters (e.g., other doctors or nurses) to convey the narrative.
\end{itemize}

\textbf{\#\# 6. Communication Constraints (Face-to-Face Only)}
\begin{itemize}[noitemsep, topsep=2pt, leftmargin=*]
\item \textbf{No Phone Calls:} Do not depict the player making or receiving phone calls to discuss patient conditions.
\item \textbf{Physical Interaction:} All consultations between the player and other specialists must occur \textbf{in person}.
\item \textbf{Staging:} If a consultation is necessary, you must arrange for the player to visit the other specialist's office (change of scene) or for the specialist to enter the player's office to discuss the case.
\end{itemize}

\textbf{\#\# 7. Output Format Constraints (CRITICAL)}
\begin{itemize}[noitemsep, topsep=2pt, leftmargin=*]
\item \textbf{JSON Only:} Output the result as a raw JSON object. Do NOT wrap it in markdown code blocks. Do NOT add any conversational text before or after the JSON.
\item \textbf{Language:} Ensure the content (dialogue, options, explanations) matches the language of the input summary (or the target language if specified), while keeping all JSON keys strictly in English as defined in the Pydantic model.
\end{itemize}

\textbf{\#\# 8. Scene and Character Requirements}

We provide you a list of scenes and characters that you can \textbf{only use} to design the plots.

\textbf{\#\#\# Scene}

\textbf{\textcolor{RoyalBlue}{\{Scenes\_reference\}}} \hfill $\rightarrow$ \textit{See Section~\ref{appendix section: Details of Narrative Elements}}

\textbf{\#\#\# Characters}

\textbf{\textcolor{RoyalBlue}{\{Characters\_reference\}}} \hfill $\rightarrow$ \textit{See Section~\ref{appendix section: Details of Narrative Elements}}

\noindent\rule{\linewidth}{0.4pt}

Now, based on the examples and requirements above, please convert the following \textbf{\textcolor{RoyalBlue}{\{input\_type\}}} into a structured (json format) medical gamified clinical story:

\textbf{\textcolor{RoyalBlue}{\{input\_patient\_summary\}}}

\end{tcolorbox}
\caption{Complete prompt $\Phi_{\text{des}}$ for Medical Narrative Designer.}
\label{fig:prompt_phi_des}
\end{figure*}

\subsection{The Instruction used for Story Director: $\Phi_{\text{dir}}$}
\label{appendix section: prompt for story director}

The instruction $\Phi_{\text{dir}}$ guides the LLM to decompose the Clinical Storyline into executable multimodal generation tasks. Given the structured script from the Medical Narrative Designer, the Story Director generates specific execution parameters for each visual asset. The complete prompt template is shown in Figure~\ref{fig:prompt_phi_dir}.

\begin{figure*}[p]
\begin{tcolorbox}[title = {Instruction Template: $\Phi_{\text{dir}}$}, 
                  fonttitle=\small, 
                  fontupper=\scriptsize,
                  colback=gray!5!white,
                  colframe=black!75!white]

\textbf{\# Story Plot}

\textbf{\textcolor{RoyalBlue}{\{story\_plot\_content\}}} \hfill $\rightarrow$ \textit{Output from Medical Narrative Designer}

\textbf{\# Available Hospital Character Resources}

\textbf{\textcolor{RoyalBlue}{\{characters\_description\}}} \hfill $\rightarrow$ \textit{See Section~\ref{appendix section: Details of Narrative Elements}}

\textbf{\# Available Scene Resources}

\textbf{\textcolor{RoyalBlue}{\{scenes\_description\}}} \hfill $\rightarrow$ \textit{See Section~\ref{appendix section: Details of Narrative Elements}}

\textbf{\# Available Image Generation APIs}

\begin{enumerate}[noitemsep, topsep=2pt, leftmargin=*]
\item \textbf{character\_gen}: Generate character portrait with pure white background (profile photo). Used for creating patient images.
\item \textbf{fusion}: Fuse one or multiple person images into a scene at specified positions. Supports both single-person and multi-person scenarios.
\item \textbf{modification}: Modify the state of person(s) in an existing image (expression, posture, clothing, etc.). Can reference previous task outputs via \texttt{[task\_XXX\_output]} placeholder.
\end{enumerate}

\textbf{\# Task List}

There are \textbf{\textcolor{RoyalBlue}{\{N\}}} image generation tasks. Based on the plot, visual continuity, and dialogue characters, determine the appropriate API type and parameters for each task.

\textbf{\textcolor{RoyalBlue}{\{task\_list\_with\_locations\_and\_content\}}}

\textbf{\# Generation Guidelines}

\textbf{\#\# 1. Path Placeholders}
\begin{itemize}[noitemsep, topsep=2pt, leftmargin=*]
\item Character paths: \texttt{[character\_source\_dir]/Nurse\_1.png}
\item Scene paths: \texttt{[scene\_source\_dir]/consultation\_room\_1365x768.png}
\item Task dependencies: \texttt{[task\_XXX\_output]} references the output of task XXX
\end{itemize}

\textbf{\#\# 2. Character Profile Generation Rules}
\begin{itemize}[noitemsep, topsep=2pt, leftmargin=*]
\item \textbf{Same Patient Evolution}: Character\_Profile in different Acts represents the same patient at different time points.
\item \textbf{First Act}: Use \texttt{character\_gen} to generate the patient's initial profile from scratch.
\item \textbf{Subsequent Acts}: Use \texttt{modification} to update the previous Act's profile via \texttt{[task\_XXX\_output]} placeholder, reflecting disease progression, treatment effects, or clothing changes.
\end{itemize}

\textbf{\#\# 3. First-Person Perspective Rules}
\begin{itemize}[noitemsep, topsep=2pt, leftmargin=*]
\item \textbf{Core Rule}: The Doctor (Player) can NEVER appear in the same frame with other characters.
\item \textbf{Dialogue Scenes}: Use first-person view showing only non-player characters (patient, specialists). Even when only the Doctor speaks, the frame shows listeners in an attentive state.
\item \textbf{Doctor\_Thinking}: Camera looks at the doctor alone in the frame.
\end{itemize}

\textbf{\#\# 4. API Selection by Character Count}
\begin{itemize}[noitemsep, topsep=2pt, leftmargin=*]
\item \textbf{Single character}: Use \texttt{fusion} (single-person) or \texttt{modification}.
\item \textbf{Multiple characters}: Use \texttt{fusion} or \texttt{modification} with LIST parameters (all lists must have equal length).
\end{itemize}

\textbf{\#\# 5. Fusion vs Modification Strategy}
\begin{itemize}[noitemsep, topsep=2pt, leftmargin=*]
\item \textbf{Prioritize fusion} for story images to ensure high quality.
\item \textbf{Avoid long modification chains} as they cause severe image quality degradation.
\item Use \texttt{modification} primarily for Character\_Profile evolution across Acts.
\end{itemize}

\textbf{\#\# 6--13. Additional Guidelines}

Additional rules cover: image continuity, description quality (visual elements only, ``Mouth closed.''), Doctor\_Thinking posture diversity, parameter completeness, correct key parameter formatting, scene adaptability, Act-based clothing changes, and medical staff expression correction.

\noindent\rule{\linewidth}{0.4pt}

\textbf{\# Output Format}

Return a JSON object with task\_id as key and corresponding parameters as value:

\begin{verbatim}
{
  "task_001": {
    "type": "character_gen|fusion|modification",
    "description": "Brief English description",
    "characters_in_image": ["patient"],
    "params": { ... }
  },
  "task_002": { ... }
}
\end{verbatim}

\end{tcolorbox}
\caption{Complete prompt $\Phi_{\text{dir}}$ for Story Director.}
\label{fig:prompt_phi_dir}
\end{figure*}

\subsection{Task Dependency}
\label{appendix section: task dependency}

The Story Director organizes all generation tasks into a Directed Acyclic Graph (DAG) $\mathcal{G} = (V, E)$ to manage inter-task dependencies and ensure execution correctness.

\subsubsection{Dependency Types}

We identify two primary categories of dependencies between tasks:

\paragraph{Identity Propagation.} The most critical dependency type ensures character identity consistency across the narrative. When a patient first appears in Act 1, the \texttt{character\_gen} API creates their canonical profile (e.g., \texttt{task\_001}). All subsequent appearances of this patient---whether in later Acts or different scenes---must reference this canonical identity. For example:
\begin{itemize}[noitemsep, topsep=2pt, leftmargin=*]
\item \texttt{task\_001} (Act 1 Character\_Profile): \texttt{character\_gen} $\rightarrow$ generates initial patient portrait
\item \texttt{task\_015} (Act 2 Character\_Profile): \texttt{modification} with \texttt{input\_image\_path: [task\_001\_output]} $\rightarrow$ updates appearance while preserving identity
\item \texttt{task\_030} (Act 3 Character\_Profile): \texttt{modification} with \texttt{input\_image\_path: [task\_015\_output]} $\rightarrow$ continues evolution chain
\end{itemize}
This concern for consistent person representation across evolving multimodal frames aligns with broader multimodal face modeling that studies identity-preserving analysis under diverse visual conditions~\cite{DBLP:conf/aaai/MaLXXY26,DBLP:conf/aaai/WangSTGZLFYYC26}.

\paragraph{Scene Continuity.} Within a single scene containing multiple visual segments (e.g., a dialogue with several exchanges), subsequent frames may depend on earlier ones to maintain visual coherence. This is particularly important for modification-based transitions where character posture or expression changes incrementally.

\subsubsection{Dependency Resolution}

The system resolves dependencies through a two-phase process:

\paragraph{Phase 1: Placeholder Injection.} During parameter generation, the LLM uses symbolic placeholders (e.g., \texttt{[task\_001\_output]}) instead of concrete file paths. This allows the model to express dependencies without knowing the actual output locations.

\paragraph{Phase 2: Topological Execution.} Before execution, the system:
\begin{enumerate}[noitemsep, topsep=2pt, leftmargin=*]
\item Parses all \texttt{[task\_XXX\_output]} references to construct the dependency graph
\item Performs topological sorting to determine a valid execution order
\item Validates that no circular dependencies exist
\item Executes tasks in order, resolving placeholders to actual file paths as upstream tasks complete
\end{enumerate}

\subsubsection{Dependency Graph Metrics}

For quality assurance, we compute several metrics on the generated dependency graphs:
\begin{itemize}[noitemsep, topsep=2pt, leftmargin=*]
\item \textbf{Root Tasks}: Tasks with no dependencies (typically Act 1 Character\_Profile and initial scene setups)
\item \textbf{Graph Depth}: The longest dependency chain, indicating the maximum sequential execution path
\item \textbf{Invalid References}: Dependencies pointing to non-existent tasks, which indicate generation errors
\item \textbf{Circular Dependencies}: Cycles in the graph that would prevent valid execution ordering
\end{itemize}

These metrics are used in our evaluation of Story Direction quality (Section~\ref{appendix subsection: Evaluation of Story Direction}).

\section{Generalizing Linear Clinical Storylines to Branching Story Graphs}
\label{appendix section: branching generalization}

The linear clinical storyline studied in the main paper can be naturally generalized to a branching clinical story graph. Instead of representing a case as a single sequence of state transitions, we can define a directed story graph
\begin{equation}
\mathcal{G}_{\text{story}} = (\mathcal{C}, \mathcal{E}_{\text{story}}),
\end{equation}
where $\mathcal{C}$ denotes the set of clinical states and each edge $e \in \mathcal{E}_{\text{story}}$ represents a decision-conditioned transition between two states:
\begin{equation}
e = (\mathcal{C}_i, a, \mathcal{C}_j).
\end{equation}
Here, $a$ denotes a learner action or selected option at a decision node. Under this view, the linear formulation in Eq.~\ref{eq:linear-story} is a special case in which each decision node has one designated successor state along the generated learning trajectory.

Branching can be incorporated by allowing different options or externally specified branch states to point to different successor states. For example, one branch may continue along the factual clinical trajectory described in the source patient summary, while another branch may instantiate a controlled counterfactual continuation under explicit medical and structural constraints. This view preserves the same state-transition abstraction used in the main paper: Acts and Scenes organize local regions of the story graph, Decision Nodes define decision-conditioned outgoing transitions, and the Story Director can consume the resulting graph by planning multimodal generation tasks for each reachable state and dependency edge.

At the Act level, this extension can be constructed through prefix-conditioned continuation. Let
\begin{equation}
\mathbf{A}_{1:k}=\langle A_1,\ldots,A_k\rangle
\end{equation}
denote a shared prefix of already generated Acts, and let $\mathcal{B}_k=\{b_1,\ldots,b_M\}$ denote a set of branch states describing alternative clinical states entering the next Act. Given the source patient summary $S_{\text{pat}}$, the shared prefix $\mathbf{A}_{1:k}$, and a branch state $b_m$, the Medical Narrative Designer can generate a branch-specific successor Act:
\begin{equation}
A_{k+1}^{(m)} = f_{\text{des}}\left(S_{\text{pat}}, \mathbf{A}_{1:k}, b_m \mid \Phi_{\text{des}}\right).
\end{equation}
Each branch-specific storyline variant is then formed as
\begin{equation}
\mathcal{T}^{(m)}=\langle A_1,\ldots,A_k,A_{k+1}^{(m)}\rangle .
\end{equation}
Figure~\ref{fig:act_branching_generalization} illustrates this Act-level view.

\begin{figure}[t]
\centering
\scriptsize
\resizebox{\linewidth}{!}{%
\begin{tikzpicture}[
    node distance=0.85cm and 0.85cm,
    act/.style={draw, rounded corners, align=center, minimum width=2.0cm, minimum height=0.55cm, inner sep=3pt, fill=blue!6},
    sharedprefix/.style={draw, rounded corners, align=center, minimum width=2.0cm, minimum height=0.55cm, inner sep=3pt, fill=green!8},
    branch/.style={draw, rounded corners, align=center, minimum width=2.05cm, minimum height=0.55cm, inner sep=3pt, fill=orange!10},
    arrow/.style={-{Latex[length=2mm]}, thick}
]
\node[act] (act1) {Act $1$};
\node[sharedprefix, right=of act1] (act2) {Act $2$};
\node[right=of act2] (dots) {$\cdots$};
\node[sharedprefix, right=of dots] (actk) {Act $k$};

\node[branch, above right=0.35cm and 1.05cm of actk] (branch1) {Act $k{+}1$\\branch $1$};
\node[branch, right=1.05cm of actk] (branch2) {Act $k{+}1$\\branch $2$};
\node[branch, below right=0.35cm and 1.05cm of actk] (branchm) {Act $k{+}1$\\branch $M$};

\draw[arrow] (act1) -- (act2);
\draw[arrow] (act2) -- (dots);
\draw[arrow] (dots) -- (actk);
\draw[arrow] (actk) -- node[above, sloped] {$b_1$} (branch1);
\draw[arrow] (actk) -- node[above] {$b_2$} (branch2);
\draw[arrow] (actk) -- node[below, sloped] {$b_M$} (branchm);
\end{tikzpicture}
}
\caption{Act-level branching as prefix-conditioned continuation. A shared Act prefix $\mathbf{A}_{1:k}$ can be extended into multiple branch-specific successor Acts by conditioning generation on alternative branch states $b_1,\ldots,b_M$.}
\label{fig:act_branching_generalization}
\end{figure}

\begin{tcolorbox}[
    title={Inference-time Act-prefix branching},
    fontupper=\small,
    colback=gray!5!white,
    colframe=black!50!white,
    left=1mm,
    right=1mm,
    top=1mm,
    bottom=1mm
]
\begin{enumerate}[noitemsep, topsep=2pt, leftmargin=*]
    \item Generate or select a shared Act prefix $\mathbf{A}_{1:k}=\langle A_1,\ldots,A_k\rangle$.
    \item Specify branch states $\mathcal{B}_k=\{b_1,\ldots,b_M\}$, where each $b_m$ describes the clinical state entering the next Act.
    \item For each branch state $b_m$, prompt the Medical Narrative Designer with $S_{\text{pat}}$, $\mathbf{A}_{1:k}$, and $b_m$ to generate only $A_{k+1}^{(m)}$.
    \item Validate the generated successor Act against the Clinical Storyline schema.
    \item Merge $\mathbf{A}_{1:k}$ with each valid $A_{k+1}^{(m)}$ to obtain branch-specific storyline variants.
\end{enumerate}
\end{tcolorbox}

This construction should be interpreted as an inference-time extension rather than as a claim that the model automatically discovers a complete branching policy. The branch states may be manually specified or semi-automatically proposed and should be medically reviewed, especially when they instantiate counterfactual clinical continuations. The key observation is that a model trained for linear storyline generation can still support Act-level branching when it is conditioned on a shared prefix and explicit branch-specific clinical states.

\section{More Details on Dataset}
\label{appendix section: more details about dataset}

\subsection{Clinical Case Source}

MedGame Bench comprises 5,000 clinical cases sourced from the PMC-Patient Dataset~\cite{PMC_Patient}, covering diverse medical domains. To ensure balanced representation, we uniformly sample cases from eight subspecialties, as shown in Figure~\ref{fig:medgame_bench_overview} in the main paper.

\subsection{Reference Trajectory Construction}

For fine-tuning and controlled task evaluation, we construct reference artifacts for the two MedGame tasks. Starting from each sampled patient summary, we use Gemini-3-Pro~\cite{team2023gemini} to produce a structured Clinical Storyline and a corresponding Story Director plan following the formulation and framework described in the main paper. These generated artifacts are used as demonstration trajectories for supervised fine-tuning and as format-compatible references for task construction.

Importantly, these reference trajectories are not treated as direct evidence of clinical correctness. Generated Clinical Storylines and Story Director plans are checked with automated schema and business-logic validation, and model outputs are evaluated using the independent benchmark protocol described in Section~\ref{section: MedGame Bench}. In particular, medical accuracy, educational quality, story adaptation, and task reasonability are assessed by separate evaluation rubrics rather than by exact matching to the reference trajectories.

\subsection{Generated Output Statistics}

\begin{figure}[h]
    \centering
    \includegraphics[width=\linewidth]{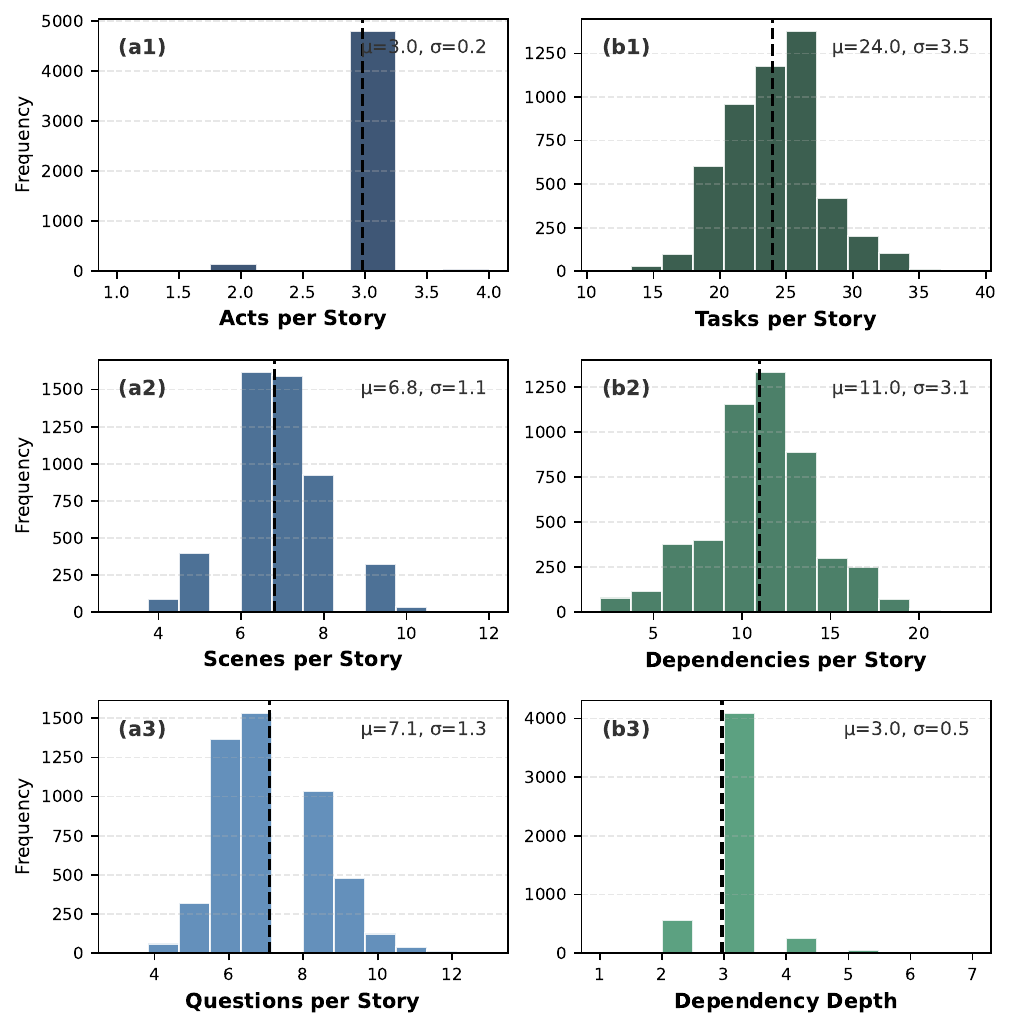}
    \caption{Statistical distributions of generated MedGame reference artifacts (N=5,000). Panels (a1--a3) summarize the hierarchical narrative structure, showing the distributions of acts, scenes, and questions per story. Panels (b1--b3) characterize the technical orchestration complexity, detailing the number of total tasks, inter-task dependencies, and the depth of the dependency graph. Mean ($\mu$) and standard deviation ($\sigma$) are provided for each metric.}
    \label{fig:DataStats}
\end{figure}

Figure~\ref{fig:DataStats} summarizes the statistical properties of the generated reference artifacts, including the distributions of narrative structure (acts, scenes, and questions) and orchestration complexity (task counts, dependency counts, and graph depth).

\subsection{Medical Narrative Designer Reference Corpus}

The Medical Narrative Designer reference corpus is used for fine-tuning models to transform patient summaries into structured clinical storylines. This subsection provides detailed statistics on token distribution and interaction types.

\paragraph{Token Statistics.}
Table~\ref{tab:token_stats_mnd} summarizes the token statistics for the Medical Narrative Designer reference corpus. It involves converting concise patient summaries (avg. 423 tokens) into structured clinical storylines (avg. 3,888 tokens).

\begin{table}[h]
\centering
\caption{Token statistics for the Medical Narrative Designer reference corpus.}
\label{tab:token_stats_mnd}
\small
\begin{tabular}{lccc}
\toprule
\textbf{Component} & \textbf{Mean} & \textbf{Min} & \textbf{Max} \\
\midrule
System Prompt & 2,771 & -- & -- \\
User Input & 423 & 38 & 2,757 \\
Assistant Output & 3,888 & 1,988 & 5,849 \\
\midrule
\textbf{Total} & \textbf{7,082} & & \\
\bottomrule
\end{tabular}
\end{table}

\paragraph{Interaction Type Distribution.}
Table~\ref{tab:interaction_types} presents the distribution of interaction types across all Decision Nodes in the Medical Narrative Designer reference corpus. The three interaction types serve different pedagogical purposes: \textit{single\_choice} for clinical decision-making assessments, \textit{interactive} for information gathering through patient interviews, and \textit{batch} for multi-selection scenarios such as ordering multiple diagnostic tests.

\begin{table}[h]
\centering
\caption{Distribution of interaction types in the Medical Narrative Designer reference corpus. The corpus contains 35,452 Decision Nodes across 5,000 clinical storylines.}
\label{tab:interaction_types}
\small
\begin{tabular}{lcc}
\toprule
\textbf{Interaction Type} & \textbf{Count} & \textbf{Percentage} \\
\midrule
single\_choice & 15,027 & 42.4\% \\
interactive & 13,249 & 37.4\% \\
batch & 7,176 & 20.2\% \\
\midrule
\textbf{Total} & \textbf{35,452} & \textbf{100\%} \\
\bottomrule
\end{tabular}
\end{table}

\subsection{Story Director Reference Corpus}

The Story Director reference corpus is used for fine-tuning models to orchestrate the generation of visual assets for each clinical storyline. This subsection provides detailed statistics on token distribution, task types, character usage, and scene distribution.

\paragraph{Token Statistics.}
Table~\ref{tab:token_stats_sd} summarizes the token statistics for the Story Director reference corpus. It takes story scripts with resource specifications (avg. 8,501 tokens) and generates multimodal orchestration plans (avg. 3,277 tokens).

\begin{table}[h]
\centering
\caption{Token statistics for the Story Director reference corpus.}
\label{tab:token_stats_sd}
\small
\begin{tabular}{lccc}
\toprule
\textbf{Component} & \textbf{Mean} & \textbf{Min} & \textbf{Max} \\
\midrule
System Prompt & 2,973 & -- & -- \\
User Input & 8,501 & 7,625 & 9,633 \\
Assistant Output & 3,277 & 2,116 & 4,738 \\
\midrule
\textbf{Total} & \textbf{14,751} & & \\
\bottomrule
\end{tabular}
\end{table}

\paragraph{Task Type Distribution.}
Table~\ref{tab:task_type_dist} shows the distribution of image generation task types in the Story Director reference corpus. The corpus comprises three task types: \textit{fusion} tasks that composite character images onto scene backgrounds, \textit{modification} tasks that apply visual transformations to existing images, and \textit{character\_gen} tasks that generate new character images based on textual descriptions.

\begin{table}[h]
\centering
\caption{Distribution of task types in the Story Director reference corpus. Fusion tasks dominate as they are required for generating each scene frame.}
\label{tab:task_type_dist}
\small
\begin{tabular}{lcc}
\toprule
\textbf{Task Type} & \textbf{Count} & \textbf{Percentage} \\
\midrule
fusion & 102,260 & 85.3\% \\
modification & 12,602 & 10.5\% \\
character\_gen & 5,024 & 4.2\% \\
\midrule
\textbf{Total} & \textbf{119,886} & \textbf{100\%} \\
\bottomrule
\end{tabular}
\end{table}

\paragraph{Character Usage.}
Table~\ref{tab:character_usage} summarizes the usage of hospital character types across all storylines. The \textit{patient} and \textit{doctor} characters appear in virtually every story, while supporting medical staff such as nurses, radiologists, and pathologists appear based on the clinical workflow requirements of each case.

\begin{table}[h]
\centering
\caption{Hospital character type usage in the Story Director reference corpus. ``Stories'' indicates the number of storylines featuring each character type, and ``Total Usage'' counts all appearances across image generation tasks.}
\label{tab:character_usage}
\small
\resizebox{\linewidth}{!}{%
\begin{tabular}{lcccc}
\toprule
\textbf{Character Type} & \textbf{Stories} & \textbf{\% Stories} & \textbf{Total Usage} & \textbf{Avg/Story} \\
\midrule
patient & 5,000 & 100.0\% & 58,932 & 11.8 \\
doctor & 4,998 & 100.0\% & 37,905 & 7.6 \\
Nurse & 3,172 & 63.4\% & 14,539 & 4.6 \\
Radiologist & 3,018 & 60.4\% & 8,852 & 2.9 \\
lab\_physician & 1,827 & 36.5\% & 4,407 & 2.4 \\
Pathologist & 1,525 & 30.5\% & 4,813 & 3.2 \\
\bottomrule
\end{tabular}%
}
\end{table}

\paragraph{Scene Distribution.}
Table~\ref{tab:scene_dist} presents the distribution of hospital scenes used in the Story Director reference corpus. The consultation room is the most frequently used scene, reflecting the central role of doctor-patient consultations in clinical workflows. Other scenes such as general wards, radiology rooms, and pathology labs are used based on the diagnostic and treatment requirements of each case.

\begin{table}[h]
\centering
\caption{Distribution of hospital scenes in the Story Director reference corpus.}
\label{tab:scene_dist}
\small
\begin{tabular}{lcc}
\toprule
\textbf{Scene} & \textbf{Count} & \textbf{Percentage} \\
\midrule
Consultation Room & 59,789 & 58.5\% \\
General Ward & 12,200 & 11.9\% \\
Radiology Room & 12,069 & 11.8\% \\
Pathology Room & 6,589 & 6.4\% \\
ICU & 6,344 & 6.2\% \\
Examination Room & 5,231 & 5.1\% \\
\midrule
\textbf{Total} & \textbf{102,222} & \textbf{100\%} \\
\bottomrule
\end{tabular}
\end{table}

\section{More Details on Evaluation}
\label{appendix section: Evaluation Details}

This section provides comprehensive details of our evaluation methodology for both the Medical Narrative Designer and the Story Director components. Our evaluation pipeline consists of two stages: (1) \textbf{Automated Structure Validation} to verify format compliance and business logic constraints, and (2) \textbf{LLM-as-a-Judge} evaluation for content quality assessment using GPT-5.2 as the evaluator.

\subsection{Student Perception Questionnaire}
\label{appendix subsection: student perception questionnaire}

For the learner-facing pilot study in RQ4, eight senior medical students each evaluated the same five clinical cases under three presentation modes: the original case material, a text-only MedGame storyline, and a multimodal MedGame simulation. After each condition, students completed the questionnaire in Table~\ref{tab:student_perception_questionnaire} using a 1--5 Likert scale, where 1 indicates strong disagreement and 5 indicates strong agreement. The Low Cognitive Load item is phrased so that higher scores consistently indicate a better perceived experience.

\begin{table}[htbp]
\centering
\small
\caption{Questionnaire used in the student perception study.}
\label{tab:student_perception_questionnaire}
\resizebox{\linewidth}{!}{
\begin{tabular}{lp{0.68\linewidth}}
\toprule
Dimension & Questionnaire Item \\
\midrule
Engagement & This presentation mode kept me engaged with the clinical case. \\
Presence & This presentation mode made me feel immersed in the clinical scenario. \\
Clinical Clarity & This presentation mode helped me clearly understand the clinical information and reasoning process. \\
Perceived Usefulness & This presentation mode would be useful for learning or reviewing clinical cases. \\
Low Cognitive Load & This presentation mode was easy to follow and did not impose unnecessary cognitive burden. \\
\bottomrule
\end{tabular}
}
\end{table}

After completing the ratings, students were also invited to provide brief comments on which presentation mode they preferred and why. These comments were used only to contextualize the quantitative perception scores.

\subsection{Evaluation of Medical Narrative Designer}
\label{appendix subsection: Evaluation of Medical Narrative Designer}

\subsubsection{Structure Validation Pipeline}

The structure validation for Medical Narrative Designer employs a three-layer verification system:

\paragraph{Layer 1: JSON Parsing.} We first verify that the model's output is valid JSON. Responses wrapped in \texttt{```json} blocks or containing \texttt{<think>...</think>} tags (from reasoning models) are preprocessed before parsing.

\paragraph{Layer 2: Schema Validation.} We validate the parsed JSON against the \texttt{Clinical\_Story\_Tree} Pydantic schema (Listing~\ref{list: pydantic_models of clinical story}). This ensures all required fields are present with correct types, including the hierarchical structure of Acts, Scenes, and Decision Nodes.

\paragraph{Layer 3: Business Logic Validation.} We apply strict business rules specific to our educational game design:
\begin{itemize}[noitemsep, topsep=2pt, leftmargin=*]
    \item \textbf{Single-Choice Questions}: Must have exactly one correct answer (\texttt{is\_correct = True}).
    \item \textbf{Batch Questions}: Must have at least one correct answer.
    \item \textbf{Interactive Questions}: All options must have valid \texttt{interaction\_type} values.
\end{itemize}

\subsubsection{LLM-as-a-Judge Evaluation Protocol}

For content quality assessment, we employ GPT-5.2 as an LLM-as-a-Judge evaluator. The evaluation is structured into \textbf{three independent dimensions}, each assessed via a separate API call to ensure focused and unbiased evaluation. This design prevents cross-contamination between dimensions and allows the judge to provide detailed reasoning for each aspect.

\paragraph{Evaluation Dimensions and Indicators.}
Table~\ref{tab:storygen_indicators} summarizes the eight evaluation indicators across three dimensions.

\begin{table}[h]
\centering
\caption{Evaluation indicators for Medical Narrative Designer. Each indicator is scored on a 1-10 scale by GPT-5.2.}
\label{tab:storygen_indicators}
\resizebox{\linewidth}{!}{
\begin{tabular}{l p{7cm}}
\toprule
\textbf{Indicator} & \textbf{Evaluation Focus} \\
\midrule
\multicolumn{2}{l}{\textit{Dimension 1: Story Adaptation Quality}} \\
\cmidrule{1-2}
CCI & Integration of key clinical information from the patient summary into the story plot \\
CSU & Compliance with and appropriate usage of predefined characters and scenes \\
NQ & Narrative coherence, immersion, and smoothness of plot transitions \\
\midrule
\multicolumn{2}{l}{\textit{Dimension 2: Medical Accuracy}} \\
\cmidrule{1-2}
CDA & Correctness of diagnostic reasoning, examination choices, and treatment plans \\
ODA & Accuracy of \texttt{is\_correct} labels and plausibility of distractor options \\
MEA & Factual accuracy of medical explanations and teaching points \\
\midrule
\multicolumn{2}{l}{\textit{Dimension 3: Educational Value}} \\
\cmidrule{1-2}
QDQ & Question clarity, depth, and ability to stimulate clinical thinking \\
FQ & Educational value of explanations, teaching points, and feedback \\
\bottomrule
\end{tabular}
}
\end{table}

\subsubsection{Detailed Indicator Definitions}

Below we provide detailed definitions and scoring criteria for each indicator.

\paragraph{Clinical Content Integration (CCI).}
This indicator evaluates whether key clinical information from the Patient Summary—including chief complaint, medical history, examinations, diagnosis, and treatment—is completely and naturally integrated into the story plot and interactive questions.

\textbf{Important Consideration}: To create smoother narratives, the LLM is allowed to add supplementary clinical content (e.g., additional examinations, intermediate findings) that is not present in the original Patient Summary. These creative additions are acceptable and should not be penalized, as long as they: (1) do not contradict the original clinical content, (2) do not alter the diagnostic reasoning or final diagnosis, and (3) do not change the treatment approach.

\textit{Scoring Priority}: Focus primarily on the integration of original key clinical information. Creative additions are a secondary consideration—they can enhance a high score or reduce it if problematic.

\paragraph{Character \& Scene Usage (CSU).}
This indicator evaluates two aspects: (1) \textbf{Compliance}—whether all used characters and scenes come from the predefined reference set ($\mathcal{P}$ and $\mathcal{L}$), and (2) \textbf{Reasonability}—whether character roles and interactions follow clinical logic (e.g., pathologists deliver diagnostic evidence, nurses manage bedside care).

\paragraph{Narrative Quality (NQ).}
This indicator evaluates the story's coherence, immersion, naturalness of \texttt{start\_role\_play} and \texttt{end\_role\_play} segments, and smoothness of plot transitions between scenes.

\paragraph{Clinical Decision Accuracy (CDA).}
This indicator evaluates whether diagnostic reasoning, examination choices, and treatment plans comply with medical standards and clinical guidelines. The evaluator assesses decisions as a medical professional would, checking for evidence-based practices.

\paragraph{Option Design Accuracy (ODA).}
This indicator evaluates two aspects: (1) whether \texttt{is\_correct} labels accurately identify the ground-truth answer(s), and (2) whether incorrect options are clinically plausible distractors representing common misconceptions, rather than obviously absurd alternatives.

\paragraph{Medical Explanation Accuracy (MEA).}
This indicator evaluates the factual accuracy of medical content in \texttt{explanation}, \texttt{teaching\_point}, and \texttt{overall\_explanation} fields. The evaluator checks whether explanations align with current medical knowledge and guidelines.

\paragraph{Question Design Quality (QDQ).}
This indicator evaluates whether questions and options are clear, deep, and can stimulate clinical thinking. Additionally, the evaluator considers \textbf{information timing}: whether sufficient context has been provided for meaningful engagement, and whether answers have been prematurely revealed in preceding content.

\paragraph{Feedback Quality (FQ).}
This indicator evaluates the combined educational value of \texttt{explanation} + \texttt{teaching\_point} + \texttt{overall\_explanation}. High-quality feedback should be memorable, clinically actionable, and reinforce key learning objectives.

\subsubsection{Complete Evaluation Prompts}

The complete prompts used for LLM-as-a-Judge evaluation are shown in Figures~\ref{fig:storygen_prompt_d1}, \ref{fig:storygen_prompt_d2}, and \ref{fig:storygen_prompt_d3}.

\begin{figure*}[p]
\begin{tcolorbox}[title = {Dimension 1 Prompt: Story Adaptation Quality (Complete)}, 
                  fonttitle=\small, 
                  fontupper=\scriptsize,
                  colback=gray!5!white,
                  colframe=black!75!white]

You are an expert evaluator for medical education content. Your task is to evaluate the \textbf{Story Adaptation Quality} of a generated Clinical Story Tree.

\textbf{Important}: Please evaluate \textbf{objectively, fairly, and meticulously}. Base your scores strictly on the evidence present in the content. Avoid being overly lenient or harsh - assign scores that accurately reflect the quality.

\textbf{\# Evaluation Focus: Story Adaptation Quality}

You will evaluate 3 indicators (each scored 1-10):

\textbf{\#\# 1.1 Clinical Content Integration}

Evaluate whether key clinical information from the Patient Summary (chief complaint, medical history, examinations, diagnosis, treatment) is completely and naturally integrated into the story plot and interactive questions.

\textbf{Important Note on Creative Additions}: To create smoother narratives, the LLM is allowed to add supplementary clinical content (e.g., additional examinations, intermediate findings, supportive procedures) that is NOT present in the original Patient Summary. These creative additions are acceptable and should NOT be penalized, as long as they: (1) Do not contradict the original clinical content; (2) Do not alter the diagnostic reasoning or final diagnosis; (3) Do not change the treatment approach.

\textbf{Scoring Priority}: Focus primarily on the integration of \textbf{original key clinical information}. Creative additions are a secondary consideration.

\textit{Scoring Criteria:}
\begin{itemize}[noitemsep, topsep=0pt, leftmargin=*]
\item 10: Perfect - All original key clinical information is perfectly and naturally integrated; creative additions (if any) enhance the narrative seamlessly
\item 9: Excellent - All original key information is integrated; creative additions are consistent, well-placed, and clinically reasonable
\item 8: Very Good - All main original information is complete; 1-2 minor secondary details slightly simplified; creative additions are appropriate
\item 7: Good - Main original information is basically complete; a few secondary details are missing; creative additions (if any) are acceptable
\item 6: Above Average - Core original information exists, but some important info is missing; OR creative additions are excessive or loosely relevant to the case
\item 5: Average - Core original information is basically covered with noticeable omissions; OR creative additions are medically questionable (though not directly harmful)
\item 4: Below Average - Multiple important pieces of original information are missing; creative additions may distract from the core case
\item 3: Poor - Large amounts of original key information are missing; story barely reflects the original case
\item 2: Very Poor - Severely deviates from the original Patient Summary; OR creative additions contradict the original case
\item 1: Unacceptable - Almost completely ignores or misrepresents the original clinical content
\end{itemize}

\textbf{\#\# 1.2 Character \& Scene Usage}

Evaluate: (1) Whether all used characters/scenes come from the reference (compliance); (2) Whether character roles and interactions follow clinical logic (reasonability).

\textit{Scoring Criteria:}
\begin{itemize}[noitemsep, topsep=0pt, leftmargin=*]
\item 10: Perfect - 100\% compliant with references + all character interactions are highly logical and clinically appropriate
\item 9: Excellent - Fully compliant + character interactions are very reasonable with only trivial improvements possible
\item 8: Very Good - Fully compliant + character interactions are reasonable; minor interaction details could be improved
\item 7: Good - Fully compliant + basically reasonable interactions; some character usage is slightly suboptimal
\item 6: Above Average - Fully compliant, but some character usage feels inappropriate or forced in context
\item 5: Average - Fully compliant, but character interactions lack clinical logic or feel generic
\item 4: Below Average - Minor boundary issues (e.g., ambiguous character/scene usage); interactions are weak
\item 3: Poor - Some characters or scenes appear to violate reference constraints; interactions are illogical
\item 2: Very Poor - Clearly introduces undefined characters or scenes; major compliance violations
\item 1: Unacceptable - Completely disregards the character/scene references; creates arbitrary characters
\end{itemize}

\textbf{\#\# 1.3 Narrative Quality}

Evaluate the story's coherence, immersion, naturalness of \texttt{start\_role\_play}/\texttt{end\_role\_play}, and plot transitions.

\textit{Scoring Criteria:}
\begin{itemize}[noitemsep, topsep=0pt, leftmargin=*]
\item 10: Perfect - Exceptionally smooth plot, seamless transitions, highly immersive narrative throughout
\item 9: Excellent - Very smooth narrative; transitions feel natural; highly engaging with trivial improvements possible
\item 8: Very Good - Coherent narrative; transitions are mostly natural; good immersion with minor rough spots
\item 7: Good - Generally coherent; occasional slightly stiff transitions; maintains reader engagement
\item 6: Above Average - Readable narrative; some transitions feel mechanical; moderate immersion
\item 5: Average - Basically readable, but narrative feels flat; transitions are noticeable
\item 4: Below Average - Plot has some jumps; transitions are often abrupt; limited immersion
\item 3: Poor - Jumping plot; abrupt transitions; difficult to follow the narrative flow
\item 2: Very Poor - Chaotic narrative structure; very difficult to understand
\item 1: Unacceptable - Completely incoherent; no discernible narrative structure
\end{itemize}

\noindent\rule{\linewidth}{0.2pt}

\textbf{\# Input Materials}

\textbf{Patient Summary (Original Input):} \textcolor{RoyalBlue}{\{patient\_summary\}}

\textbf{Characters Reference (Available Characters):} \textcolor{RoyalBlue}{\{characters\_reference\}}

\textbf{Scenes Reference (Available Scenes):} \textcolor{RoyalBlue}{\{scenes\_reference\}}

\textbf{Generated Clinical Story Tree (To Evaluate):} \textcolor{RoyalBlue}{\{generated\_story\}}

\noindent\rule{\linewidth}{0.2pt}

\textbf{\# Output Format}

\textbf{Important}: For each indicator, first provide your reasoning/analysis, then give the score.

\begin{verbatim}
{
  "clinical_content_integration": {
    "reasoning": "<your detailed analysis of how well clinical content is integrated>",
    "score": <1-10>
  },
  "character_scene_usage": {
    "reasoning": "<detailed analysis of character/scene compliance and reasonability>",
    "score": <1-10>
  },
  "narrative_quality": {
    "reasoning": "<detailed analysis of narrative coherence and immersion>",
    "score": <1-10>
  }
}
\end{verbatim}

Please evaluate carefully. Remember: reasoning FIRST, then score.

\end{tcolorbox}
\caption{Complete LLM-as-a-Judge prompt for Dimension 1: Story Adaptation Quality.}
\label{fig:storygen_prompt_d1}
\end{figure*}

\begin{figure*}[p]
\begin{tcolorbox}[title = {Dimension 2 Prompt: Medical Accuracy (Complete)}, 
                  fonttitle=\small, 
                  fontupper=\scriptsize,
                  colback=gray!5!white,
                  colframe=black!75!white]

You are an expert medical professional and educator. Your task is to evaluate the \textbf{Medical Accuracy} of a generated Clinical Story Tree.

\textbf{Important}: Please evaluate \textbf{objectively, fairly, and meticulously}. Base your scores strictly on the medical evidence and clinical guidelines. Avoid being overly lenient or harsh - assign scores that accurately reflect the medical accuracy.

\textbf{Important Context}: The LLM that generated this story was \textbf{required to only use characters and scenes from the provided references}. Therefore, \textbf{DO NOT} consider the choice of characters or scenes as a factor in your scoring. Focus purely on medical accuracy.

\textbf{\# Evaluation Focus: Medical Accuracy}

You will evaluate 3 indicators (each scored 1-10):

\textbf{\#\# 2.1 Clinical Decision Accuracy}

Evaluate whether diagnostic reasoning, examination choices, and treatment plans comply with medical standards and clinical guidelines.

\textit{Scoring Criteria:}
\begin{itemize}[noitemsep, topsep=0pt, leftmargin=*]
\item 10: Perfect - All clinical decisions are textbook-accurate, following the latest guidelines and best practices flawlessly
\item 9: Excellent - All decisions are medically accurate; only trivial alternative approaches might exist
\item 8: Very Good - Decisions are accurate and follow guidelines; 1-2 minor points could be optimized but don't affect outcomes
\item 7: Good - Most decisions are accurate; minor issues exist that don't affect patient safety
\item 6: Above Average - Generally acceptable decisions; some choices are suboptimal but not harmful
\item 5: Average - Some questionable decisions exist, but overall acceptable for educational purposes
\item 4: Below Average - Multiple suboptimal clinical decisions; some may confuse learners
\item 3: Poor - Multiple inaccurate clinical decisions that could mislead learners
\item 2: Very Poor - Serious medical errors present that could harm patients if followed
\item 1: Unacceptable - Dangerous medical misinformation throughout; completely unreliable
\end{itemize}

\textbf{\#\# 2.2 Option Design Accuracy}

Evaluate whether \texttt{is\_correct} labels are accurate + whether incorrect options are clinically plausible distractors (common misconceptions) rather than obviously absurd.

\textit{Scoring Criteria:}
\begin{itemize}[noitemsep, topsep=0pt, leftmargin=*]
\item 10: Perfect - All is\_correct labels are accurate; all distractors represent common clinical misconceptions and are medically plausible
\item 9: Excellent - All labels are accurate; distractors are plausible with trivial improvements possible
\item 8: Very Good - All labels are accurate; most distractors are reasonable and clinically plausible
\item 7: Good - Labels are mostly accurate; distractors are generally reasonable; 1-2 weak options
\item 6: Above Average - Labels mostly accurate; some distractors are too obvious or too obscure
\item 5: Average - Some labeling issues exist; distractor plausibility is inconsistent
\item 4: Below Average - Multiple weak distractors; some labeling accuracy concerns
\item 3: Poor - Multiple incorrect is\_correct labels; many absurd or implausible distractors
\item 2: Very Poor - Severely incorrect labeling; distractors are nonsensical or dangerous
\item 1: Unacceptable - Completely unreliable labeling; misleading content throughout
\end{itemize}

\textbf{\#\# 2.3 Medical Explanation Accuracy}

Evaluate whether medical content in \texttt{explanation}, \texttt{teaching\_point}, and \texttt{overall\_explanation} is accurate.

\textit{Scoring Criteria:}
\begin{itemize}[noitemsep, topsep=0pt, leftmargin=*]
\item 10: Perfect - All medical explanations are impeccably accurate, up-to-date, and align with current guidelines
\item 9: Excellent - All explanations are accurate; only trivial wording improvements possible
\item 8: Very Good - Explanations are accurate; minor imprecisions that don't affect understanding
\item 7: Good - Mostly accurate explanations; 1-2 minor factual imprecisions
\item 6: Above Average - Generally accurate; some explanations could be more precise
\item 5: Average - Some inaccuracies present, but no dangerous misinformation
\item 4: Below Average - Multiple minor factual errors; some explanations are unclear
\item 3: Poor - Multiple factual errors that could confuse learners
\item 2: Very Poor - Seriously inaccurate information; potentially harmful if trusted
\item 1: Unacceptable - Pervasive medical misinformation; dangerous content
\end{itemize}

\noindent\rule{\linewidth}{0.2pt}

\textbf{\# Input Materials}

\textbf{Patient Summary (Original Clinical Case):} \textcolor{RoyalBlue}{\{patient\_summary\}}

\textbf{Characters Reference (for context only):} \textcolor{RoyalBlue}{\{characters\_reference\}}

\textbf{Scenes Reference (for context only):} \textcolor{RoyalBlue}{\{scenes\_reference\}}

\textbf{Generated Clinical Story Tree (To Evaluate):} \textcolor{RoyalBlue}{\{generated\_story\}}

\noindent\rule{\linewidth}{0.2pt}

\textbf{\# Output Format}

\textbf{Important}: (1) For each indicator, first provide your reasoning/analysis, then give the score. (2) DO NOT penalize for character/scene choices.

\begin{verbatim}
{
  "clinical_decision_accuracy": {
    "reasoning": "<detailed analysis of diagnostic reasoning, examination choices, treatment plans>",
    "score": <1-10>
  },
  "option_design_accuracy": {
    "reasoning": "<detailed analysis of is_correct labels and distractor quality>",
    "score": <1-10>
  },
  "medical_explanation_accuracy": {
    "reasoning": "<detailed analysis of medical content accuracy in explanations>",
    "score": <1-10>
  }
}
\end{verbatim}

Please evaluate carefully as a medical expert. Remember: reasoning FIRST, then score.

\end{tcolorbox}
\caption{Complete LLM-as-a-Judge prompt for Dimension 2: Medical Accuracy.}
\label{fig:storygen_prompt_d2}
\end{figure*}

\begin{figure*}[p]
\begin{tcolorbox}[title = {Dimension 3 Prompt: Educational Value (Complete)}, 
                  fonttitle=\small, 
                  fontupper=\scriptsize,
                  colback=gray!5!white,
                  colframe=black!75!white]

You are an expert in medical education and instructional design. Your task is to evaluate the \textbf{Educational Value} of a generated Clinical Story Tree.

\textbf{Important}: Please evaluate \textbf{objectively, fairly, and meticulously}. Base your scores strictly on the educational quality and pedagogical effectiveness. Avoid being overly lenient or harsh - assign scores that accurately reflect the educational value.

\textbf{Important Context}: The LLM that generated this story was \textbf{required to only use characters and scenes from the provided references}. Therefore, \textbf{DO NOT} consider the choice of characters or scenes as a factor in your scoring. Focus purely on educational value.

\textbf{\# Evaluation Focus: Educational Value}

You will evaluate 2 indicators (each scored 1-10):

\textbf{\#\# 3.1 Question Design Quality}

Evaluate whether questions and options are clear, deep, and can stimulate clinical thinking.

\textbf{Additional Consideration - Information Timing}: The story progresses sequentially through acts and scenes. For each question, also consider:
\begin{itemize}[noitemsep, topsep=0pt, leftmargin=*]
\item \textbf{Information Sufficiency}: Has enough information been provided up to that point for the learner to meaningfully engage with the question?
\item \textbf{Answer Obviousness}: Has the answer been too explicitly revealed in preceding content, making the question trivially easy and losing its assessment value?
\end{itemize}

\textit{Scoring Criteria:}
\begin{itemize}[noitemsep, topsep=0pt, leftmargin=*]
\item 10: Perfect - Exceptional questions that deeply promote clinical reasoning; options are expertly crafted to test understanding; timing is ideal with sufficient context and no premature answer revelation
\item 9: Excellent - Outstanding questions with clear learning objectives; highly effective at stimulating critical thinking; appropriate timing
\item 8: Very Good - Strong questions that promote clinical reasoning; well-designed options with minor improvements possible; good timing
\item 7: Good - Good questions with clear learning objectives; effective options; 1-2 questions could be stronger in depth or timing
\item 6: Above Average - Adequate questions; most stimulate some thinking but could be more thought-provoking; minor timing issues possible
\item 5: Average - Passable questions; some lack depth or clarity; moderate educational engagement; occasional timing concerns
\item 4: Below Average - Weak questions; many are superficial or unclear; limited thinking stimulation; some timing issues
\item 3: Poor - Questions are superficial, confusing, or fail to target key learning points; timing issues affect educational value
\item 2: Very Poor - Questions fail to engage learners; options are poorly designed; severe timing or design issues
\item 1: Unacceptable - No meaningful questions; complete failure to educate
\end{itemize}

\textbf{\#\# 3.2 Feedback Quality}

Evaluate the combined educational value of \texttt{explanation} + \texttt{teaching\_point} + \texttt{overall\_explanation}.

\textit{Scoring Criteria:}
\begin{itemize}[noitemsep, topsep=0pt, leftmargin=*]
\item 10: Perfect - Exceptional feedback that teaches brilliantly; teaching points are memorable and clinically actionable
\item 9: Excellent - Outstanding explanations; teaching points are insightful and will stick with learners
\item 8: Very Good - Strong feedback with useful teaching points; minor improvements to memorability possible
\item 7: Good - Good explanations with useful teaching points; effectively reinforces learning
\item 6: Above Average - Adequate feedback; teaching points exist but lack depth or memorability
\item 5: Average - Passable feedback; somewhat generic; moderate educational value
\item 4: Below Average - Weak feedback; often generic or fails to add value beyond the question itself
\item 3: Poor - Feedback is generic, unhelpful, or misses key teaching opportunities
\item 2: Very Poor - Minimal or misleading feedback; teaching points are absent or wrong
\item 1: Unacceptable - No meaningful feedback; zero educational value
\end{itemize}

\noindent\rule{\linewidth}{0.2pt}

\textbf{\# Input Materials}

\textbf{Patient Summary (Original Clinical Case):} \textcolor{RoyalBlue}{\{patient\_summary\}}

\textbf{Characters Reference (for context only):} \textcolor{RoyalBlue}{\{characters\_reference\}}

\textbf{Scenes Reference (for context only):} \textcolor{RoyalBlue}{\{scenes\_reference\}}

\textbf{Generated Clinical Story Tree (To Evaluate):} \textcolor{RoyalBlue}{\{generated\_story\}}

\noindent\rule{\linewidth}{0.2pt}

\textbf{\# Output Format}

\textbf{Important}: (1) For each indicator, first provide your reasoning/analysis, then give the score. (2) DO NOT penalize for character/scene choices.

\begin{verbatim}
{
  "question_design_quality": {
    "reasoning": "<detailed analysis of question clarity, depth, and ability to stimulate thinking>",
    "score": <1-10>
  },
  "feedback_quality": {
    "reasoning": "<detailed analysis of explanation, teaching_point, and overall_explanation quality>",
    "score": <1-10>
  }
}
\end{verbatim}

Please evaluate carefully as an education expert. Remember: reasoning FIRST, then score.

\end{tcolorbox}
\caption{Complete LLM-as-a-Judge prompt for Dimension 3: Educational Value.}
\label{fig:storygen_prompt_d3}
\end{figure*}

\subsection{Evaluation of Story Direction}
\label{appendix subsection: Evaluation of Story Direction}

\subsubsection{Structure Validation Pipeline}

The structure validation for Story Direction employs a comprehensive multi-layer verification system that checks format compliance, API schema adherence, resource path validity, and dependency graph integrity.

\paragraph{Layer 1: JSON Parsing.}
Similar to Medical Narrative Designer, we first verify that the model's output is valid JSON, with preprocessing for reasoning model outputs.

\paragraph{Layer 2: Task-Level Schema Validation.}
Each task in the generated output is validated against the API schema:
\begin{itemize}[noitemsep, topsep=2pt, leftmargin=*]
    \item \textbf{Required Task Fields}: Every task must include \texttt{type}, \texttt{description}, \texttt{characters\_in\_image}, and \texttt{params}.
    \item \textbf{Valid API Types}: The \texttt{type} field must be one of: \texttt{character\_gen}, \texttt{fusion}, or \texttt{modification}.
    \item \textbf{API-Specific Parameters}: Each API type has its own required parameters (Table~\ref{tab:api_required_params}).
\end{itemize}

\begin{table}[h]
\centering
\caption{Required parameters for each API type in Story Direction.}
\label{tab:api_required_params}
\resizebox{\linewidth}{!}{
\begin{tabular}{ll}
\toprule
\textbf{API Type} & \textbf{Required Parameters} \\
\midrule
character\_gen & age, ethnicity, gender, appearance, expression, clothing \\
fusion & person\_image\_path, scene\_image\_path, location\_description, posture\_expression \\
modification & input\_image\_path, modification\_target, modification\_details \\
\bottomrule
\end{tabular}
}
\end{table}

\paragraph{Layer 3: Resource Path Validation.}
We verify that all resource paths reference valid assets:
\begin{itemize}[noitemsep, topsep=2pt, leftmargin=*]
    \item \textbf{Character Paths}: Must reference files from the predefined character asset library (e.g., \texttt{[character\_source\_dir]/Nurse\_1.png}).
    \item \textbf{Scene Paths}: Must reference files from the predefined scene asset library (e.g., \texttt{[scene\_source\_dir]/\allowbreak ICU\_facing\_doctor.png}).
    \item \textbf{Task References}: Cross-task dependencies (e.g., \texttt{[task\_001\_output]}) must reference existing tasks.
\end{itemize}

\paragraph{Layer 4: Dependency Graph Analysis.}
We construct and analyze the task dependency graph $\mathcal{G} = (V, E)$:
\begin{itemize}[noitemsep, topsep=2pt, leftmargin=*]
    \item \textbf{Connectivity}: All tasks with dependencies should be traceable to root nodes (tasks with no dependencies).
    \item \textbf{Invalid References}: Dependencies pointing to non-existent tasks are flagged as errors.
    \item \textbf{Cycle Detection}: The graph must be acyclic to allow valid topological ordering for execution.
    \item \textbf{Depth Analysis}: We compute the maximum dependency chain depth as a complexity metric.
\end{itemize}

\subsubsection{LLM-as-a-Judge Evaluation Protocol}

Similar to Medical Narrative Designer, we use GPT-5.2 as an LLM-as-a-Judge evaluator with three independent dimensions, each assessed via a separate API call.

\paragraph{Evaluation Dimensions and Indicators.}
Table~\ref{tab:filmgen_indicators} summarizes the three evaluation indicators for Story Direction.

\begin{table}[h]
\centering
\caption{Evaluation indicators for Story Direction. Each indicator is scored on a 1-10 scale by GPT-5.2.}
\label{tab:filmgen_indicators}
\resizebox{\linewidth}{!}{
\begin{tabular}{l p{7cm}}
\toprule
\textbf{Indicator} & \textbf{Evaluation Focus} \\
\midrule
Resource Assignment (RA) & Whether scene and character selections match the plot context appropriately \\
API Type Selection (ATS) & Whether API type choices follow the specified rules (e.g., \texttt{character\_gen} for Act 1 profiles) \\
Parameter Content (PC) & Whether parameter values appropriately reflect the narrative requirements \\
\bottomrule
\end{tabular}
}
\end{table}

\subsubsection{Detailed Indicator Definitions}

\paragraph{Resource Assignment (RA).}
This indicator evaluates whether the scene and character selections are appropriate for each task's plot context.

\textbf{Scene Selection Criteria}:
\begin{itemize}[noitemsep, topsep=2pt, leftmargin=*]
    \item Does the selected scene match the story location? (e.g., ward dialogue $\rightarrow$ \texttt{General\_ward\_*})
    \item Is the scene perspective appropriate? (e.g., doctor thinking $\rightarrow$ \texttt{facing\_doctor})
\end{itemize}

\textbf{Character Selection Criteria}:
\begin{itemize}[noitemsep, topsep=2pt, leftmargin=*]
    \item For Doctor\_Thinking scenes: Only the doctor should be in frame (third-person perspective looking at the doctor).
    \item For Dialogue scenes: Only non-doctor participants should be in frame (first-person perspective from the doctor's viewpoint).
    \item Character IDs should be used consistently throughout the story.
\end{itemize}

\textbf{Important Flexibility Consideration}: Although character descriptions mention typical work locations, characters can appear in different scenes when the plot requires it (e.g., a Radiologist coming to the ICU to discuss imaging results). Only assignments that clearly contradict the plot description should be penalized.

\paragraph{API Type Selection (ATS).}
This indicator evaluates whether API type selections follow the specified rules:

\textbf{Character Profile Rules}:
\begin{itemize}[noitemsep, topsep=2pt, leftmargin=*]
    \item Act 1 Character\_Profile: Must use \texttt{character\_gen} (generate from scratch)
    \item Act 2/3 Character\_Profile: Must use \texttt{modification} with \texttt{input\_image\_path} referencing the previous Act's profile
    \item The modification chain should be: Act 1 $\rightarrow$ Act 2 $\rightarrow$ Act 3 (not Act 1 $\rightarrow$ Act 3 directly)
\end{itemize}

\textbf{Perspective Rules}:
\begin{itemize}[noitemsep, topsep=2pt, leftmargin=*]
    \item Doctor\_Thinking scenes: Camera looks at the doctor alone in frame
    \item Dialogue scenes: Doctor never appears with other characters; use first-person view showing others
\end{itemize}

\textbf{Quality Strategy}:
\begin{itemize}[noitemsep, topsep=2pt, leftmargin=*]
    \item Story/Plot images should prioritize \texttt{fusion} to maintain image quality
    \item Avoid long chains of \texttt{modification} as they cause quality degradation
\end{itemize}

\paragraph{Parameter Content (PC).}
This indicator evaluates whether the content of parameters is appropriate and matches the plot:

\textbf{Key Parameters Evaluated}:
\begin{itemize}[noitemsep, topsep=2pt, leftmargin=*]
    \item \textbf{description}: Should accurately describe visual content; should not contain plot words like ``after'', ``again'', ``returned''
    \item \textbf{posture\_expression}: Should match emotional context; serious scenes require serious/concerned expressions
    \item \textbf{location\_description}: Should be specific position descriptions (e.g., ``sitting on the hospital bed''), not generic placeholders (e.g., ``main\_position'')
    \item \textbf{modification\_target}: Should describe visual features (e.g., ``the elderly man lying in bed''), not identity words (e.g., ``patient'')
    \item \textbf{modification\_details}: Should describe specific changes reflecting plot progression
\end{itemize}

\textbf{Reasonable Adaptation}: Minor prop adaptations are acceptable when the actual scene differs from the story description. For example, if the story mentions ``examination bed'' but the scene only has chairs, describing the patient sitting on a chair is acceptable—unless it violates medical logic or contradicts the patient's physical condition.

\subsubsection{Complete Evaluation Prompts}

The complete prompts used for Story Direction LLM-as-a-Judge evaluation are shown in Figures~\ref{fig:filmgen_prompt_d1}, \ref{fig:filmgen_prompt_d2}, and \ref{fig:filmgen_prompt_d3}.

\begin{figure*}[p]
\begin{tcolorbox}[title = {Story Direction - Dimension 1 Prompt: Resource Assignment (Complete)}, 
                  fonttitle=\small, 
                  fontupper=\scriptsize,
                  colback=gray!5!white,
                  colframe=black!75!white]

You are an expert evaluator for medical educational game image generation. Your task is to evaluate the \textbf{Resource Assignment Reasonability} of the generated image task parameters.

\textbf{Important}: Please evaluate \textbf{objectively, fairly, and meticulously}. Base your scores strictly on the evidence present in the content.

\textbf{IMPORTANT - Camera Perspective Rules (Do NOT penalize for following these rules):}
\begin{itemize}[noitemsep, topsep=0pt, leftmargin=*]
\item \textbf{Doctor\_Thinking scenes}: Third-person perspective looking AT the doctor (player). Only the doctor (player) appears in frame.
\item \textbf{Dialogue/Roleplay scenes}: First-person perspective FROM the doctor (player)'s viewpoint. The doctor (player) NEVER appears in frame—only the conversation partner(s) (patient, nurse, etc.) are shown. This is intentional design, NOT a ``missing participant'' issue.
\end{itemize}

\textbf{IMPORTANT - Character Location Flexibility (Do NOT penalize for this):}

Although the character descriptions mention their typical work locations (e.g., Radiologist $\rightarrow$ Radiology department), \textbf{characters CAN appear in different scenes when the plot requires it}. Example: A Radiologist may come to the ICU to discuss imaging results with the doctor. \textbf{Only penalize} if a character appears in a scene that is clearly inconsistent with the plot description.

\textbf{\# Evaluation Focus: Scene \& Character Assignment Reasonability}

\textbf{Scene Selection:}
\begin{itemize}[noitemsep, topsep=0pt, leftmargin=*]
\item Does the selected scene match the story location? (e.g., ward dialogue $\rightarrow$ General\_ward\_*, lab discussion $\rightarrow$ Examination\_room\_*)
\item Is the scene perspective appropriate? (e.g., doctor thinking $\rightarrow$ facing\_doctor, patient speaking $\rightarrow$ facing\_patient)
\end{itemize}

\textbf{Character Selection:}
\begin{itemize}[noitemsep, topsep=0pt, leftmargin=*]
\item For Doctor\_Thinking: Only doctor in frame (correct)
\item For Dialogue scenes: Only the NON-doctor participants in frame (correct—this is first-person perspective)
\item When the story mentions ``Nurse reports'', is a Nurse character actually in the frame?
\item Are the correct specific character IDs used consistently? (e.g., always using Nurse\_2 for the same nurse throughout)
\end{itemize}

\textit{Scoring Criteria:}
\begin{itemize}[noitemsep, topsep=0pt, leftmargin=*]
\item 10: Perfect - All scene and character assignments perfectly match the plot context
\item 9: Excellent - Nearly all assignments are correct; only trivial improvements possible
\item 8: Very Good - Most assignments are correct; 1-2 minor mismatches that don't affect understanding
\item 7: Good - Generally correct assignments; a few noticeable but non-critical mismatches
\item 6: Above Average - Mostly reasonable; some assignments are suboptimal but acceptable
\item 5: Average - Several questionable assignments; some scenes/characters don't quite fit the plot
\item 4: Below Average - Multiple incorrect assignments; affects the visual narrative coherence
\item 3: Poor - Many assignments are wrong; significant mismatch between plot and visual representation
\item 2: Very Poor - Most assignments are incorrect or nonsensical
\item 1: Unacceptable - Completely random or irrelevant assignments
\end{itemize}

\noindent\rule{\linewidth}{0.2pt}

\textbf{\# Input Materials}

\textbf{Story Plot (Original Story Content):} \textcolor{RoyalBlue}{\{story\_plot\}}

\textbf{Available Scene Resources:} \textcolor{RoyalBlue}{\{scene\_resources\}}

\textbf{Available Character Resources:} \textcolor{RoyalBlue}{\{character\_resources\}}

\textbf{Generated Image Tasks (To Evaluate):} \textcolor{RoyalBlue}{\{generated\_tasks\}}

\noindent\rule{\linewidth}{0.2pt}

\textbf{\# Output Format}

\textbf{Important}: First provide your reasoning/analysis, then give the score.

\begin{verbatim}
{
  "resource_assignment": {
    "reasoning": "<detailed analysis of scene and character assignment quality>",
    "score": <1-10>
  }
}
\end{verbatim}

Please evaluate carefully. Remember: reasoning FIRST, then score.

\end{tcolorbox}
\caption{Complete LLM-as-a-Judge prompt for Story Direction Dimension 1: Resource Assignment.}
\label{fig:filmgen_prompt_d1}
\end{figure*}

\begin{figure*}[p]
\begin{tcolorbox}[title = {Story Direction - Dimension 2 Prompt: API Type Selection (Complete)}, 
                  fonttitle=\small, 
                  fontupper=\scriptsize,
                  colback=gray!5!white,
                  colframe=black!75!white]

You are an expert evaluator for medical educational game image generation. Your task is to evaluate the \textbf{API Type Selection Compliance} of the generated image task parameters.

\textbf{Important}: Please evaluate \textbf{objectively, fairly, and meticulously}. Check whether the API type selections follow the specified rules.

\textbf{\# Available API Types}

The system provides 3 API types for image generation:
\begin{itemize}[noitemsep, topsep=0pt, leftmargin=*]
\item \textbf{character\_gen}: Generate a character portrait with pure white background. Use Case: Create patient's profile photo.
\item \textbf{fusion}: Fuse character(s) into a scene at specified positions. Use Case: Place characters into scene environments; preferred for story/plot images.
\item \textbf{modification}: Modify the state of character(s) in an existing image. Use Case: Update existing character profile or make subtle changes.
\end{itemize}

\textbf{\# Key Rules to Check}

\textbf{1. Character Profile Rules:}
\begin{itemize}[noitemsep, topsep=0pt, leftmargin=*]
\item Act 1 Character\_Profile: MUST use \texttt{character\_gen} (generate from scratch)
\item Act 2/3 Character\_Profile: MUST use \texttt{modification} with \texttt{input\_image\_path} referencing previous Act's profile
\item The modification chain should be: Act 1 $\rightarrow$ Act 2 $\rightarrow$ Act 3 (not Act 1 $\rightarrow$ Act 3 directly)
\end{itemize}

\textbf{2. First-Person Perspective Rules:}
\begin{itemize}[noitemsep, topsep=0pt, leftmargin=*]
\item Doctor\_Thinking scenes: Only doctor in frame, camera looks at doctor
\item Dialogue scenes: Doctor NEVER appears with other characters; use first-person view showing others
\end{itemize}

\textbf{3. Multi-Person Scene Rules:}
\begin{itemize}[noitemsep, topsep=0pt, leftmargin=*]
\item 2+ characters in frame: Must use \texttt{fusion} or \texttt{modification} with LIST parameters
\item Single character: Can use \texttt{fusion} (single) or \texttt{modification}
\end{itemize}

\textbf{4. Fusion vs Modification Strategy:}
\begin{itemize}[noitemsep, topsep=0pt, leftmargin=*]
\item Story/Plot images: Should prioritize \texttt{fusion} to maintain image quality
\item Avoid long chains of \texttt{modification} (causes quality degradation)
\item \texttt{modification} should only be used when plot requires subtle changes to existing image
\end{itemize}

\textit{Scoring Criteria:}
\begin{itemize}[noitemsep, topsep=0pt, leftmargin=*]
\item 10: Perfect - All API type selections strictly follow the rules
\item 9: Excellent - Nearly all selections are correct; only trivial issues
\item 8: Very Good - Most selections follow rules; 1-2 minor violations that don't affect functionality
\item 7: Good - Generally follows rules; a few noticeable violations
\item 6: Above Average - Mostly compliant; some rule violations but results are still usable
\item 5: Average - Several rule violations; some API choices are inappropriate
\item 4: Below Average - Multiple rule violations; affects image generation logic
\item 3: Poor - Many incorrect API type selections; fundamental misunderstanding of rules
\item 2: Very Poor - Most selections violate the rules
\item 1: Unacceptable - Complete disregard for API selection rules
\end{itemize}

\noindent\rule{\linewidth}{0.2pt}

\textbf{\# Input Materials}

\textbf{Task List with Content Types:} \textcolor{RoyalBlue}{\{task\_list\}}

(This shows what type of content each task should generate)

\textbf{Generated Image Tasks (To Evaluate):} \textcolor{RoyalBlue}{\{generated\_tasks\}}

\noindent\rule{\linewidth}{0.2pt}

\textbf{\# Output Format}

\textbf{Important}: First provide your reasoning/analysis, then give the score.

\begin{verbatim}
{
  "api_type_selection": {
    "reasoning": "<detailed analysis of API type selection compliance>",
    "score": <1-10>
  }
}
\end{verbatim}

Please evaluate carefully. Remember: reasoning FIRST, then score.

\end{tcolorbox}
\caption{Complete LLM-as-a-Judge prompt for Story Direction Dimension 2: API Type Selection.}
\label{fig:filmgen_prompt_d2}
\end{figure*}

\begin{figure*}[p]
\begin{tcolorbox}[title = {Story Direction - Dimension 3 Prompt: Parameter Content Quality (Complete)}, 
                  fonttitle=\small, 
                  fontupper=\scriptsize,
                  colback=gray!5!white,
                  colframe=black!75!white]

You are an expert evaluator for medical educational game image generation. Your task is to evaluate the \textbf{Parameter Content Quality} of the generated image task parameters.

\textbf{Important}: Please evaluate \textbf{objectively, fairly, and meticulously}. The auto-checking system has judged the parameter completeness; you only need to evaluate content quality.

\textbf{IMPORTANT - Reasonable Adaptation for Scene Props (Do NOT penalize for this):}

During Clinical Storyline generation, the Medical Narrative Designer does NOT know the exact scene visuals (only functional descriptions like ``consultation room''). During Story Direction, the Story Director sees actual scene details (e.g., ``room contains office chairs''). Therefore, \textbf{minor prop adaptations are acceptable}: If the story says ``patient sits on examination bed'' but the actual scene only has office chairs, describing ``patient sits on office chair'' is a reasonable adaptation.

\textbf{ONLY penalize} if the adaptation violates medical logic or common sense:
\begin{itemize}[noitemsep, topsep=0pt, leftmargin=*]
\item \ding{55} Infant who ``cannot sit independently'' described as ``sitting on chair alone'' (medical violation)
\item \ding{55} Patient ``wheeled in on stretcher, cannot sit up'' described as ``sitting in chair'' (plot contradiction)
\item \ding{51} ``Examination bed'' $\rightarrow$ ``office chair'' for a patient who CAN sit (acceptable prop substitution)
\end{itemize}

\textbf{\# Key Parameters to Evaluate}

\textbf{1. description} (character\_gen, fusion):
\begin{itemize}[noitemsep, topsep=0pt, leftmargin=*]
\item Should accurately describe the visual content of the image
\item Should only contain visual elements, NOT plot words like ``after'', ``again'', ``returned''
\item Should be concise and clear
\end{itemize}

\textbf{2. posture\_expression} (fusion):
\begin{itemize}[noitemsep, topsep=0pt, leftmargin=*]
\item Should match the emotional context of the plot
\item Serious scene $\rightarrow$ serious/concerned expression (NOT smiling)
\item Should explicitly state posture: Sitting/Standing/Leaning, etc.
\item Note: Whether ``mouth closed'' is included does NOT affect scoring
\end{itemize}

\textbf{3. location\_description} (fusion):
\begin{itemize}[noitemsep, topsep=0pt, leftmargin=*]
\item Should be specific position descriptions
\item WRONG: ``main\_position'', ``secondary\_position'', ``left\_side''
\item CORRECT: ``standing in the center of the room'', ``sitting on the hospital bed''
\end{itemize}

\textbf{4. modification\_target} (modification):
\begin{itemize}[noitemsep, topsep=0pt, leftmargin=*]
\item Should describe visual features, NOT identity
\item WRONG: ``patient'', ``doctor'', ``the nurse''
\item CORRECT: ``the elderly man lying in bed'', ``the young woman in hospital gown''
\end{itemize}

\textbf{5. modification\_details} (modification):
\begin{itemize}[noitemsep, topsep=0pt, leftmargin=*]
\item Should describe specific changes needed
\item For Character\_Profile (Act 2/3): Should reflect disease progression (e.g., more tired, paler skin, changed expression)
\end{itemize}

\textbf{6. Doctor\_Thinking Diversity}:
\begin{itemize}[noitemsep, topsep=0pt, leftmargin=*]
\item Doctor's postures should be varied across different thinking scenes
\item NOT always the same ``standing and thinking''
\item Examples: arms crossed, hand on chin, looking at report, leaning forward, etc.
\end{itemize}

\textbf{7. character\_gen parameters} (for Act 1 Character\_Profile only):
\begin{itemize}[noitemsep, topsep=0pt, leftmargin=*]
\item expression: Should match the patient's initial emotional state
\item appearance: Should establish the patient's baseline look
\item clothing: Appropriate attire for initial hospital visit
\item All parameters should be specific and visual, not generic placeholders
\end{itemize}

\textit{Scoring Criteria:}
\begin{itemize}[noitemsep, topsep=0pt, leftmargin=*]
\item 10: Perfect - All parameter contents are perfectly crafted and match the plot
\item 9: Excellent - Nearly all parameters are well-written; only trivial improvements possible
\item 8: Very Good - Most parameters are appropriate; 1-2 minor content issues
\item 7: Good - Generally good content; a few parameters could be improved
\item 6: Above Average - Mostly acceptable; some parameters are generic or slightly off
\item 5: Average - Several content issues; some parameters don't match plot well
\item 4: Below Average - Multiple content problems; affects visual narrative quality
\item 3: Poor - Many parameters have inappropriate content; doesn't reflect the plot
\item 2: Very Poor - Most parameter contents are wrong or generic
\item 1: Unacceptable - Parameter contents are nonsensical or completely mismatched
\end{itemize}

\noindent\rule{\linewidth}{0.2pt}

\textbf{\# Input Materials}

\textbf{Story Plot (Original Story Content):} \textcolor{RoyalBlue}{\{story\_plot\}}

\textbf{Generated Image Tasks (To Evaluate):} \textcolor{RoyalBlue}{\{generated\_tasks\}}

\noindent\rule{\linewidth}{0.2pt}

\textbf{\# Output Format}

\textbf{Important}: First provide your reasoning/analysis, then give the score.

\begin{verbatim}
{
  "parameter_content": {
    "reasoning": "<detailed analysis of parameter content quality>",
    "score": <1-10>
  }
}
\end{verbatim}

Please evaluate carefully. Remember: reasoning FIRST, then score.

\end{tcolorbox}
\caption{Complete LLM-as-a-Judge prompt for Story Direction Dimension 3: Parameter Content Quality.}
\label{fig:filmgen_prompt_d3}
\end{figure*}

\subsection{Human Validation of LLM-as-a-Judge}
\label{appendix subsection: human validation of llm judge}

To assess whether GPT-5.2 provides a reliable signal for expert-facing analysis, we collected human judgments on stratified random samples using the same rubrics defined above. For Medical Narrative Generation, three medical Ph.D.\ holders evaluated the same 50 cases. For Story Direction, two experienced game developers, each with over 150 hours of gameplay experience in narrative interactive games, evaluated the same 50 cases. Evaluators were blinded to model identities and used the same indicator-level rubrics as GPT-5.2. For each sample and indicator, we averaged human scores across evaluators and computed Pearson correlation against the corresponding GPT-5.2 score. The same three medical Ph.D.\ holders who evaluated Medical Narrative Generation outputs also performed the expert-in-the-loop revision study; the game-development evaluators were only involved in Story Direction evaluation. The evaluation involved professional judgment of generated educational content rather than collection of identifiable personal data.

\begin{figure}[htbp]
    \centering
    \includegraphics[width=\linewidth]{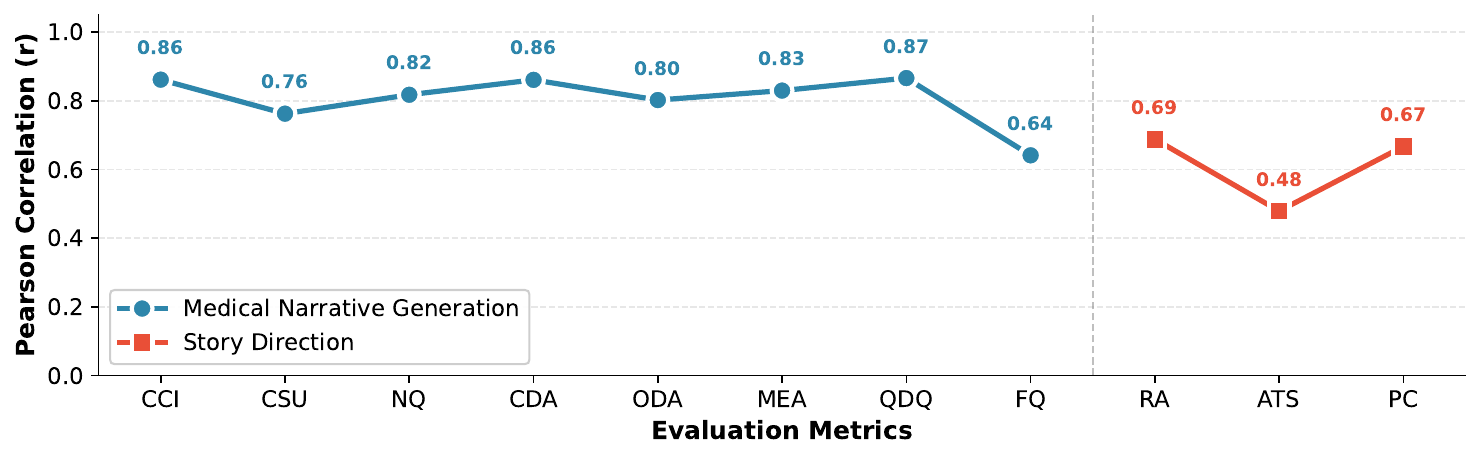}
    \caption{Pearson correlation between GPT-5.2 and human evaluators. Medical Narrative Generation achieves avg.\ $r=0.81$; Story Direction achieves avg.\ $r=0.61$.}
    \label{fig:human_eval}
\end{figure}

Figure~\ref{fig:human_eval} shows strong agreement for Medical Narrative Generation (avg.\ $r=0.81$), suggesting that our LLM-as-a-Judge protocol aligns well with medical experts on narrative, clinical, and educational dimensions. Agreement is moderate for Story Direction (avg.\ $r=0.61$), indicating that orchestration-oriented indicators may require more cautious interpretation or additional human review.

\section{More Experimental Results}

\subsection{Radar Chart Analysis for Medical Narrative Generation}

\begin{figure}[!h]
    \centering
    \includegraphics[width=\linewidth]{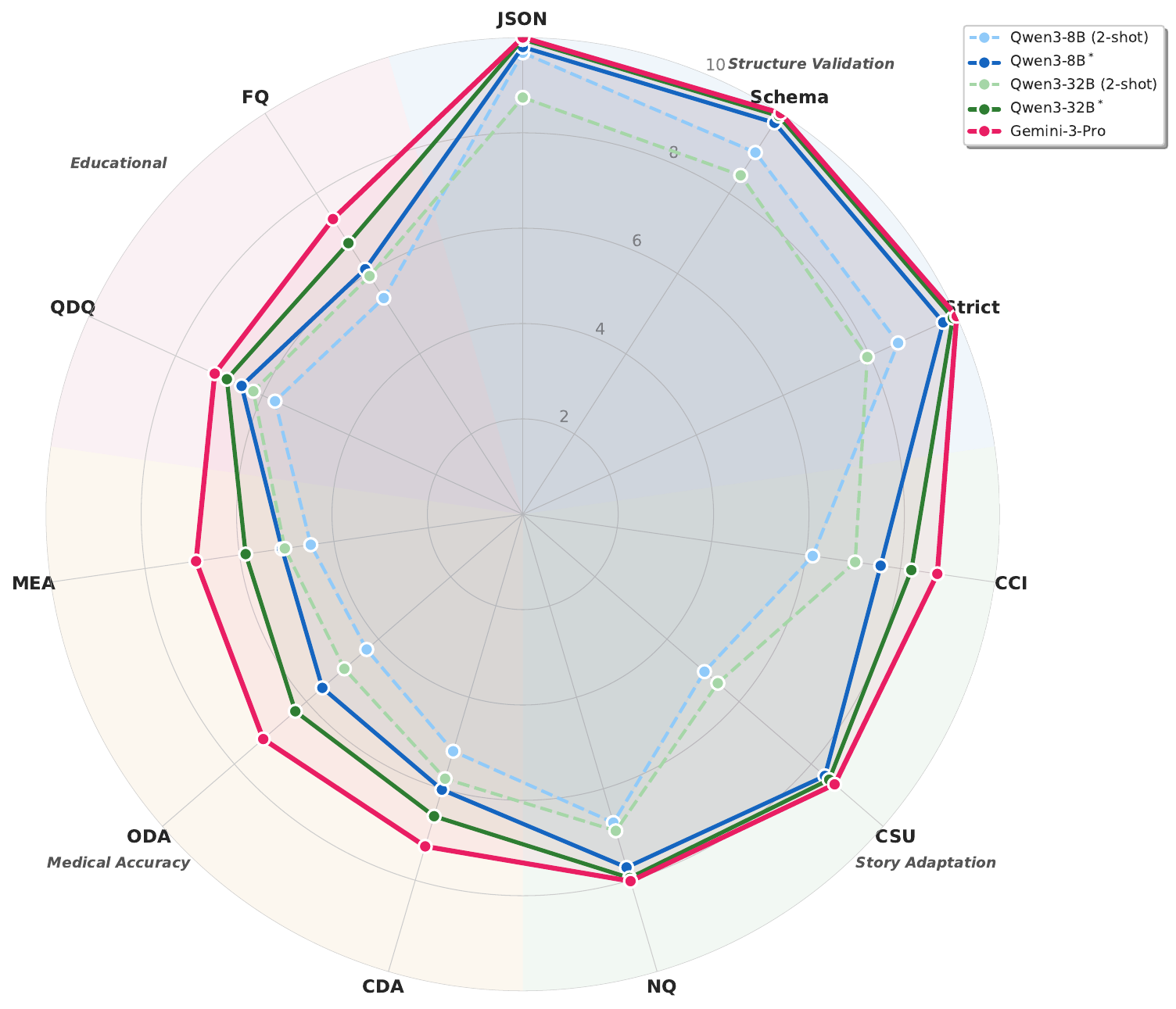}
    \caption{Radar chart comparing key models on Medical Narrative Generation. Dashed lines represent 2-shot prompted models, while solid lines indicate fine-tuned models (marked with $^*$). The background colors delineate four evaluation dimensions.}
    \label{fig:storygen_radar}
\end{figure}

Figure~\ref{fig:storygen_radar} presents a comprehensive radar chart visualization comparing the performance of key models across all 11 evaluation metrics for the Medical Narrative Generation task. The metrics are organized into four distinct dimensions, represented by different background colors: \textit{Structure Validation} (blue), \textit{Story Adaptation} (green), \textit{Medical Accuracy} (orange), and \textit{Educational Quality} (pink).

Several key observations emerge from this visualization:

\paragraph{Impact of Fine-tuning.} Comparing the 2-shot and fine-tuned versions of the same base model reveals substantial improvements across structure and adaptation metrics. For Qwen3-32B, fine-tuning increases the \textit{Character \& Scene Usage} (CSU) score from 5.41 to 8.50 and improves the \textit{Strict Validation} rate from 79.4\% to 99.1\%, demonstrating better use of the provided narrative assets and stronger compliance with the Clinical Storyline schema.

\paragraph{Dimension-wise Performance Patterns.} Models exhibit varying strengths across different evaluation dimensions. While fine-tuned models achieve near-perfect scores in \textit{Structure Validation} (approaching 10 on the normalized scale), the \textit{Medical Accuracy} dimension remains more challenging, with even the best-performing Gemini-3-Pro achieving scores around 7.0. This suggests that generating medically accurate content requires further attention in future work.

\paragraph{Commercial vs. Open-Source Models.} Gemini-3-Pro achieves perfect structure validation (100\%) and the highest scores in most medical accuracy metrics. However, fine-tuned open-source models become competitive in structural and adaptation-oriented dimensions, indicating that task-specific fine-tuning narrows the gap mainly in format compliance and storyline adaptation while medical accuracy remains more difficult.

\subsection{Effect of Medical-Domain Pretraining on Medical Accuracy}
\label{appendix subsection: medical llm comparison}

As observed in the main results, task-specific fine-tuning substantially improves structural validity and story adaptation, but a persistent gap remains in the \textit{Medical Accuracy} dimension. We therefore examine whether medical-domain specialization provides additional benefit beyond learning the MedGame output format.

To investigate this, we compare scale-matched general-purpose models with medical-domain counterparts under the same Medical Narrative Generation fine-tuning protocol. The 8B and 32B comparisons pair Qwen3 with the II-Medical series, medical post-trained variants built on the corresponding Qwen3 backbones~\cite{2025II-Medical-8B,2025II-Medical-32B-Preview}; the 27B comparison uses Gemma-3-27B and MedGemma-27B. Table~\ref{tab:medical_llm_medical_accuracy} summarizes the Medical Accuracy comparison, with significance computed using one-sided paired t-tests over matched test samples.

\begin{table}[htbp]
\centering
\scriptsize
\caption{Medical Accuracy comparison between scale-matched general-purpose and medical LLMs after Medical Narrative Generation fine-tuning. Significance is computed with one-sided paired t-tests over matched test samples; $^{***}$: $p<0.001$.}
\label{tab:medical_llm_medical_accuracy}
\resizebox{\columnwidth}{!}{
\begin{tabular}{lcccc}
\toprule
Model & CDA$\uparrow$ & ODA$\uparrow$ & MEA$\uparrow$ & Avg.$\uparrow$ \\
\midrule
Qwen3-8B & 6.02 & 5.56 & 5.10 & 5.56 \\
II-Medical-8B & 6.36 & 6.02 & 5.60 & 5.99 \\
\rowcolor{gray!15} $\Delta$ & +0.34$^{***}$ & +0.46$^{***}$ & +0.49$^{***}$ & +0.43$^{***}$ \\
\midrule
Qwen3-32B & 6.60 & 6.31 & 5.87 & 6.26 \\
II-Medical-32B & 6.72 & 6.46 & 6.02 & 6.40 \\
\rowcolor{gray!15} $\Delta$ & +0.12$^{***}$ & +0.15$^{***}$ & +0.15$^{***}$ & +0.14$^{***}$ \\
\midrule
Gemma-3-27B & 6.39 & 6.01 & 5.56 & 5.98 \\
MedGemma-27B & 6.64 & 6.33 & 5.91 & 6.29 \\
\rowcolor{gray!15} $\Delta$ & +0.25$^{***}$ & +0.33$^{***}$ & +0.35$^{***}$ & +0.31$^{***}$ \\
\bottomrule
\end{tabular}
}
\end{table}

\paragraph{Key Findings.} Medical-domain pretraining consistently improves Medical Accuracy across all model pairs. The average CDA/ODA/MEA gain is largest for the 8B pair (+0.43), smaller for the 32B pair (+0.14), and substantial for the 27B Gemma/MedGemma pair (+0.31). This pattern suggests that task-specific fine-tuning and medical-domain pretraining contribute different strengths: fine-tuning helps models learn MedGame's structure and interaction requirements, while medical pretraining improves clinical decision, option design, and explanation quality. We therefore interpret medical pretraining as a targeted improvement for clinical correctness rather than as a replacement for task-specific supervision or expert review.

\subsection{Expert Revision Analysis}
\label{appendix subsection: expert revision analysis}

Table~\ref{tab:progressive_revision} summarizes progressive expert revision on 10 low-scoring story drafts per model from Gemini-3-Pro and Qwen3.5-27B$^*$ (20 drafts total). Experts marked targeted revision regions (e.g., inaccurate decisions, misleading options, incomplete explanations, and weak feedback), averaging 8.3 and 8.9 regions per sample, respectively. We apply these revisions progressively and re-evaluate each checkpoint with GPT-5.2 LLM-as-a-Judge at 25\%, 50\%, 75\%, and 100\% completion.

\begin{table}[htbp]
\centering
\scriptsize
\caption{LLM-as-a-Judge scores under progressive expert revision. Experts marked 8.3 and 8.9 revision regions per sample for Gemini-3-Pro and Qwen3.5-27B$^*$, respectively. Revision percentage denotes the fraction of expert-marked regions applied before GPT-5.2 re-evaluation.}
\label{tab:progressive_revision}
\resizebox{\columnwidth}{!}{
\begin{tabular}{llccccc}
\toprule
Model & Revision & CDA & ODA & MEA & QDQ & FQ \\
\midrule
\multirow{5}{*}{Gemini} & 0\% & 6.70 & 6.40 & 6.10 & 6.70 & 6.20 \\
 & 25\% & 7.10 & 6.80 & 6.70 & 6.90 & 6.80 \\
 & 50\% & 7.00 & 6.60 & 7.00 & 6.80 & 7.10 \\
 & 75\% & 7.50 & 7.20 & 7.50 & 7.00 & 7.40 \\
 & 100\% & 8.10 & 7.90 & 8.20 & 7.80 & 8.10 \\
\midrule
\multirow{5}{*}{Qwen3.5-27B$^*$} & 0\% & 6.50 & 6.20 & 5.90 & 6.60 & 6.10 \\
 & 25\% & 6.90 & 6.50 & 6.40 & 6.70 & 6.60 \\
 & 50\% & 6.70 & 6.30 & 6.80 & 6.60 & 6.80 \\
 & 75\% & 7.20 & 6.90 & 7.20 & 6.90 & 7.20 \\
 & 100\% & 7.90 & 7.70 & 8.00 & 7.60 & 7.90 \\
\bottomrule
\end{tabular}
}
\end{table}

\subsection{Impact of Similar Cases on Model Performance}
\label{appendix subsection: similar patient analysis}

The PMC-Patients~\cite{PMC_Patient} dataset includes a \texttt{similar\_patients} field that links patients discussed together in the same publication. It is important to note that this similarity is defined by the original authors' editorial choices rather than by clinical or demographic criteria—patients may be grouped together because they represent different manifestations of the same disease, belong to a case series from a single institution, or serve as comparison cases in the original study.

We investigate whether test samples with such related cases present in the training set would exhibit higher model performance. Specifically, we ask: \textit{Does exposure to author-grouped similar cases during training provide any measurable advantage in generating clinical narratives for related patients?}

\paragraph{Experimental Setup.} Across our test set of 1,000 samples, we identified 409 samples (40.9\%) that had at least one author-defined similar patient in the training set of 4,000 samples. We compared the LLM-as-a-Judge scores between these two groups using independent samples t-tests and computed Cohen's $d$ to measure effect size.

\paragraph{Results.} Table~\ref{tab:similar_patient_analysis} presents the comparison results. For both Qwen3-8B and Qwen3-32B, we observe no statistically significant difference between samples with and without related cases in training (all $p > 0.05$). The effect sizes are negligible (Cohen's $d < 0.05$), indicating that any observed differences are practically meaningless.

\begin{table}[htbp]
    \centering
    \caption{Comparison of model performance between test samples with and without author-defined similar cases in the training set. Total score is out of 80 points. None of the differences are statistically significant ($p > 0.05$).}
    \resizebox{\linewidth}{!}{
    \begin{tabular}{l|cc|ccc}
    \toprule
    \multirow{2}{*}{Model} & \multicolumn{2}{c|}{Total Score} & \multirow{2}{*}{$\Delta$} & \multirow{2}{*}{$p$-value} & \multirow{2}{*}{Cohen's $d$} \\
    & With Similar & Without Similar & & & \\
    \midrule
    Qwen3-8B & 53.14 $\pm$ 5.96 & 52.84 $\pm$ 6.27 & +0.30 & 0.447 & 0.049 \\
    Qwen3-32B & 57.07 $\pm$ 4.50 & 57.03 $\pm$ 4.50 & +0.04 & 0.896 & 0.008 \\
    \bottomrule
    \end{tabular}
    }
    \label{tab:similar_patient_analysis}
\end{table}

\paragraph{Dimension-wise Analysis.} We further examined whether specific evaluation dimensions might be affected. Table~\ref{tab:similar_patient_dimension} shows that none of the three dimensions (Story Adaptation, Medical Accuracy, Educational Value) exhibit significant differences for either model.

\begin{table}[htbp]
    \centering
    \caption{Dimension-wise comparison for Qwen3-32B between test samples with and without similar cases in training. SA: Story Adaptation, MA: Medical Accuracy, EV: Educational Value.}
    \resizebox{\linewidth}{!}{
    \begin{tabular}{l|cc|cc}
    \toprule
    Dimension & With Similar & Without Similar & $\Delta$ & $p$-value \\
    \midrule
    SA & 24.73 $\pm$ 1.51 & 24.66 $\pm$ 1.56 & +0.07 & 0.505 \\
    MA & 18.71 $\pm$ 3.29 & 18.81 $\pm$ 3.18 & $-$0.10 & 0.631 \\
    EV & 13.62 $\pm$ 0.81 & 13.55 $\pm$ 0.87 & +0.07 & 0.188 \\
    \bottomrule
    \end{tabular}
    }
    \label{tab:similar_patient_dimension}
\end{table}

\paragraph{Conclusion.} These results provide no evidence that related-case overlap inflates the reported scores. The small and statistically insignificant differences suggest that the fine-tuned models are not simply benefiting from exposure to high-similarity patient variants in the training split.

\section{Platform Design Logic}
\label{sec:appendix-platform}

To bridge the gap between generated clinical narratives and learner interaction, we develop an interactive platform that renders MedGame outputs as playable simulations. This section describes the platform design and core interaction principles.

\subsection{Platform Overview}

The platform adopts a client-server architecture. The frontend (React + TypeScript) handles multimodal content rendering and user interaction, while the backend (FastAPI) serves pre-generated game assets through RESTful APIs. All game content—story trees, character videos, dialogue audio, and scene images—is pre-generated by the MedGame pipeline and stored as static files, enabling the platform to render experiences without runtime generation overhead.

\begin{figure*}[htbp]
\centering
\includegraphics[width=\textwidth]{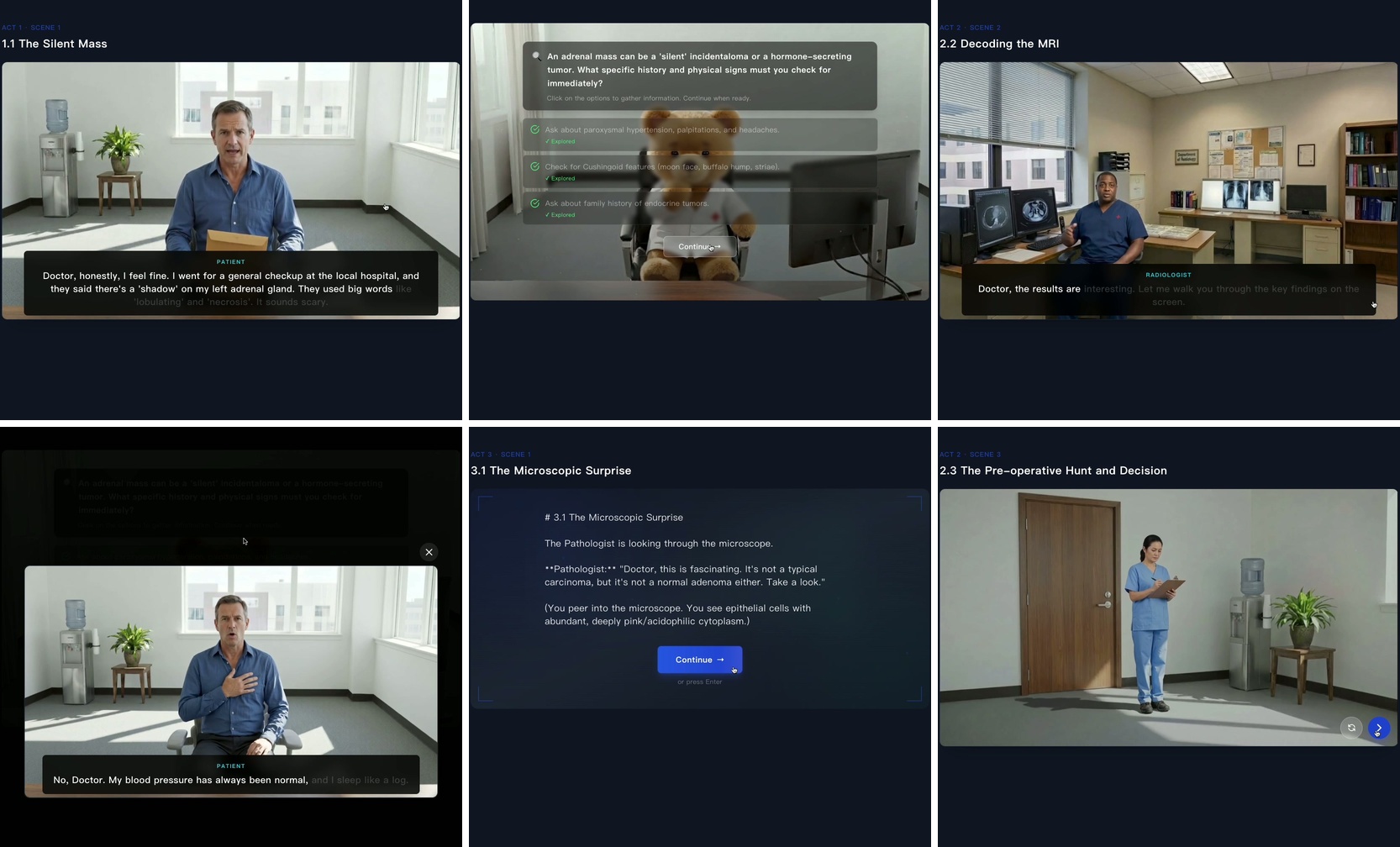}
\caption{Platform interface showcase. \textbf{Top row}: (Left) Film mode with patient dialogue and progressive subtitles; (Center) Interactive mode with explorable options and Doctor Thinking background; (Right) Film mode in radiology room with specialist dialogue. \textbf{Bottom row}: (Left) Interactive dialogue popup triggered by option selection; (Center) Text mode with typewriter-style narrative display; (Right) End roleplay transition with scene navigation controls.}
\label{fig:platform_showcase}
\end{figure*}

\begin{figure*}[htbp]
\centering
\includegraphics[width=\textwidth]{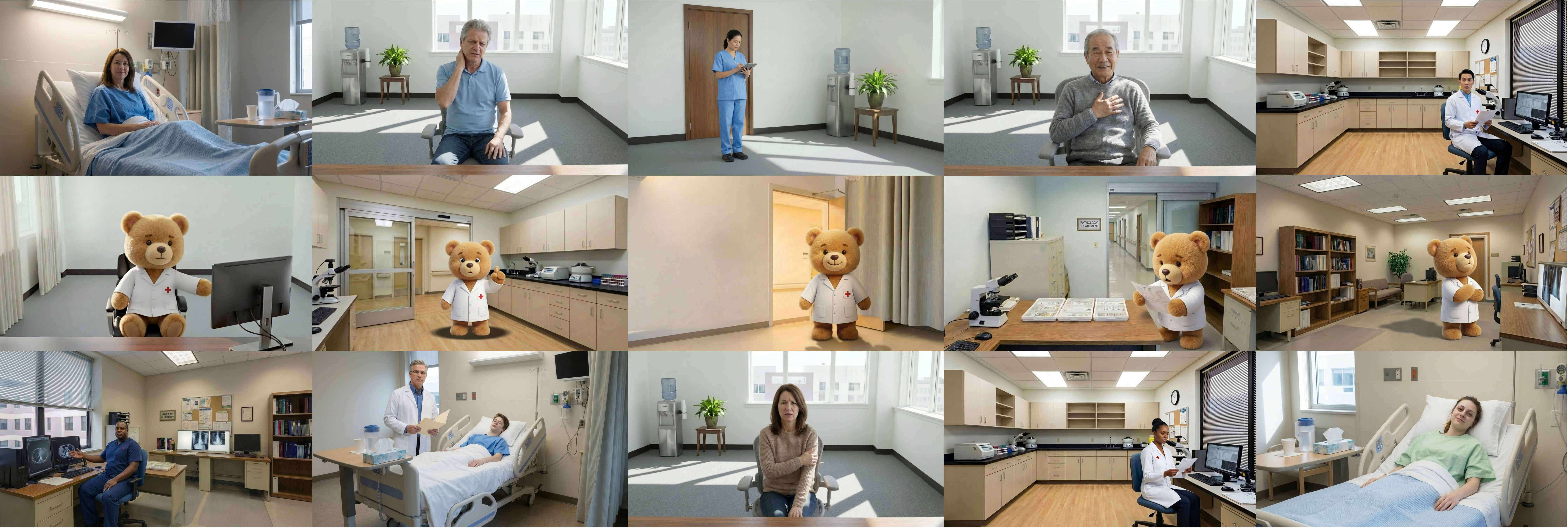}
\caption{Representative MedGame stills rendered for clinical simulation scenes, including patient encounters, clinical staff, and hospital environments.}
\label{fig:medgame_photos}
\end{figure*}

Figure~\ref{fig:platform_showcase} presents representative screenshots from the platform, demonstrating the various content rendering modes and interaction types described below. Figure~\ref{fig:medgame_photos} shows representative MedGame stills rendered for clinical simulation scenes.

\subsection{Game Flow State Machine}

The platform implements a hierarchical state machine that governs the progression through clinical narratives. Figure~\ref{fig:game_state_machine} shows the state transition logic.

\begin{figure}[htbp]
\centering
\resizebox{\columnwidth}{!}{%
\begin{tikzpicture}[
    state/.style={rectangle, rounded corners=5pt, draw=#1!70, fill=#1!12, minimum width=3.2cm, minimum height=1.1cm, align=center, font=\normalsize\bfseries},
    decision/.style={diamond, draw=gray!60, fill=gray!8, minimum width=2.2cm, minimum height=1.1cm, align=center, font=\small, aspect=2},
    arrow/.style={-{Stealth[length=2.5mm]}, very thick, draw=gray!50},
    note/.style={font=\small, text=gray!60, align=left},
    sceneloop/.style={rectangle, rounded corners=8pt, draw=purple!50, fill=purple!5, dashed, thick, inner sep=14pt}
]

\node[state=gray, minimum width=2.5cm] (start) at (0, 7) {Case Selection};

\node[state=blue] (actintro) at (0, 5.2) {ACT\_INTRO};
\node[note, right=0.5cm of actintro] {Display act background\\and patient profile};

\node[state=teal] (startrp) at (0, 3) {START\_ROLEPLAY};
\node[note, right=0.5cm of startrp] {Film / Text / Interleaved};

\node[state=orange] (interact) at (0, 1.2) {INTERACTION};
\node[note, right=0.5cm of interact] {Single / Batch / Interactive};

\node[state=teal] (endrp) at (0, -0.6) {END\_ROLEPLAY};
\node[note, right=0.5cm of endrp] {Film / Text / Interleaved};

\begin{scope}[on background layer]
\node[sceneloop, fit=(startrp)(interact)(endrp), label={[font=\normalsize\bfseries, text=purple!70]above:Scene Loop}] (sceneloop) {};
\end{scope}

\node[decision] (morescene) at (0, -2.5) {More\\Scenes?};

\node[decision] (moreact) at (0, -4.6) {More\\Acts?};

\node[state=green!70!black] (summary) at (3.5, -4.6) {SUMMARY};

\draw[arrow] (start) -- (actintro);
\draw[arrow] (actintro) -- (startrp);
\draw[arrow] (startrp) -- (interact);
\draw[arrow] (interact) -- (endrp);
\draw[arrow] (endrp) -- (morescene);

\draw[arrow] (morescene.west) -- ++(-1.8, 0) node[above, font=\small, pos=0.5] {Yes} |- (startrp.west);

\draw[arrow] (morescene.south) -- node[right, font=\small] {No} (moreact);

\draw[arrow] (moreact.west) -- ++(-2.8, 0) node[above, font=\small, pos=0.5] {Yes} |- (actintro.west);

\draw[arrow] (moreact.east) -- node[above, font=\small] {No} (summary);

\end{tikzpicture}%
}
\caption{Game flow state machine. The platform cycles through roleplay-interaction-roleplay phases within each scene, advancing through scenes and acts until completion.}
\label{fig:game_state_machine}
\end{figure}

Each scene follows a three-phase structure: (1) \textbf{Start Roleplay} presents the narrative context through cinematic videos or text; (2) \textbf{Interaction} engages the learner with clinical decision points; and (3) \textbf{End Roleplay} concludes the scene and transitions to the next. This design mirrors the pedagogical structure defined in the Clinical Story schema (Appendix~\ref{appendix section: clinical story structure}).

\subsection{Content Rendering Modes}

The platform supports three rendering modes for narrative content, corresponding to the \texttt{type} field in the story schema:

\begin{itemize}[leftmargin=*, nosep]
    \item \textbf{Film Mode}: Renders synchronized video with separate audio track and progressive subtitle overlay. The audio-video synchronization ensures lip-sync accuracy, while subtitles reveal progressively to match speech timing.
    \item \textbf{Text Mode}: Displays narrative text with typewriter animation for immersive reading.
    \item \textbf{Interleaved Mode}: Alternates between film and text segments within a single roleplay, enabling hybrid presentations (e.g., dialogue video followed by examination report text).
\end{itemize}

\subsection{Interaction Components}

The platform implements three interaction types that map directly to the Decision Node schema:

\begin{itemize}[leftmargin=*, nosep]
    \item \textbf{Single-Choice}: Radio-button selection for discrete clinical decisions. Immediate feedback with explanation is provided upon selection.
    \item \textbf{Batch}: Checkbox multi-selection for scenarios requiring multiple actions (e.g., ordering a panel of laboratory tests). Feedback is provided after submission.
    \item \textbf{Interactive}: Card-based information gathering that simulates clinical inquiry. Each option reveals different information (patient responses, examination findings), and may trigger dialogue videos.
\end{itemize}

During interactions, a looping ``Doctor Thinking'' video plays in the background, showing the player character in a contemplative state. This maintains visual continuity and reinforces the first-person perspective.

\stopcontents[appendices]

\end{document}